\documentclass{article}

 \PassOptionsToPackage{numbers, compress}{natbib}
\usepackage[preprint]{neurips_2023}
\usepackage[utf8]{inputenc} % allow utf-8 input
\usepackage[T1]{fontenc}    % use 8-bit T1 fonts
\usepackage{hyperref}       % hyperlinks
\usepackage{url}            % simple URL typesetting
\usepackage{booktabs}       % professional-quality tables
\usepackage{amsfonts}       % blackboard math symbols
\usepackage{nicefrac}       % compact symbols for 1/2, etc.
\usepackage{microtype}      % microtypography
\usepackage[table,xcdraw]{xcolor}
\usepackage{wrapfig}
\usepackage{times}
\usepackage{latexsym}
\usepackage{subfigure}
\usepackage{xcolor}
\newcommand{\stitle}[1]{\vspace{1ex}\noindent{\bf #1}}
\newcommand{\softmax}{\mathrm{softmax}}
\usepackage{amsmath} 
\DeclareMathOperator*{\argmax}{arg\,max}

\usepackage{inconsolata}
\usepackage{algorithm}
\usepackage{algorithmic}
\usepackage{listings}
\usepackage{newfloat}
\usepackage{listings}
\usepackage{titlesec}
\usepackage{verbatim}
\usepackage{multirow}
\usepackage{booktabs}
\usepackage{amssymb}
\usepackage{amsmath}
\usepackage{bm}
\usepackage{color, soul, colortbl}
\usepackage[table]{xcolor}
\usepackage{colortbl}
\usepackage{makecell}
\usepackage{caption}

\usepackage{subfigure}
\usepackage{graphicx}
\usepackage{pdfpages}
\usepackage[capitalise]{cleveref}

\usepackage[algo2e]{algorithm2e} 
\definecolor{wingreen}{rgb}{0,0.45,0.24}
\definecolor{losered}{rgb}{1.0,0.1,0.24}
\definecolor{lightgray2}{rgb}{0.8,0.8,0.8}
\newcommand{\done}{\cellcolor{lightgray2}}

\newcommand{\methodname}[1]{DP}
\newcommand{\basea}[1]{base}
\newcommand{\baseb}[1]{base}
\newcommand{\data}[1]{data}

\title{Dynamic Prompting: A Unified Framework for Prompt Tuning}

\author{
Xianjun Yang$^1$\thanks{ Work done during the internship at NEC Laboratories America. }, Wei Cheng$^2$, Xujiang Zhao$^2$, Wenchao Yu$^2$,\\
\textbf{ Linda Petzold$^1$, Haifeng Chen$^2$} \\
\\
$^1$University of California, Santa Barbara \\
\texttt{\{xianjunyang, petzold\}@ucsb.edu }  \\
\\
$^2$NEC Laboratories America  \\
\texttt{\{weicheng, xuzhao, wyu, haifeng\}@nec-labs.com }\\
\\
}

\begin{document}

\maketitle

\begin{abstract}
It has been demonstrated that the art of prompt tuning is highly effective in efficiently extracting knowledge from pretrained foundation models, encompassing pretrained language models (PLMs), vision pretrained models, and vision-language (V-L) models. However, the efficacy of employing fixed soft prompts with a predetermined position for concatenation with inputs for all instances, irrespective of their inherent disparities, remains uncertain. Variables such as the position, length, and representations of prompts across diverse instances and tasks can substantially influence the performance of prompt tuning.
In this context, we provide a theoretical analysis, which reveals that optimizing the position of the prompt to encompass the input can capture additional semantic information that traditional prefix or postfix prompt tuning methods fail to capture. 
Building upon our analysis, we present a unified dynamic prompt (DP) tuning strategy that dynamically determines different factors of prompts based on specific tasks and instances. To accomplish this, we employ a lightweight learning network with Gumble-Softmax, allowing us to learn instance-dependent guidance. Experimental results underscore the significant performance improvement achieved by dynamic prompt tuning across a wide range of tasks, including NLP tasks, vision recognition tasks, and vision-language tasks.
Furthermore, we establish the universal applicability of our approach under full-data, few-shot, and multitask scenarios. Codes are available at \url{https://github.com/Xianjun-Yang/DPT}.
\end{abstract}

\section{Introduction}

In recent years, the prevailing trend in the field of machine learning has been to employ a two-step approach for achieving remarkable performance in natural language processing (NLP) tasks \citep{devlin-etal-2019-bert, Liu2019RoBERTaAR,raffel2020exploring, Lewis2019BARTDS}, computer vision tasks \citep{jia2022vpt,khattakMaPLe,zhou2022coop,zhou2022cocoop}, and multi-modal vision-language tasks\citep{radford2021learning,zhou2022cocoop,zhou2022coop,shu2022tpt,jin2022good}. This approach involves initial pre-training on vast and diverse datasets, followed by fine-tuning (FT) on smaller, task-specific supervised datasets. While this methodology has yielded remarkable results, it poses challenges when dealing with the ever-growing size of models, ranging from 3 billion parameters for T5 \citep{raffel2020exploring} to a staggering 175 billion parameters for GPT-3 \citep{brown2020language}.
Consequently, the research community has fervently dedicated efforts to developing novel methods aimed at achieving parameter-efficient adaptation. Two prominent strategies in this realm are prefix-tuning (PFT) \citep{li-liang-2021-prefix} and prompt-tuning (PT) \citep{lester-etal-2021-power}. These approaches selectively fine-tune a small subset of parameters, distinct from the original pre-trained model, thereby circumventing the costly process of fine-tuning the entire foundation model. Another exploration involves sparse fine-tuning \citep{ben-zaken-etal-2022-bitfit}, which focuses solely on optimizing the bias terms of pre-trained language models (PLMs). Besides, researchers have delved into designing sparse modules, exemplified by the ingenious employment of hypercomplex multiplication (PHM) layers \citep{zhang2021beyond}, compacter models \citep{karimi2021compacter}, and Low-Rank Adaptation (LoRA) \citep{Hu2021LoRALA}. By manipulating a considerably reduced number of parameters, these methods often exhibit comparable performance to traditional fine-tuning, particularly in the context of billion-scale PLMs, a phenomenon commonly referred to as the power of scale \citep{lester-etal-2021-power}.

\begin{figure*}[!t]\vspace{-0.3cm}
    \centering
    \subfigure[Average acc. across different pre-trained model sizes]{\includegraphics[width=2.4in]{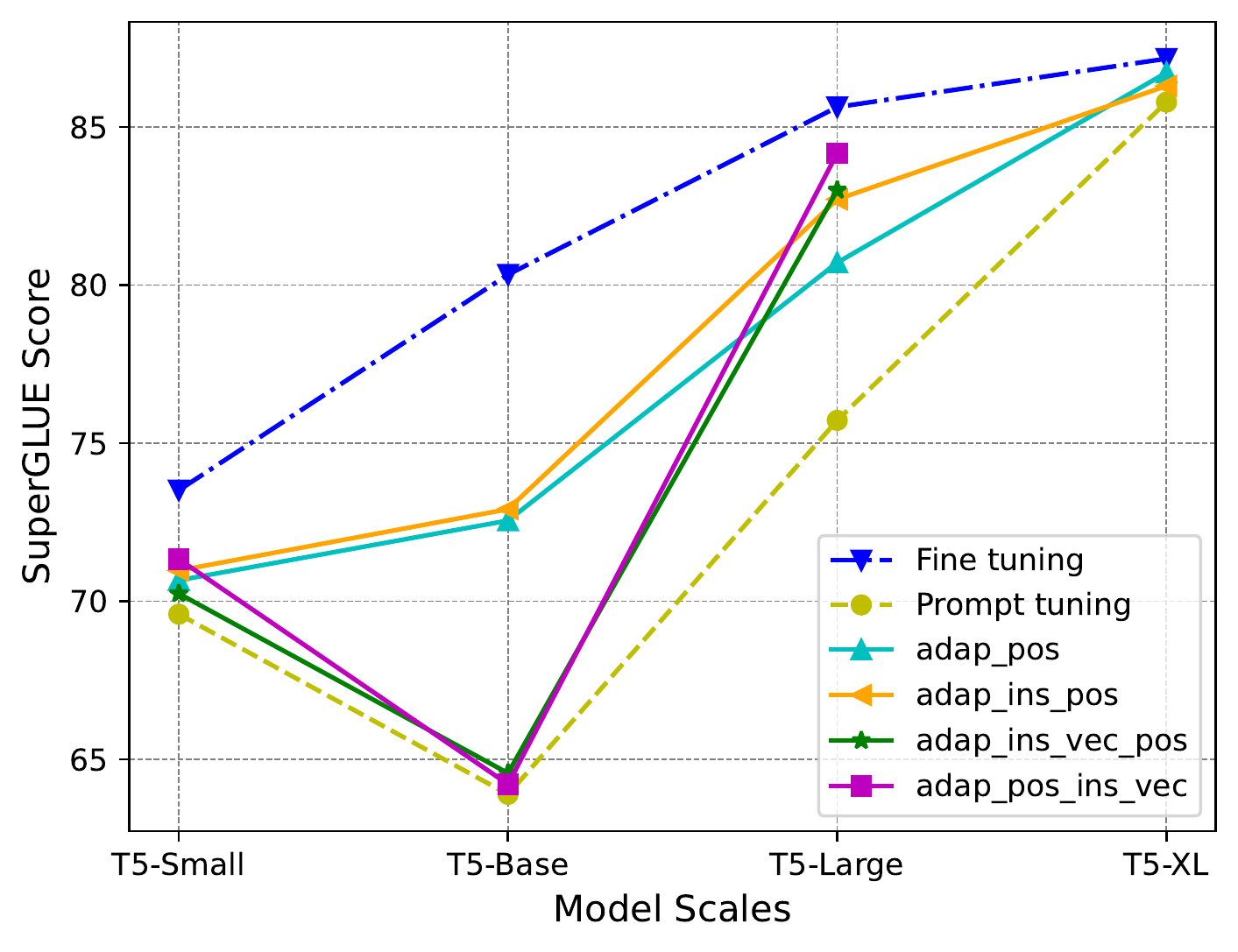}\label{fig:results}}
     % \hfill
     \hspace{1cm}
    \subfigure[Performance comparison across different SuperGLUE datasets on T5-Large]{\includegraphics[width=2.3in]{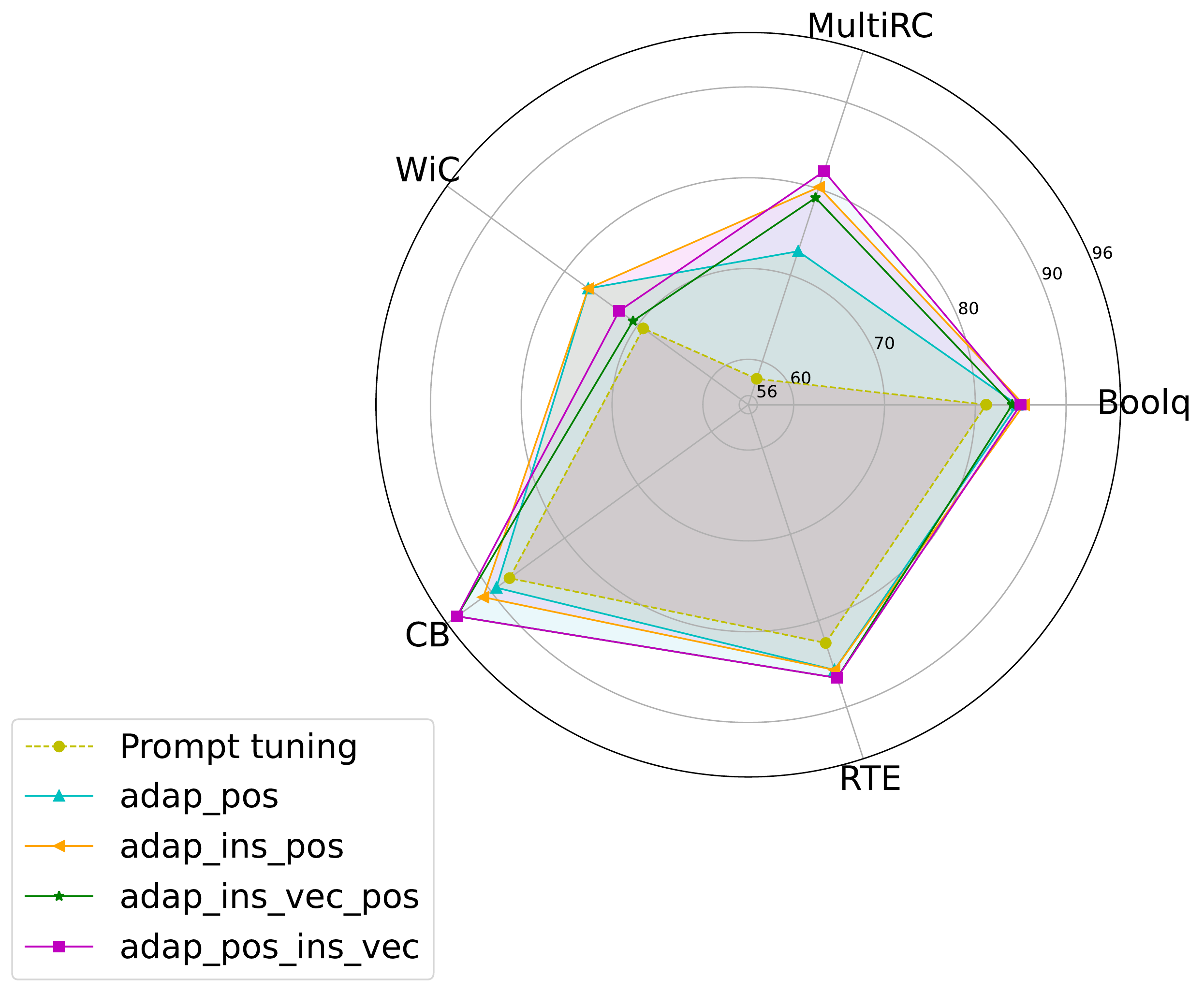}\label{fig:results_radar}}
    \caption{Standard prompt tuning achieves suboptimal scores on SuperGLUE. Our dynamic prompting (\textit{adap\_*}) consistently yields superior results. Fine-tuning results on T5-XL are reported in \citep{aribandi2021ext5}. }
    \label{fig:results_all}\vspace{-0.2cm}
\end{figure*}

The pre-train, prompt, and predict paradigm, as elucidated in \citep{liu2023pre}, may be classified into the realms of soft and hard prompts, thereby enabling the seamless optimization and resplendent visualization of this groundbreaking approach. Nonetheless, the majority of previous research has either maintained a static set of optimized prompts \citep{Ma2022XPromptET, Liu2021PTuningVP} across all instances in a task or exclusively added prompts to the beginning of all inputs \citep{vu-etal-2022-spot, Guo2022ImprovingTS, gu-etal-2022-ppt}. Very recently, \citep{wu-etal-2022-idpg} proposes to generate the instance-dependent prompts by an Adapter \citep{houlsby2019parameter} module for NLU tasks and show noticeable performance improvement. Similarly, \citep{Asai2022ATTEMPTPM} dynamically changes the instance-dependent soft prompts via attentional mixtures of prompts learned from multi-tasks. However, they still adopt a fixed position and fixed length for concatenating prompts with inputs, which might be the suboptimal strategy. 

\begin{figure*}\vspace{-0.33cm}
\centering    \includegraphics[width=0.83\textwidth]{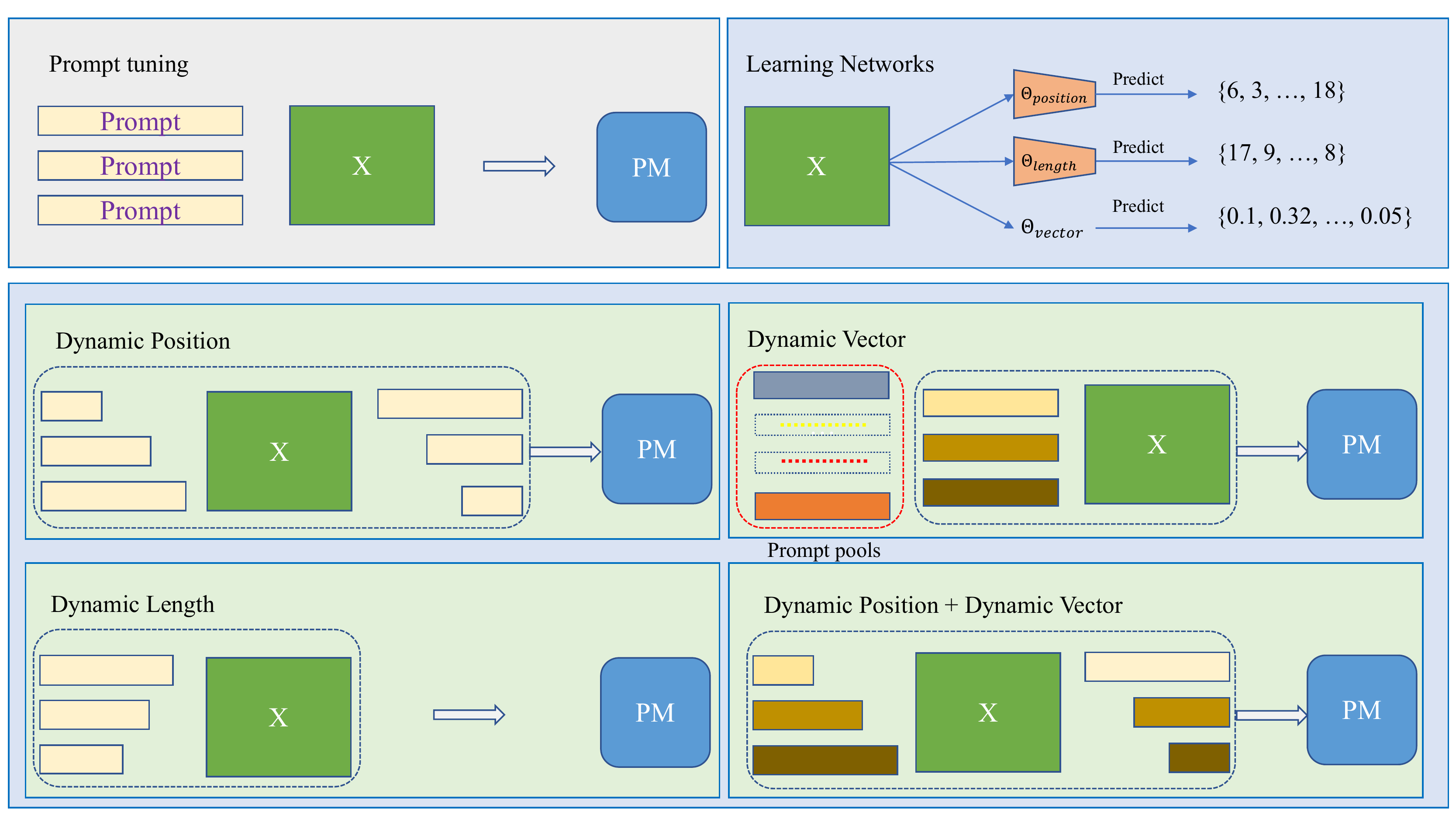} \vspace{-0.2cm}
    \caption{ An overview of our approach. The learning networks first predict the task- or instance-dependent prompt position, length, and prompt pools. Then the new soft prompt is concatenated with instances to be fed into the frozen language, vision, or V-L models for prediction. Parameters in the prompts and learning networks are simultaneously updated. Canonical prompt tuning can be seen as a special case of our dynamic prompting, while the same soft prompt is prepended for all instances. }\label{fig:overview} \vspace{-0.3cm}
\end{figure*}

To the best of our knowledge, a systematic exploration of the dynamic manipulation of prompt position, length, and prompt pools remains absent from the literature. Driven by this gap, we are inspired to embark on an investigation of various approaches to dynamic prompting (DP), aiming to extract valuable insights from large foundation models. Our endeavor encompasses a comprehensive theoretical analysis that unravels the potential benefits of optimizing the position for concatenating prompts with inputs, thereby capturing additional semantics that conventional prefix or postfix prompt tuning methods fail to encapsulate. Motivated by our analysis, we propose a unified strategy for dynamic prompt (DP) tuning, wherein different factors are dynamically determined based on the specific tasks or instances at hand. An overview of our novel approach is summarized in Figure \ref{fig:overview}. Specifically, we employ a one-layer feedforward network in conjunction with the Gumbel-Softmax technique \citep{DBLP:conf/iclr/JangGP17} to learn the categorical distribution of position or length, facilitating optimization at both the task and instance levels. To thoroughly evaluate the efficacy of dynamic concatenation position, length, and prompt pools, we conduct independent as well as simultaneous experiments on the SuperGlue benchmark \citep{wang2019superglue}. The empirical results of our study showcase remarkable performance improvements attained through dynamic prompting, spanning models of varying sizes. This advancement effectively narrows the gap between prompt tuning and traditional fine-tuning, as illustrated in Figure \ref{fig:results_all}. Our findings underscore the immense potential of dynamic prompting in enhancing model capabilities and shedding light on its significance in the realm of large-scale pretrained models.

Our approach serves as a versatile and potent component, capable of seamlessly integrating into a wide array of problem domains to unlock superior performance. Beyond its application in prompt tuning and P-tuning v2 for NLP tasks, as established in prior research \citep{Liu2021PTuningVP}, we extend the reach of our approach to encompass other methodologies. Notably, we successfully apply our framework to vision prompt tuning (VPT) \citep{jia2022vpt} and MaPLe \citep{khattakMaPLe}, catering to the realm of multi-modal prompt learning. The incorporation of our dynamic prompting technique yields additional accuracy gains across these diverse methodologies. Furthermore, our study showcases the effectiveness of dynamic prompting not only in the single-task setting but also in multi-task and few-shot learning scenarios. This broader applicability further amplifies the potential impact of our work and solidifies its relevance in various learning paradigms. %In conclusion, our research not only enhances the performance of parameter-efficient prompt tuning but also opens up new avenues for advancements in this field. 

The key contributions of this work include the following:
\begin{itemize}
\item We are the first to propose dynamic prompting with instance-dependent prompt position, length, and representation.
\item Our innovative research has yielded a comprehensive framework for elucidating the mechanism underlying the superior performance of dynamically adjusted soft prompts in comparison to conventional prompt tuning methods.
\item We conduct experiments to validate the efficacy of our methods across a wide range of tasks, including NLP tasks, vision recognition tasks, and vision-language tasks.
\end{itemize}

\section{Related Work}
\stitle{Prompt Tuning.}
Prompt tuning was introduced by \citep{lester-etal-2021-power}, a simple yet effective mechanism for learning “soft prompts” to condition frozen language models to perform specific downstream tasks as an alternative to prefix-tuning \citep{li-liang-2021-prefix}, which optimizes some specific layers prefixed to each transformer layer. Those kinds of parameter-efficient tuning methods \citep{Liu2022FewShotPF, Jia2022VisualPT, Chen2022RevisitingPT} have been shown to be powerful while tuning much fewer parameters compared with traditional fine-tuning. 
In P-Tuning v2 \citep{Liu2021PTuningVP}, prompt tuning is improved to be comparable to fine-tuning by prepending prompts on each transformer input layer, increasing the complexity and tunable parameters.
\citep{su-etal-2022-transferability} improves PT via prompt transfer and investigates the transferability of soft prompts across different downstream tasks. They find that the trained soft prompts can effectively transfer to similar tasks, significantly accelerate training, and improve performance when used as initialization. \citep{vu-etal-2022-spot} also shows that learning a prompt on one or more source tasks and then using it to initialize the prompt for a target task can outperform standard model tuning. \citep{Ma2022XPromptET} further explores the limits of prompt tuning by eliminating the negative prompt tokens through a hierarchically structured pruning. \citep{wei2021pretrained} theoretically proves that prompt tuning obtains downstream guarantees with weaker non-degeneracy conditions. Recently, prompt tuning was also introduced to vision tasks \citep{jia2022vpt,chen2022adaptformer,Lian_2022_SSF}, such as vision prompt for continual learning \citep{wang2022learning}, and for image inpainting \citep{visprompt}. Besides, prompting techniques are also proposed for addressing multi-modal applications, such as vision-language applications \citep{khattakMaPLe,radford2021learning, zhou2022cocoop, zhou2022coop, shu2022tpt, jin2022good}.

\stitle{Instance-dependent Prompt Tuning.}
IDPG \citep{wu2022idpg} proposes an instance-dependent prompt generation method by an up-and-down adapter module. And they utilize the parameterized hypercomplex multiplication (PHM) layers \citep{zhang2021beyond} to reduce the number of parameters of the generation module. \citep{Asai2022ATTEMPTPM} improves instance-dependent prompts by an attentional mixture of source multi-task prompts, where the source prompts are pre-trained in a multi-task way, which is also adopted in \citep{Sun2022MultiTaskPO, Asai2022AttentionalMO}. Very recently, \citep{Dai2022WhyCG, Oswald2022TransformersLI} shows that transformer-based PLMs learn in-context by gradient descent, which might explain why different in-context examples can significantly change the results \citep{liu2021makes}. Diversified in-context examples can be seen as hard prompts, and different hard prompts selection can significantly impact more challenging reasoning tasks \citep{zhang2022automatic}. Inspired by such observation, we believe dynamically adjusted soft prompts are also vital for efficient prompt tuning. Besides, \citep{he2021towards} gives a unified view of various parameter-efficient learning methods by looking at the attention and feedforward layers and treating prompt tuning as a simplified prefix tuning \citep{lester-etal-2021-power, Liu2021PTuningVP}. In this work, we derive a unified framework to include various dynamic prompting methods.

\section{Method}
In this section, we first derive a unified view of prompt tuning in Sec. \ref{unified}, then we describe several dynamic prompting strategies: dynamic position for concatenation with inputs, dynamic length, and dynamic representation in Sec. \ref{form:dp}, as depicted in Figure \ref{fig:overview}.

\subsection{A Unified View}\label{unified}
Unlike canonical prompt tuning \citep{lester-etal-2021-power}, where soft prompts are prepended to inputs, we split the prompts into two parts: \textit{prefix} and \textit{postfix}. Formally,
for a sequence $x\in \mathbb{R}^{m \times d}$, the query matrix is $Q = xW^Q \in \mathbb{R}^{m \times d}$, the key and value matrix are $K = xW^K\in \mathbb{R}^{m \times d}$, and $V = xW^V \in \mathbb{R}^{m \times d_v}$, respectively. The soft prompt $P$ with length $l$ is split into two parts, $P = [P_1; P_2]$, where $P_1 \in \mathbb{R}^{l_1 \times d}$ and $P_2 \in \mathbb{R}^{l_2 \times d}$.
The resulting new input becomes $x' = [P_1; x; P_2] \in \mathbb{R}^{(l_1+m+l_2) \times d}$, and the new key and value become $K' = x'W^K\in \mathbb{R}^{(l_1+m+l_2) \times d}$ and $V' = x'W^V \in \mathbb{R}^{(l_1+m+l_2) \times d_v}$. Here, $[;]$ denotes the concatenation operation. By matrix decomposition, we have:
\begin{equation}
Q' = \begin{bmatrix} Q_1 \\ Q \\ 
Q_2 \end{bmatrix}, K' = \begin{bmatrix} K_1 \\ K \\ K_2 \end{bmatrix}, V' = \begin{bmatrix} V_1 \\ V \\ V_2 \end{bmatrix}, 
\end{equation}
 where $Q_1, K_1 \in \mathbb{R}^{l_1 \times d} $,  $Q_2, K_2 \in \mathbb{R}^{l_2 \times d} $ and $V_1 \in \mathbb{R}^{l_1 \times d_v} $,  $V_2 \in \mathbb{R}^{l_2 \times d_v} $. 

For the new query $x' = [P_1; x; P_2] $, the attention head module becomes:
\begin{align}
    \textit{Head} & = \textit{Attn}\left(\bigl[P_1; x; P_2\bigr]W^Q, \right.
    \left. \bigl[P_1; x; P_2\bigr]W^K, \bigl[P_1; x; P_2\bigr]W^V\right)\\
    &= \softmax\left(\frac{Q'*K'^T}{\sqrt{d}}\right)V'\\[-1.2em]
    \intertext{omitting $\sqrt{d}$ for brevity}\nonumber\\[-1.5em]
    & =\left[ \softmax(P_1W^QK'^T)V'; \softmax(xW^Q K'^T)V' ; \softmax(P_2W^Q K'^T)V' \right]. 
\end{align}

With the definition above, we can derive a unified view of prompt tuning as shown in the following formula. The detailed derivation is included in Appendix \ref{app:derivation}.

\small{
\begin{align}
    &  Head= Attn(x', K', V')\nonumber \\ 
    & = \left [ \lambda_1* \underbrace{Attn(Q_1, K_1, V_1)}_{\textit{ \textcolor{red} {prompt tuning} }} + \lambda_2*\underbrace{Attn(Q_1, K_2, V_2)}_{\textit{  {postfix} }}\right.  +( 1 - \lambda_1 - \lambda_2)*\underbrace{Attn(Q_1, K, V )}_{ \textcolor{red} {\textit{ prompt tuning}}} ; \nonumber \\
    & \beta_1* \underbrace{Attn(Q, K_1, V_1)}_{\textit{\textcolor{red} {prompt tuning} }} + \beta_2*\underbrace{Attn(Q, K_2, V_2)}_{\textit{postfix}} +( 1 - \beta_1 - \beta_2)*\underbrace{ \textcolor{blue} {Attn(Q, K, V )} }_{\textit{ \textcolor{blue} { standard} }};\nonumber\\[-1.2em]
    & \gamma_1* \underbrace{Attn(Q_2, K_1, V_1)}_{\textit{postfix}} + \gamma_2*\underbrace{Attn(Q_2, K_2, V_2)}_{\textit{postfix}}\left. +( 1 - \gamma_1 - \gamma_2)*\underbrace{Attn(Q_2, K, V )}_{\textit{postfix}} \right ].
\end{align}
}\normalsize

In such a unified formulation, $\{\lambda_i, \beta_i, \gamma_i\}_{i=1,2}$ are normalized weights to control how attention is distributed among the (prefix) \textit{prompt tuning}, \textit{postfix}, and \textit{standard} attention. When $P_2$ does not exist, postfix weights are equal to 0, so the result is equivalent to standard prompt tuning. By introducing $P_2$, the overall $Head$ is more flexible to accommodate different query $x$ and potentially provides additional semantics that $P_1$ can not capture. The theoretical analysis reveals that optimizing the position of the prompt to encompass the input can capture additional semantic information that traditional prefix or postfix prompt tuning methods fail to capture. Our dynamic prompting is inspired by such a formulation that diversified prompts are expected for coupling with different queries.%\wc{how?}

\subsection{Dynamic Prompting}\label{form:dp}
In this section, we introduce how we use dynamic prompting (DP) to accommodate tuning with respect to the task- or instance-aware insertion position, length, and representation of soft prompt. 

Following the ancestral prompt tuning \citep{lester-etal-2021-power}, given a sequence $x$ of $n$ tokens, $x = \{x_1, x_2, ..., x_n\}$, a pre-trained foundation model, such as the language model T5 \citep{raffel2020exploring}, generates embedding of the tokens $X \in \mathbb{R}^{n \times d}$ where $d$ is the dimension of the encoded representation. For vision prompt tuning, $x$ is a sequence of visual hidden features \citep{jia2022vpt}. The prompt tuning introduces a soft prompt $P \in \mathbb{R}^{l \times d}$ where $l$ is the length of the soft prompt. The next step is to \textit{prepend} the prompt $P$ with actual inputs $X$ into a matrix $X' = [ P; X ]$, then $X'$ is fed into the model $LM$ for optimization, where only parameters in $P$ is optimized while the backbone $LM$ is frozen.

\stitle{Dynamic Position for Concatenation with Inputs.}\label{sec:dynamicpos}
As noticed above, the concatenation of soft prompt and inputs is simply a prefix of $P$ into $X$. However, we assume this kind of concatenation might not be the optimal strategy. Intuitively, the prefix $P$ provides extra information for the input sequence and offers an optimized alternative, but it might not be sufficient. Thus, we propose dynamic position to fill the gap: integer $dpos$ is a parameter to be learned for different tasks or instances, then the original $P$ can be split into two parts $P = [P_{before}, P_{after}]$, where $P_{before} = [P_1, P_2, ..., P_{dpos}]$ and $P_{after} = [P_{dpos+1}, ..., P_l]$. Thus, the new input to $LM$ becomes 
\begin{equation}
X' = [P_{before}; X; P_{after}],
\end{equation}
where $dpos \in [0, l]$ is an integer to be learned and the ancestral prompt tuning is a special case when $dpos{=}l$. Since $dpos$ is categorical, we use a one-layer network $POS_{\theta}$ and the Gumbel-Softmax \citep{DBLP:conf/iclr/JangGP17} to optimize it. Specifically, given the output of $POS_{\theta}$, 
$\bm{\alpha}\in \mathbb{R}^{l+1}$, we need to estimate a binary vector of the same size. A simple way to implement the binarization function is to select the position with a maximum value of \{$\alpha_0$, $\alpha_1$,$\cdots$,$\alpha_{l}$\}, however, this approach is non-differentiable. 
There are several ways that allow us to propagate gradients through the discrete nodes \citep{bengio2013estimating}. In this work, we adopt the Gumbel-Softmax sampling approach \citep{jang2017categorical,maddison2017concrete}. Thereby, we have
\begin{equation}
logit =\textit{ Gumbel-Softmax}( POS_{\theta}(x), \tau ), 
\end{equation}
where $\tau$ is the annealing temperature adjusted by the total training steps as detailed in Sec. \ref{exp:gumbel}. The $logit$ is an $(l{+}1)$-dimensional binary vector where only one element is equal to one and all other elements are zero. A detailed derivation of using Gumbel-Softmax to get the insertion position is included in Appendix \ref{sec:gumbel}.

Previous research \citep{lester-etal-2021-power} has shown that prompt length is crucial for specific models and tasks, while going beyond 20 soft tokens gives marginal gains. We thus adopt $l{=}20$ for most experiments. In this way, the parameters of the soft prompt are the same as PT when $l$ is fixed, making the comparison fair. The only additional parameters are brought by the small network of $POS_{\theta}$ with one linear layer, which is with size $d {*} (l {+} 1)$. We denote this instance-dependant position selection method as \textit{Adaptive position on instance-level}, abbreviated as \textit{adap\_ ins\_pos}. Notice that for our experiments on learning an optimal position for all instances in a task, we only use a vector $v \in \mathbb{R}^{l{+}1} $ to learn a global best position for all instances within that task. We refer to this method as the \textit{Adaptive Position on Task-Level} (abbreviated as \textit{adap\_ pos}). Then the number of additional parameters is $l{+}1$. 

\stitle{Dynamic Length.}\label{dy_len}
Previous research has shown that prompt length plays a vital role in prompt tuning \citep{lester-etal-2021-power}, and larger LMs usually require a shorter prompt length. But the effect of prompt length on tasks or instances level has been underexplored. We propose that the prompt length can also be dynamically learned:
\begin{equation}
P\in \mathbb{R}^{l^* \times d}, l^*=\arg\!\min_{i} loss( LM([\hat{P}_i;X]|\hat{P}_i \in \{\hat{P}_1,\cdots,\hat{P}_l\}, \hat{P}_i\in \mathbb{R}^{i\times d}) ).
\end{equation}
Similarly, $l^* \in [0, l] $ is also categorical and can be optimized by a one-layer network $LEN_{\theta}$ and Gumbel-Softmax. Here, $l$ represents the maximum permissible length for the selection process. Also, the number of  additional parameters will be $l{+}1$ and $d {*} (l{+}1)$ for task and instance-level, respectively. Nevertheless, implementing such a mechanism poses a practical challenge since models typically necessitate fixed input matrix dimensions. In light of this, we employ a surrogate strategy, which is elaborated upon in Appendix \ref{app:dl}.

\stitle{Dynamic Vector.}\label{sec:dynamicvec}
Extensive evidence \citep{wu2022idpg, Asai2022ATTEMPTPM} supports the advantage of utilizing instance-dependent prompts in downstream tasks. The prompt can either be generated directly through an adapter module, as demonstrated in the study conducted by \citep{wu-etal-2022-idpg}, or by employing attentional mixtures during multi-task training, as illustrated in the research conducted by \citep{Asai2022ATTEMPTPM}. We propose a novel and streamlined method for generating dynamic prompts using prompt pools. This approach simplifies the process and allows for the seamless generation of dynamic prompts. Specifically, suppose there are a set of prompt pools $Pool = \{P^{(1)}, ..., P^{(k)} \}$, where $k$ is the size of the pool. Then given any input $x$, we learn a small network $Po_{\theta}$ to get the attention score of every prompt $P^{(k)}$ with respect to $x$, finally the new soft prompt become:
\begin{equation}
P_{new} = \sum_{i=1}^{k} \beta_i*P^{(i)}, \bm\beta = \softmax(Po_{\theta}(x) ).
\end{equation}
In practice, $k$ controls the size of the prompt pool and increased parameters. Since the $P_{new}$ depends on a specific input instance, we denote this setting as \textit{Adaptive vector on instance-level}.

\stitle{Combination.}\label{sec:combination}
Notice that the previously mentioned methods can be combined to unleash the power of dynamic prompting further. For example, we can simultaneously update dynamic position and prompt pool together, which we denote as \textit{Adaptive instance-vector-position}, shortened as \textit{adap\_ins\_vec\_pos}.
Alternatively, we first use dynamic position to learn the best task-level position and update the instance-level prompt pool, denoted as \textit{Adaptive position-instance-vector}, shortened as \textit{adap\_pos\_ins\_vec}. We leave more combinations for future work.

\section{Experiments}

\stitle{Models.}
For language tasks, we use the OpenPrompt \citep{ding-etal-2022-openprompt} framework for implementing our experiments, which is built on Huggincaface\footnote{https://huggingface.co/} and Pytorch \footnote{https://pytorch.org/}. We use the T5-adaptive-LM\footnote{https://github.com/google-research/text-to-text-transfer-transformer/blob/main/released\_checkpoints.md} version for its superiority for prompt tuning. In all our experiments, we freeze the backbone LMs and only optimize the soft prompts and learning networks for acquiring dynamic information. We choose the initial learning rate \textit{lr} from $[0.1, 0.2, 0.3]$, weight decay to be $1e{-}5$, and Adafactor \citep{shazeer2018adafactor} as the optimizer. Besides, we use the default settings for prompt templates and verbalizers from OpenPrompt\footnote{https://github.com/thunlp/OpenPrompt/blob/main/tutorial/1.4\_soft\_template.py}. Unless otherwise stated, we keep the soft tokens to 20 for all experiments. We evaluate the validation set every 500 steps. For all fine-tuning experiments, we keep the same setting as our prompt tuning, except for initializing the \textit{lr} to $1e{-}5$, removing the added soft prompts, and tuning the whole LMs. 

To assess the efficacy of dynamically learning the optimal position of prompts to comprehensively cover the input, as discussed in Sec. \ref{unified}, we have incorporated our dynamic optimization method into various methodologies, namely P-tuning v2\citep{Liu2021PTuningVP}, vision prompt tuning (VPT)\citep{jia2022vpt}, and MaPLe \citep{khattakMaPLe} for multi-modal prompt learning. Notably, since OpenPrompt solely supports prompt addition at the input layer, we extended our experimentation beyond T5-series pretrained models to include BERT-Large and Roberta-Large models using the P-tuning v2 approach. Moreover, we have verified the effectiveness of our approach in the domain of visual recognition tasks, employing the vision prompt tuning (VPT) framework \citep{jia2022vpt} with ViT-B backbone(``sup\_vitb16\_imagenet21k''), as well as in the context of multi-modal prompt learning for the novel class generalization task, utilizing MaPLe \citep{khattakMaPLe} as the underlying backbone. For vision-language model validation, we conduct the generalization from the Base-to-Novel classes task. Specifically, we evaluate
the generalizability of MaPLe with our dynamic prompt insertion technique and follow a zero-shot setting where the datasets are split into base and novel classes. The model is trained only on the base classes in a few-shot
setting and evaluated on base and novel categories. The backbone model is the ViT-B/16 CLIP model.
For further details on the experimental configurations, please refer to Appendix \ref{sec:expsetting}.

\stitle{Datasets.}
Following previous work \citep{ding-etal-2022-openprompt}, we evaluate our approach on five SuperGLUE \citep{wang2019superglue} datasets to test the language understanding ability, namely BoolQ \citep{clark-etal-2019-boolq}, MultiRC \citep{khashabi-etal-2018-looking}, CB \citep{de2019commitmentbank}, RTE \citep{giampiccolo-etal-2007-third}, and WiC \citep{pilehvar-camacho-collados-2019-wic}. We use the default train/dev split and report the default metric on the validation set since the test set is not directly available. For comparison with P-tuning v2, we use eight SuperGLUE datasets. For the vision prompt tuning setting, we follow \citep{jia2022vpt} and use the well-known FGVC datasets consisting of 5 benchmarked datasets. For the vision-language setting, we follow MaPLe \citep{khattakMaPLe} and use 11 different datasets. Details on the datasets are included in Appendix \ref{sec:datasetapp}.

\section{Results}
\begin{table*}
\centering
\caption{ Compared with only adjusting position in Table \ref{tab:pos}, combining together with the adaptive vector can further close the gap between fine-tuning.  }\label{tab:vec}
\setlength\tabcolsep{3pt}
\scalebox{0.7}{
\begin{tabular}{l@{\hspace{0.8\tabcolsep}}|ccc|ccc|ccc}
\hline
\multirow{3}{*}{\textbf{Dataset}}
 & \multicolumn{3}{c|}{\textbf{T5-LM-Small}} & \multicolumn{3}{c|}{\textbf{T5-LM-Base}} & \multicolumn{3}{c}{\textbf{T5-LM-Large}}   \\
 %\cline{2-7}
 %& \multicolumn{3}{c|}{test}  & \multicolumn{3}{c}{test} \\
  \cline{2-10}
  & \thead{Adaptive \\ Ins\_vec\_pos} & \thead{Adaptive \\ Pos\_ins\_vec} & \thead{Fine \\ tuning} & \thead{Adaptive \\ Ins\_vec\_pos} & \thead{Adaptive \\ Pos\_ins\_vec} & \thead{Fine \\ Tuning} & \thead{Adaptive \\ Ins\_vec\_pos} & \thead{Adaptive \\ Pos\_ins\_vec} & \thead{Fine \\ Tuning} \\
\hline
Boolq & $67.40$ & $\textbf{68.04}$ & \done $71.02$  & $\textbf{62.51}$ & $62.39$&  \done$81.33$    & $84.07$ & $\textbf{84.98}$ & \done$87.25$ \\
MultiRC & $68.92$ & $\textbf{69.12}$ & \done$69.58$  & $\textbf{57.96}$ & $57.65$&  \done$77.91$     & $78.96$ & $\textbf{82.03}$ & \done$85.85$ \\
WiC & $66.30$ & $\textbf{66.61}$ & \done$65.25$     & $60.97$ & $\textbf{64.29}$ &  \done$69.18$     & $70.69$ & $\textbf{72.57}$ & \done$73.82$ \\
CB & $82.14$ & $\textbf{85.71}$ &  \done$92.86$     & $\textbf{80.36}$ & $75.00$&  \done$94.62$    & $\textbf{94.64}$ & $\textbf{94.64}$ & \done$94.64$ \\
RTE & $66.42$ & $\textbf{67.15}$ & \done$68.84$     & $61.01$ & $\textbf{61.73}$ &   \done$78.62$   & $\textbf{86.64}$ & $\textbf{86.64}$& \done$86.59$ \\
\hline
Avg. & $70.24$ & $\textbf{71.33}$& \done$73.51$   & $\textbf{64.56}$ & $64.21$&  \done$80.33$   & $83.00$ & $\textbf{84.17}$ & \done$85.63$ \\
\hline
\end{tabular}
}
\vspace{-0.2cm}
\end{table*}

\begin{table} 
\centering
\begin{minipage}{0.46\linewidth}
\centering
\caption{Fixed length PT v.s. adaptive length.
}\label{tab:len}
\setlength\tabcolsep{3pt}
\scalebox{0.8}{
\begin{tabular}{l@{\hspace{0.8\tabcolsep}}|cc|cc}
\hline
\multirow{3}{*}{\textbf{Dataset}}
& \multicolumn{2}{c|}{\textbf{T5-LM-Base}} & \multicolumn{2}{c}{\textbf{T5-LM-Large}}  \\
 %\cline{2-7}
 %& \multicolumn{3}{c|}{test}  & \multicolumn{3}{c}{test} \\
  \cline{2-5}
  &  \thead{Fixed\\Length}  & \thead{Adaptive \\ Length}  &  \thead{Fixed\\Length}  & \thead{Adaptive \\ Ins\_Length} \\
\hline
Boolq & $62.35$  & $ \textbf{67.28}$ &  $81.20$ & \textbf{83.46} \\
MultiRC   & $\textbf{57.41} $ & $ 57.34$ &  $58.00$ & $ \textbf{66.30}$\\
WiC  & $53.61$ & $ \textbf{60.50}$ & $69.30$  & $ \textbf{71.47}$\\
CB &  $78.57$ & $ \textbf{80.36}$ & $\textbf{87.50} $ & $ 84.32 $\\
RTE   & $67.51$ & $ \textbf{68.32}$ & $\textbf{82.60}$  & $ 79.78 $\\\hline
Avg. & 63.89 & \textbf{66.76} & 75.72 & \textbf{77.07}\\
\hline
%Avg.  & $-$ & $ \textbf{-}$ & $-$  & $ \textbf{-}$\\
%\hline
\end{tabular}
}

\end{minipage}
  \hfill
\begin{minipage}{0.53\linewidth}
\centering
\caption{Few-shot results on T5-LM-Large.
}\label{tab:few-shot}
%\vspace{-0.39cm}
\setlength\tabcolsep{2.3pt}
\scalebox{0.8}{
\begin{tabular}{l@{\hspace{0.8\tabcolsep}}|ccccc}
\hline
\multirow{3}{*}{\textbf{Dataset}} & \multicolumn{5}{c}{\textbf{T5-LM-Large}}  \\
 %\cline{2-7}
 %& \multicolumn{3}{c|}{test}  & \multicolumn{3}{c}{test} \\
  \cline{2-6}
 & \thead{Fixed\\Position} & \thead{Adaptive\\Position}  & \thead{Adaptive \\ Ins\_vec} & \thead{Adaptive \\Ins\_vec\_pos } & \thead{Adaptive \\ Pos\_ins\_vec}\\
\hline
Boolq  & $54.37$  & $60.24$ & $60.18$ & $\textbf{61.07}$   & $60.64$    \\
MultiRC  &$53.49$   & $55.26$ & $54.52$ & $56.13$   & $\textbf{56.79}$    \\
WiC  & $52.66$ & $53.29$ & $53.92$ & $55.33$   & $\textbf{55.33}$    \\
CB   & $76.79$   & $78.57$ & $73.21$ & $80.36$   & $\textbf{87.50}$    \\
RTE  & $51.26$  & $55.96$ & $54.43$ & $54.87$   & $\textbf{57.40}$    \\
\hline
Avg.  & $58.55$ & $60.66$ & $59.05$   & $61.55$ &$\textbf{63.53}$   \\
\hline
\end{tabular}
}

\end{minipage}
\end{table}

To demonstrate the efficiency of our method, in this section, we show that our proposed simple yet powerful approach leads to substantial accuracy gains across various methodologies, including Prompt tuning, P-tuning v2 \citep{Liu2021PTuningVP}, vision prompt tuning (VPT) \citep{jia2022vpt}, and MaPLe \citep{khattakMaPLe} for multi-modal prompt learning, across a wide range of tasks, such as NLP tasks, vision recognition tasks, and vision-language tasks. More experiments, such as the case and ablation study (\ref{analysis}), parameter sensitivity analysis (\ref{sec:apppara}), and additional results (\ref{sec:appadditional}) are included in the Appendix.

\stitle{Adaptive Position.}
As presented in Table \ref{tab:pos}, we compare two variations of adaptive position: $adap\_pos$ represents the dynamically learned position for all instances in a single task, while $adap\_ins\_pos$ indicates that an optimal position is expected to exist for each instance. The experiments are conducted using the T5-LM-Adapt version (Small, Base, Large, and XL), and we report the best results in the table. We can see a general trend that $\textit{adap\_ins\_pos > adap\_pos} $ $\textit{ > fixed\_pos }$ on almost all five datasets\footnote{The T5-Base sometimes demonstrate suboptimal results, as also reported in \citep{Asai2022ATTEMPTPM}}. On average, T5-Large demonstrates substantial improvements of approximately 5 and 7 points compared to the fixed position PT. These improvements are less pronounced for smaller LMs, aligning with the findings of \citep{lester-etal-2021-power} that larger models are better suited for prompt tuning. Considering that the number of additional parameters for \textit{adap\_pos} is merely 20, while several thousand are required for \textit{adap\_ins\_pos}, we can conclude that the fixed position is suboptimal for prompt tuning, and adaptive position consistently provides gains.
\begin{table*} 
\centering
\caption{Three strategies for the dynamic position of soft prompts. Fixed Position is the default prompt tuning. Adaptive Position means the position is learned for every task but fixed for all instances within a task, while Adaptive ins\_position learns a dynamic position for each instance. 
}\label{tab:pos}%\vspace{-0.1cm}
\setlength\tabcolsep{3pt}
\scalebox{0.73}{
\begin{tabular}{l@{\hspace{0.8\tabcolsep}}|ccc|ccc|ccc|ccc}
\hline
\multirow{3}{*}{\textbf{Dataset}}
 & \multicolumn{3}{c|}{\textbf{T5-LM-Small}} & \multicolumn{3}{c|}{\textbf{T5-LM-Base}} & \multicolumn{3}{c|}{\textbf{T5-LM-Large}} & \multicolumn{3}{c}{\textbf{T5-LM-XL}}   \\
  \cline{2-13}
 &  \thead{Fixed\\Position}  & \thead{Adaptive \\ Position} & \thead{Adaptive \\ Ins\_Position}  &  \thead{Fixed\\Position}  & \thead{Adaptive \\ Position} & \thead{Adaptive \\ Ins\_Position} &  \thead{Fixed\\Position}  & \thead{Adaptive \\ Position} & \thead{Adaptive \\ Ins\_Position} &  \thead{Fixed\\Position}  & \thead{Adaptive \\ Position} & \thead{Adaptive \\ Ins\_Position}\\
\hline
Boolq & $67.31$ & $67.55$ & $\textbf{67.61}$      & $62.35$ & $\textbf{69.88}$ & $69.17$       & $81.20$ & $84.60$ & $ \textbf{85.35}$ & $89.02$ & $88.89$ & $\textbf{89.16}$ \\
MultiRC & $68.68$ & $68.89$ & $\textbf{69.29}$    & $57.42$ & $70.19$ & $\textbf{71.08}$       &$58.00^*$ & $72.77$ & $ \textbf{80.20}$ & $\textbf{84.49}$ & $84.31$ & $84.41$ \\
WiC & $62.69$ & $66.14$ & $\textbf{68.34}$        & $53.61$ & $64.42$ & $\textbf{64.89}$       & $69.30$ & $\textbf{71.20}$ & $ \textbf{71.20}$ & $\textbf{72.57}$ & $71.22$ & $70.91$ \\
CB & $\textbf{83.93}$ & $\textbf{83.93}$ & $\textbf{83.93}$         & $78.57$ & $\textbf{87.50}$ & $\textbf{87.50}$       & $87.50$ & $89.29$ & $ \textbf{91.07}$ & $94.64$ & $\textbf{98.21}$ & $ 96.43 $ \\
RTE & $65.34$ & $\textbf{66.79}$ & $65.70$        & $67.51$ & $70.75$ & $\textbf{71.93}$       & $82.60$ & $\textbf{85.71}$ & $ \textbf{85.71}$ & $88.21$ & $\textbf{90.94}$ & $90.58$ \\
\hline
Avg. & $69.59$ & $70.66$ & $\textbf{70.97}$       & $63.89$ & $72.55$ & $\textbf{72.91}$       & $75.72$ & $80.71$ & $ \textbf{82.71}$ & $85.79$ & $\textbf{86.72}$& $86.30$\\
\hline
\end{tabular}
}
\end{table*}

\stitle{Adaptive Length.}
As mentioned in Sec. \ref{dy_len}, we only use a surrogate strategy for length adjustment. Table \ref{tab:len} shows the adaptive length results. For simplicity, we only test $adap\_length$ on T5-base and $adap\_ins\_length$ on T5-large. Overall, adjusting length on task or instance-level helps, compared with fixed prompt length. However, compared with the adaptive position strategy in Table \ref{tab:pos}, the performance gain is lower, which might be caused by the difficulty of tuning. Thus, we recommend this strategy for quickly locating the proper length for different models instead of a greedy search. %We will explore how to efficiently implement adaptive length in the future.

\begin{wrapfigure}{r}{0.55\textwidth}\vspace{-0.48cm}
\centering
    \includegraphics[width=0.54\textwidth]{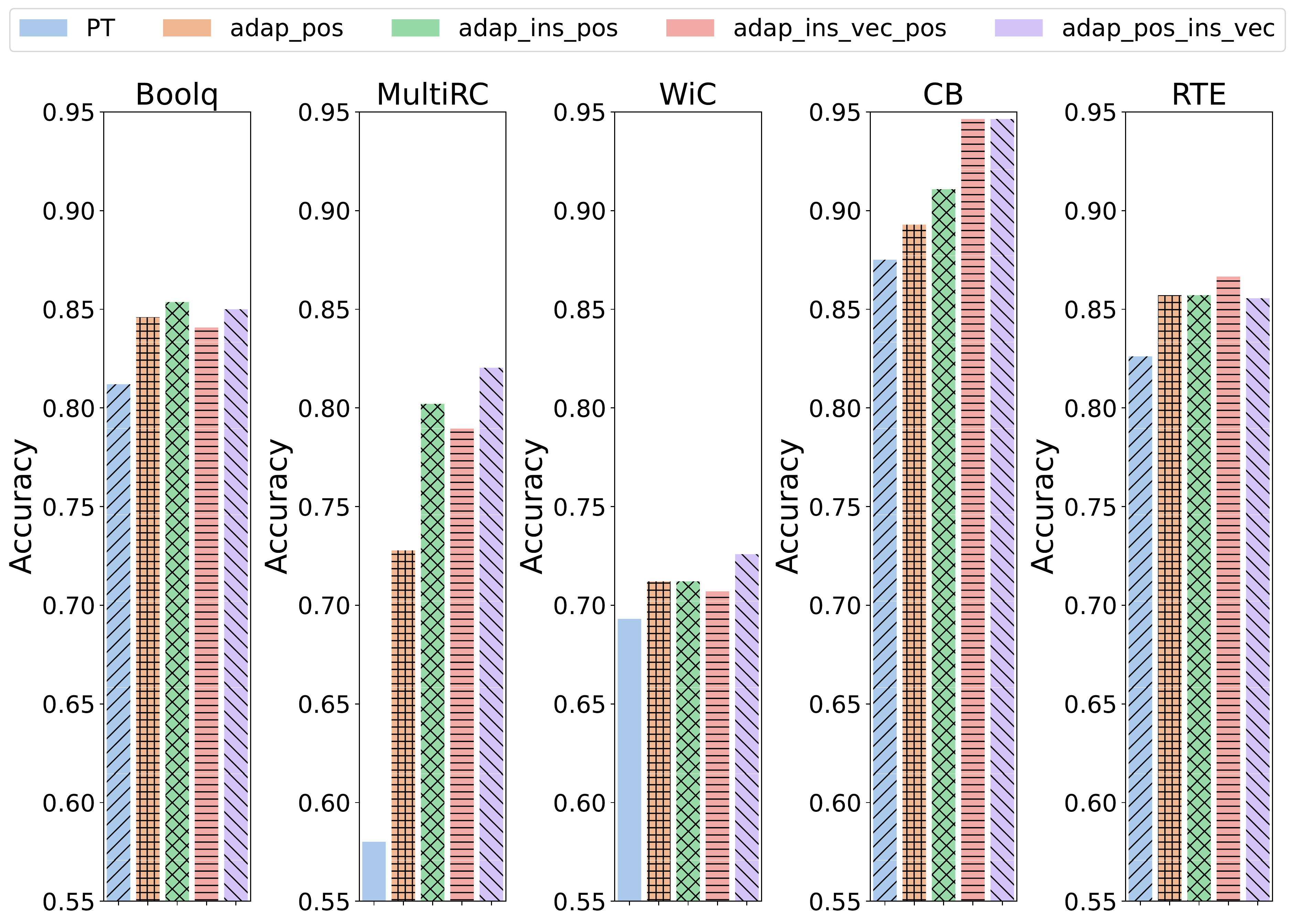} 
    \captionof{figure}{Results on SuperGLUE with T5-Large.  }\label{fig:adap_vec}\vspace{-0.1cm}
\end{wrapfigure}

\stitle{Adaptive Prompt.} \label{5:adap_vec}By adaptively adjusting the synthesized prompt from the prompt pool, the soft prompts are expected to more efficiently utilize the frozen LMs. The results are reported in Table \ref{tab:vec}, and we also give the histogram comparison in Figure \ref{fig:adap_vec}. In general, compared with adaptive position only, adding an adaptive prompt vector increases the performance. But when both position and 
prompts are optimized for each instance, we see slightly lower results in most cases, which might be caused by the increasing difficulty of optimization. 
We leave the work of better optimization methods for future work. Additional results are included in Appendix \ref{app:adap_vec}.

\stitle{Few-Shot.}
We illustrate the few-shot results in Table \ref{tab:few-shot} on T5-Large. As we can see from all datasets, our dynamic prompting consistently improves the results given only 32-shot training examples, demonstrating the broad generalization ability under the low-resource regime.

\stitle{Multi-Task.}
To furthermore demonstrate that multi-task tuning could benefit across tasks for learning a better shared prompt pool, we also show multi-task results in Table \ref{tab:multi}. Here we randomly sample 10\% or 30\% samples from all five datasets and report the average performance. The results confirm that sharing a prompt pool across multiple tasks brings universal benefits.

\begin{figure}[!t]\vspace{-0.3cm}
    \centering
    \subfigure[BERT-Large]{\includegraphics[width=1.9in]{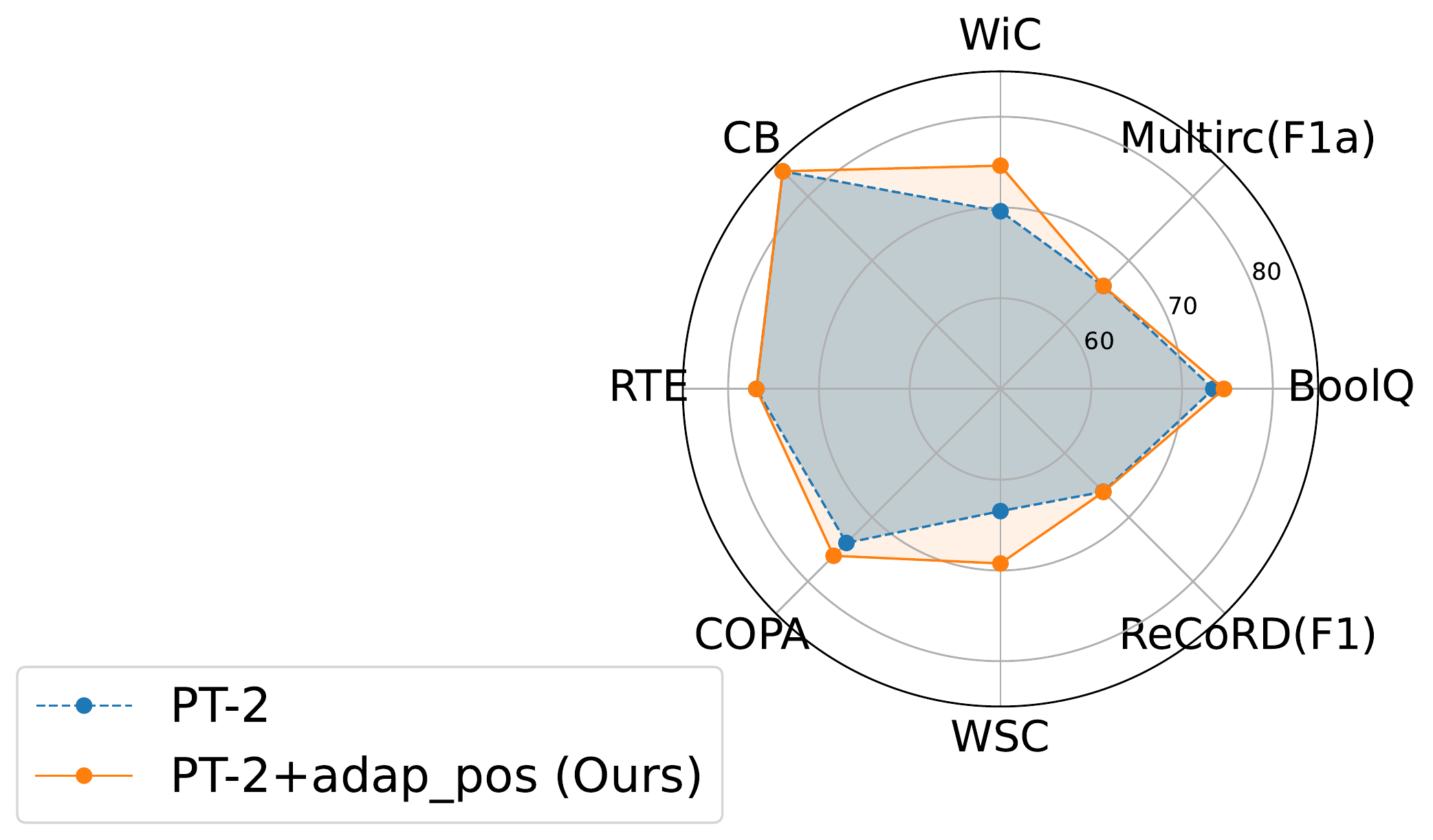}\label{fig:bert_radar}}
     \hspace{1cm}
    \subfigure[RoBERTa-Large]{\includegraphics[width=1.9in]{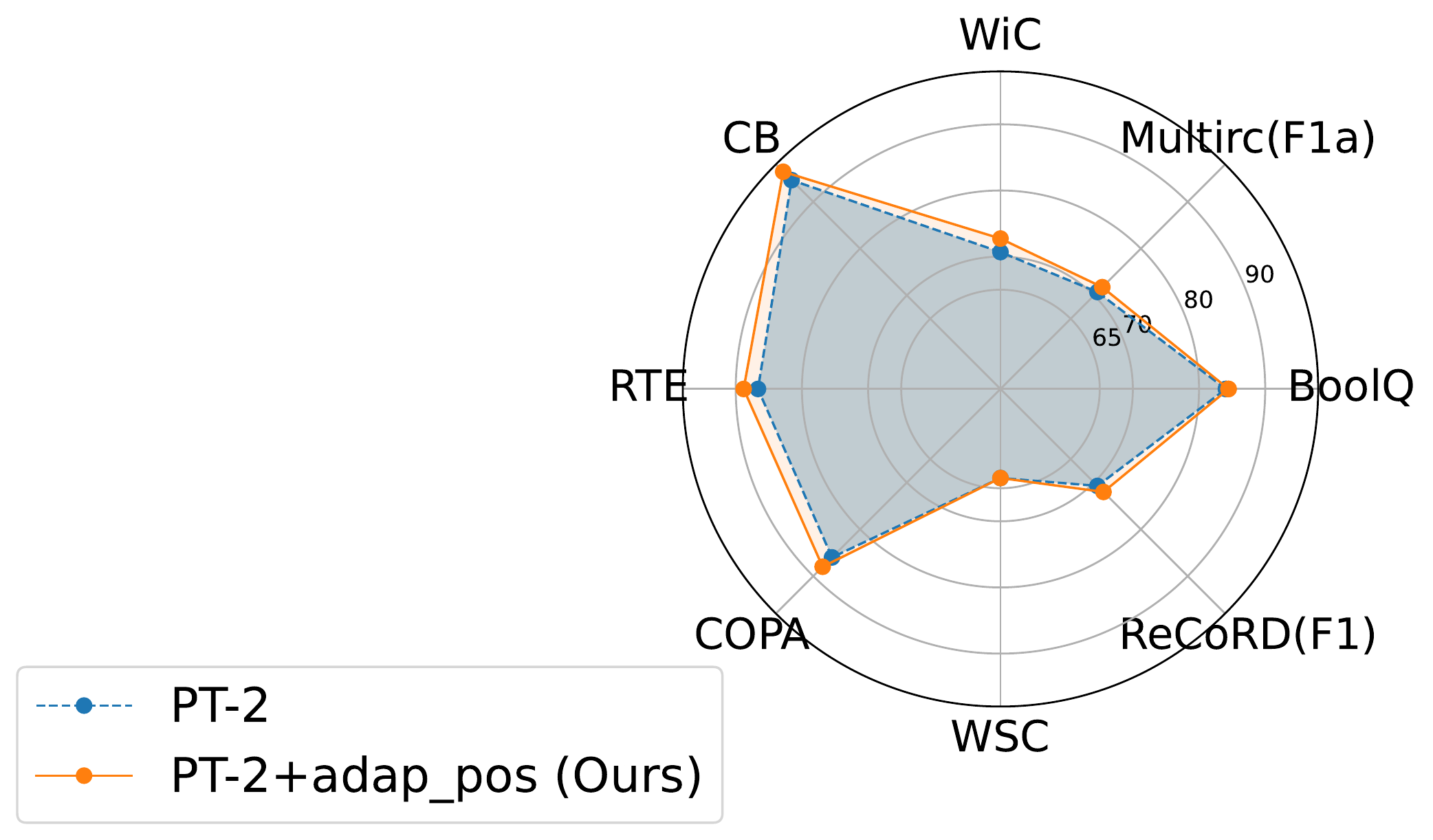}\label{fig:roberta_radar}}
    \caption{Performance comparison of PT-2 and PT-2 with adaptive position (ours) on SuperGLUE with different PLMs. }
    \label{fig:superglue_pt2}\vspace{-0.2cm}
\end{figure}

\stitle{P-tuning V2.} In the pursuit of pushing the boundaries of prompt tuning, \citep{Liu2021PTuningVP} delved deeper into the inner workings of language models (LMs) by attaching a prompt to each transformer layer, transcending the conventional practice of simply affixing a soft prompt to the original input sequence. Remarkably, their efforts yielded comparable results to fine-tuning, highlighting the remarkable potential of prompt tuning. We thus embarked on our investigation to explore the full potential of dynamic prompting. We sought to ascertain the extent to which our dynamic manipulation of soft prompts, reaching deeper into the LM, could enhance accuracy. To ensure a fair comparison, we adopted the identical setup employed by \citep{Liu2021PTuningVP}, employing the backbone models BERT-Large\citep{devlin2018bert} and RoBERTa-Large\citep{liu2019roberta} on the SuperGlue datasets. The results, as depicted in Figure \ref{fig:superglue_pt2}, demonstrate the efficacy of our adaptive prompt position approach. Across most datasets, we observe performance improvements. Our technique achieves an impressive absolute average gain of 1.74\% on BERT-Large and 1.34\% on the RoBERTa-Large over P-tuning V2. Theoretically, as expounded upon in Sec. \ref{unified}, our technique enables soft prompts to encompass the input, capturing additional semantic information that traditional prefix or postfix prompt tuning methods fail to capture. The effectiveness of our adaptive position becomes increasingly apparent as we manipulate prompts across more transformer layers. More results are included in Appendix \ref{sec:apppt2}.
\begin{table*}[!t]
\centering

\begin{minipage}{0.4\linewidth}
\centering
\caption{Multi-task results comparing prompt tuning (PT) and \textit{adap\_ins\_vec} on T5-Large under few-shot setting. 
}\label{tab:multi}
\setlength\tabcolsep{2.5pt}
\scalebox{0.86}{
\begin{tabular}{l@{\hspace{0.8\tabcolsep}}|cc}
\hline
\multirow{2}{*}{\textbf{Methods}}
 & \multicolumn{2}{c}{\textbf{k=8}}  \\
  \cline{2-3}
 &  10\%-shot & 30\%-shot \\
\hline
PT & $67.15$  & $69.79$   \\
\textit{adap\_ins\_vec}  & $\textbf{68.59}$ & $\textbf{72.56}$ \\
\hline
\end{tabular}
}

\end{minipage}
 \hfill
\begin{minipage}{0.57\linewidth}
\centering
\caption{Comparison of MaPLe with and without dynamic prompt position averaged over 11 datasets.
}\label{tab:vsionlanguageMain}

\setlength\tabcolsep{2.1pt}
\scalebox{0.89}{
\begin{tabular}{
>{\columncolor[HTML]{FFFFFF}}c 
>{\columncolor[HTML]{FFFFFF}}c 
>{\columncolor[HTML]{FFFFFF}}c |
>{\columncolor[HTML]{FFFFFF}}c }
\hline
\multicolumn{1}{c}{\cellcolor[HTML]{FFFFFF}Method}                        & Base Acc. & Novel Acc. & \begin{tabular}[c]{@{}c@{}}HM \\ (Base+Novel)\end{tabular} \\ \hline
\multicolumn{1}{c}{\cellcolor[HTML]{FFFFFF}MaPLle}                         & 83.74   & 73.64    & 77.08                                                    \\ \hline
\multicolumn{1}{c}{\cellcolor[HTML]{FFFFFF}MaPLe+\textit{adap\_pos}} & 83.96   & 75.68    & 79.25                                                    \\ %\hline
                                                                      (Ours)      & \textcolor{blue!50}{+0.22}      & \textcolor{blue!50}{+2.04}       & \textcolor{blue!50}{+2.17}                                                        \\ \cline{2-4} 
\end{tabular}
}

\end{minipage}

\vspace{-0.2cm}
\end{table*}

\stitle{Vision Prompt Tuning (VPT).} In the realm of adapting large pre-trained Transformers for downstream vision tasks, a remarkable piece of work known as VPT \citep{jia2022vpt} has emerged. VPT integrated additional parameters into the input sequence of each Transformer layer, simultaneously learned alongside a linear head during the fine-tuning process. 
Our methodology takes a step further by embedding our approach within the model, allowing for adaptive optimization of the prompt position. When applying VPT in a deep setting, we maintain the prompt position of the deep transformer layers consistent with that of the input layer. The results, as depicted in Table \ref{tab:vpt_full}, demonstrate the performance gains achieved by intelligently optimizing the prompt position. The instance-aware prompt position selection further improves accuracy. Remarkably, these benefits manifest across both shallow and deep settings of VPT, underscoring the robustness and efficacy of our approach. More results are included in Appendix \ref{sec:vpt_app}.

\stitle{Vision-Language Model.}
The Vision-language (V-L) model, such as the remarkable CLIP \citep{radford2021learning}, has garnered widespread acclaim for its exceptional ability to align language and vision modalities. The pioneering work of MaPLe \citep{khattakMaPLe} introduced a coupling function to effectively condition vision prompts based on their language counterparts, bridging the gap between the two modalities. Inspired by these advancements, we incorporate our adaptive prompt position approach into the text input layer, leveraging the power of dynamic prompt manipulation. We outline the detailed experimental settings in the Appendix. The compelling results,  as summarized in Table \ref{tab:vsionlanguageMain}, substantiate the potency of our approach. Remarkably, by incorporating adaptive prompt position into MaPLe, we achieve an impressive absolute average gain of 2.04\% on novel classes and 2.17\% on the harmonic mean, surpassing the state-of-the-art method MaPLe \citep{khattakMaPLe}. This performance improvement serves as a compelling testament to the effectiveness of our dynamic prompting methodology, firmly establishing its efficacy in the realm of V-L models. More detailed results are included in Appendix \ref{sec:vision_language_app}.
\begin{table}[htbp] 
\caption{Comparison between basic VPT model and VPT with the adaptive position.
}\label{tab:vpt_full}
\setlength\tabcolsep{2.5pt}
\scalebox{1.}{
\scriptsize{
\begin{tabular}{
>{\columncolor[HTML]{FFFFFF}}c |
>{\columncolor[HTML]{FFFFFF}}c |
>{\columncolor[HTML]{FFFFFF}}c |
>{\columncolor[HTML]{FFFFFF}}c |
>{\columncolor[HTML]{FFFFFF}}c |
>{\columncolor[HTML]{FFFFFF}}c |
>{\columncolor[HTML]{FFFFFF}}c |
>{\columncolor[HTML]{FFFFFF}}c }
\hline
                                                      & Dataset                                                                                     & CUB-200-2011                          & NABirds                               & Oxford Flowers                        & Stanford Dogs                         & Stanford Cars                         & Avg.                         \\ \hline
\cellcolor[HTML]{FFFFFF}                              & {\color[HTML]{000000} prompt length}                                                        & {\color[HTML]{000000} 100}            & {\color[HTML]{000000} 50}             & {\color[HTML]{000000} 100}            & {\color[HTML]{000000} 100}            & {\color[HTML]{000000} 100}            & {\color[HTML]{000000} }      \\ \arrayrulecolor{black}\cline{2-8} 
\cellcolor[HTML]{FFFFFF}                              & {\color[HTML]{000000} VPT}                                                                  & {\color[HTML]{000000} 85.42}          & {\color[HTML]{000000} 75.11}          & {\color[HTML]{000000} 98.29}          & {\color[HTML]{000000} 90.42}          & {\color[HTML]{000000} 54.60}          & {\color[HTML]{000000} 80.77} \\ \arrayrulecolor{black}\cline{2-8} 
\cellcolor[HTML]{FFFFFF}                              & \cellcolor[HTML]{FFFFFF}{\color[HTML]{000000} }                                             & {\color[HTML]{000000} 86.26}          & {\color[HTML]{000000} 75.56}          & {\color[HTML]{000000} 98.44}          & {\color[HTML]{000000} 91.27}          & {\color[HTML]{000000} 57.02}          & {\color[HTML]{000000} 81.71} \\ %\cline{3-8} 
\cellcolor[HTML]{FFFFFF}                              & \multirow{-2}{*}{\cellcolor[HTML]{FFFFFF}{\color[HTML]{000000} VPT+\textit{adap\_pos}  (Ours)}}      & {\color[HTML]{34CDF9} +0.84}          & {\color[HTML]{34CDF9} +0.45}          & {\color[HTML]{34CDF9} +0.15}          & {\color[HTML]{34CDF9} +0.85}          & {\color[HTML]{34CDF9} +0.42}          & {\color[HTML]{34CDF9} +0.94} \\ \arrayrulecolor{black}\cline{2-8} 
\cellcolor[HTML]{FFFFFF}                              & \cellcolor[HTML]{FFFFFF}{\color[HTML]{000000} }                                             & {\color[HTML]{000000} 86.31}          & {\color[HTML]{000000} 76.63}          & {\color[HTML]{000000} 98.52}          & {\color[HTML]{000000} 91.39}          & {\color[HTML]{000000} 58.13}          & {\color[HTML]{000000} 82.20} \\ %\cline{3-8} 
\multirow{-6}{*}{\cellcolor[HTML]{FFFFFF}VPT-Shallow} & \multirow{-2}{*}{\cellcolor[HTML]{FFFFFF}{\color[HTML]{000000} VPT+\textit{adap\_ins\_pos}  (Ours)}} & {\color[HTML]{3166FF} +0.89}          & {\color[HTML]{3166FF} +1.52}          & {\color[HTML]{3166FF} +0.23}          & {\color[HTML]{3166FF} +0.97}          & {\color[HTML]{3166FF} +3.53}          & {\color[HTML]{3166FF} +1.43} \\ \hline
\cellcolor[HTML]{FFFFFF}                              & {\color[HTML]{000000} prompt length}                                                        & {\color[HTML]{000000} 10}             & {\color[HTML]{000000} 50}             & {\color[HTML]{000000} 5}              & {\color[HTML]{000000} 5}              & {\color[HTML]{000000} 100}            & {\color[HTML]{000000} }      \\ \arrayrulecolor{black}\cline{2-8} 
\cellcolor[HTML]{FFFFFF}                              & {\color[HTML]{000000} VPT}                                                                  & {\color[HTML]{000000} 87.81}          & {\color[HTML]{000000} 81.43}          & {\color[HTML]{000000} 98.91}          & {\color[HTML]{000000} 90.57}          & {\color[HTML]{000000} 82.99}          & {\color[HTML]{000000} 88.34} \\ \arrayrulecolor{black}\cline{2-8} 
\cellcolor[HTML]{FFFFFF}                              & \cellcolor[HTML]{FFFFFF}{\color[HTML]{000000} }                                             & {\color[HTML]{000000} 88.06}          & {\color[HTML]{000000} 82.98}          & {\color[HTML]{000000} 98.99}          & {\color[HTML]{000000} 91.27}          & {\color[HTML]{000000} 83.26}          & {\color[HTML]{000000} 88.91} \\ %\cline{3-8} 
\cellcolor[HTML]{FFFFFF}                              & \multirow{-2}{*}{\cellcolor[HTML]{FFFFFF}{\color[HTML]{000000} VPT+\textit{adap\_pos}  (Ours)}}      & {\color[HTML]{34CDF9} +0.25}          & {\color[HTML]{34CDF9} +1.55}          & {\color[HTML]{34CDF9} +0.08}          & {\color[HTML]{34CDF9} +0.70}          & {\color[HTML]{34CDF9} +0.27}          & {\color[HTML]{34CDF9} +0.57} \\ \arrayrulecolor{black}\cline{2-8} 
\cellcolor[HTML]{FFFFFF}                              & \cellcolor[HTML]{FFFFFF}{\color[HTML]{000000} }                                             & {\color[HTML]{000000} \textbf{88.15}} & {\color[HTML]{000000} \textbf{83.02}} & {\color[HTML]{000000} \textbf{99.01}} & {\color[HTML]{000000} \textbf{91.32}} & {\color[HTML]{000000} \textbf{83.42}} & {\color[HTML]{000000} 88.98} \\ %\cline{3-8} 
\multirow{-6}{*}{\cellcolor[HTML]{FFFFFF}VPT-Deep}    & \multirow{-2}{*}{\cellcolor[HTML]{FFFFFF}{\color[HTML]{000000} VPT+\textit{adap\_ins\_pos}  (Ours)}} & {\color[HTML]{3166FF} +0.34}          & {\color[HTML]{3166FF} +1.59}          & {\color[HTML]{3166FF} +0.10}          & {\color[HTML]{3166FF} +0.75}          & {\color[HTML]{3166FF} +0.43}           & {\color[HTML]{3166FF} +0.64} \\ \hline
\end{tabular}
}
}
\end{table}

\section{Conclusion}
In this work, we first derive a unified view of prompt tuning, then present a novel approach to parameter-efficient prompt tuning called dynamic prompting that can significantly improve the performance of prompt tuning while adding only a few additional parameters. The key contributions of this work include exploring the effectiveness of the dynamic position, length, and prompt representation in improving traditional prompt tuning and systematically exploring dynamic prompting under the combination of different dynamic methodologies in various scenarios. Comprehensive experiments on a broad spectrum of datasets validate that dynamic prompting consistently achieves superior results across a diverse range of tasks, including language understanding tasks, vision recognition tasks, and vision-language tasks, regardless of the model sizes being employed. We also demonstrate that dynamic prompting is effective in multi-task and few-shot settings. Overall, our work can further unleash the power of prompt tuning across various modalities. We hope our method can help with the efficient use of large pretrained models and additionally close the gap between fine-tuning.

\clearpage
% Entries for the entire Anthology, followed by custom entries

\bibliographystyle{plain}
\bibliography{anthology,custom}

\begin{thebibliography}{10}

\bibitem{aribandi2021ext5}
Vamsi Aribandi, Yi~Tay, Tal Schuster, Jinfeng Rao, Huaixiu~Steven Zheng,
  Sanket~Vaibhav Mehta, Honglei Zhuang, Vinh~Q Tran, Dara Bahri, Jianmo Ni,
  et~al.
\newblock Ext5: Towards extreme multi-task scaling for transfer learning.
\newblock {\em arXiv preprint arXiv:2111.10952}, 2021.

\bibitem{Asai2022ATTEMPTPM}
Akari Asai, Mohammadreza Salehi, Matthew~E. Peters, and Hannaneh Hajishirzi.
\newblock Attempt: Parameter-efficient multi-task tuning via attentional
  mixtures of soft prompts.
\newblock In {\em Conference on Empirical Methods in Natural Language
  Processing}, 2022.

\bibitem{Asai2022AttentionalMO}
Akari Asai, Mohammadreza Salehi, Matthew~E. Peters, and Hannaneh Hajishirzi.
\newblock Attentional mixtures of soft prompt tuning for parameter-efficient
  multi-task knowledge sharing.
\newblock {\em ArXiv}, abs/2205.11961, 2022.

\bibitem{visprompt}
Amir Bar, Yossi Gandelsman, Trevor Darrell, Amir Globerson, and Alexei~A.
  Efros.
\newblock Visual prompting via image inpainting.
\newblock {\em Advances in Neural Information Processing Systems (NeurIPS)},
  2022.

\bibitem{ben-zaken-etal-2022-bitfit}
Elad Ben~Zaken, Yoav Goldberg, and Shauli Ravfogel.
\newblock {B}it{F}it: Simple parameter-efficient fine-tuning for
  transformer-based masked language-models.
\newblock In {\em Proceedings of the 60th Annual Meeting of the Association for
  Computational Linguistics (Volume 2: Short Papers)}, pages 1--9, Dublin,
  Ireland, May 2022. Association for Computational Linguistics.

\bibitem{bengio2013estimating}
Yoshua Bengio, Nicholas L{\'e}onard, and Aaron Courville.
\newblock Estimating or propagating gradients through stochastic neurons for
  conditional computation.
\newblock {\em arXiv preprint arXiv:1308.3432}, 2013.

\bibitem{brown2020language}
Tom Brown, Benjamin Mann, Nick Ryder, Melanie Subbiah, Jared~D Kaplan, Prafulla
  Dhariwal, Arvind Neelakantan, Pranav Shyam, Girish Sastry, Amanda Askell,
  et~al.
\newblock Language models are few-shot learners.
\newblock {\em Advances in neural information processing systems},
  33:1877--1901, 2020.

\bibitem{Chen2022RevisitingPT}
Guanzheng Chen, Fangyu Liu, Zaiqiao Meng, and Shangsong Liang.
\newblock Revisiting parameter-efficient tuning: Are we really there yet?
\newblock In {\em Conference on Empirical Methods in Natural Language
  Processing}, 2022.

\bibitem{chen2022adaptformer}
Shoufa Chen, Chongjian Ge, Zhan Tong, Jiangliu Wang, Yibing Song, Jue Wang, and
  Ping Luo.
\newblock Adaptformer: Adapting vision transformers for scalable visual
  recognition.
\newblock {\em 36th Conference on Neural Information Processing Systems
  (NeurIPS)}, 2022.

\bibitem{clark-etal-2019-boolq}
Christopher Clark, Kenton Lee, Ming-Wei Chang, Tom Kwiatkowski, Michael
  Collins, and Kristina Toutanova.
\newblock {B}ool{Q}: Exploring the surprising difficulty of natural yes/no
  questions.
\newblock In {\em Proceedings of the 2019 Conference of the North {A}merican
  Chapter of the Association for Computational Linguistics: Human Language
  Technologies, Volume 1 (Long and Short Papers)}, pages 2924--2936,
  Minneapolis, Minnesota, June 2019. Association for Computational Linguistics.

\bibitem{Dai2022WhyCG}
Damai Dai, Yutao Sun, Li~Dong, Yaru Hao, Zhifang Sui, and Furu Wei.
\newblock Why can gpt learn in-context? language models secretly perform
  gradient descent as meta-optimizers.
\newblock {\em ArXiv}, abs/2212.10559, 2022.

\bibitem{de2019commitmentbank}
Marie-Catherine De~Marneffe, Mandy Simons, and Judith Tonhauser.
\newblock The commitmentbank: Investigating projection in naturally occurring
  discourse.
\newblock In {\em proceedings of Sinn und Bedeutung}, volume~23, pages
  107--124, 2019.

\bibitem{devlin2018bert}
Jacob Devlin, Ming-Wei Chang, Kenton Lee, and Kristina Toutanova.
\newblock Bert: Pre-training of deep bidirectional transformers for language
  understanding.
\newblock {\em arXiv preprint arXiv:1810.04805}, 2018.

\bibitem{devlin-etal-2019-bert}
Jacob Devlin, Ming-Wei Chang, Kenton Lee, and Kristina Toutanova.
\newblock {BERT}: Pre-training of deep bidirectional transformers for language
  understanding.
\newblock In {\em Proceedings of the 2019 Conference of the North {A}merican
  Chapter of the Association for Computational Linguistics: Human Language
  Technologies, Volume 1 (Long and Short Papers)}, pages 4171--4186,
  Minneapolis, Minnesota, June 2019. Association for Computational Linguistics.

\bibitem{ding-etal-2022-openprompt}
Ning Ding, Shengding Hu, Weilin Zhao, Yulin Chen, Zhiyuan Liu, Haitao Zheng,
  and Maosong Sun.
\newblock {O}pen{P}rompt: An open-source framework for prompt-learning.
\newblock In {\em Proceedings of the 60th Annual Meeting of the Association for
  Computational Linguistics: System Demonstrations}, pages 105--113, Dublin,
  Ireland, May 2022. Association for Computational Linguistics.

\bibitem{giampiccolo-etal-2007-third}
Danilo Giampiccolo, Bernardo Magnini, Ido Dagan, and Bill Dolan.
\newblock The third {PASCAL} recognizing textual entailment challenge.
\newblock In {\em Proceedings of the {ACL}-{PASCAL} Workshop on Textual
  Entailment and Paraphrasing}, pages 1--9, Prague, June 2007. Association for
  Computational Linguistics.

\bibitem{gu-etal-2022-ppt}
Yuxian Gu, Xu~Han, Zhiyuan Liu, and Minlie Huang.
\newblock {PPT}: Pre-trained prompt tuning for few-shot learning.
\newblock In {\em Proceedings of the 60th Annual Meeting of the Association for
  Computational Linguistics (Volume 1: Long Papers)}, pages 8410--8423, Dublin,
  Ireland, May 2022. Association for Computational Linguistics.

\bibitem{Guo2022ImprovingTS}
Xu~Guo, Boyang~Albert Li, and Han Yu.
\newblock Improving the sample efficiency of prompt tuning with domain
  adaptation.
\newblock In {\em Conference on Empirical Methods in Natural Language
  Processing}, 2022.

\bibitem{he2021towards}
Junxian He, Chunting Zhou, Xuezhe Ma, Taylor Berg-Kirkpatrick, and Graham
  Neubig.
\newblock Towards a unified view of parameter-efficient transfer learning.
\newblock {\em arXiv preprint arXiv:2110.04366}, 2021.

\bibitem{houlsby2019parameter}
Neil Houlsby, Andrei Giurgiu, Stanislaw Jastrzebski, Bruna Morrone, Quentin
  De~Laroussilhe, Andrea Gesmundo, Mona Attariyan, and Sylvain Gelly.
\newblock Parameter-efficient transfer learning for nlp.
\newblock In {\em International Conference on Machine Learning}, pages
  2790--2799. PMLR, 2019.

\bibitem{Hu2021LoRALA}
Edward~J. Hu, Yelong Shen, Phillip Wallis, Zeyuan Allen-Zhu, Yuanzhi Li, Shean
  Wang, and Weizhu Chen.
\newblock Lora: Low-rank adaptation of large language models.
\newblock {\em ArXiv}, abs/2106.09685, 2021.

\bibitem{DBLP:conf/iclr/JangGP17}
Eric Jang, Shixiang Gu, and Ben Poole.
\newblock Categorical reparameterization with gumbel-softmax.
\newblock In {\em 5th International Conference on Learning Representations,
  {ICLR} 2017, Toulon, France, April 24-26, 2017, Conference Track
  Proceedings}. OpenReview.net, 2017.

\bibitem{jang2017categorical}
Eric Jang, Shixiang Gu, and Ben Poole.
\newblock Categorical reparameterization with gumbel-softmax.
\newblock In {\em International Conference on Learning Representations}, 2017.

\bibitem{jia2022vpt}
Menglin Jia, Luming Tang, Bor-Chun Chen, Claire Cardie, Serge Belongie, Bharath
  Hariharan, and Ser-Nam Lim.
\newblock Visual prompt tuning.
\newblock In {\em European Conference on Computer Vision (ECCV)}, 2022.

\bibitem{Jia2022VisualPT}
Menglin Jia, Luming Tang, Bor-Chun Chen, Claire Cardie, Serge~J. Belongie,
  Bharath Hariharan, and Ser~Nam Lim.
\newblock Visual prompt tuning.
\newblock {\em ArXiv}, abs/2203.12119, 2022.

\bibitem{jin2022good}
Woojeong Jin, Yu~Cheng, Yelong Shen, Weizhu Chen, and Xiang Ren.
\newblock A good prompt is worth millions of parameters: Low-resource
  prompt-based learning for vision-language models.
\newblock In {\em Association for Computational Linguistics (ACL)}, 2022.

\bibitem{karimi2021compacter}
Rabeeh Karimi~Mahabadi, James Henderson, and Sebastian Ruder.
\newblock Compacter: Efficient low-rank hypercomplex adapter layers.
\newblock {\em Advances in Neural Information Processing Systems},
  34:1022--1035, 2021.

\bibitem{khashabi-etal-2018-looking}
Daniel Khashabi, Snigdha Chaturvedi, Michael Roth, Shyam Upadhyay, and Dan
  Roth.
\newblock Looking beyond the surface: A challenge set for reading comprehension
  over multiple sentences.
\newblock In {\em Proceedings of the 2018 Conference of the North {A}merican
  Chapter of the Association for Computational Linguistics: Human Language
  Technologies, Volume 1 (Long Papers)}, pages 252--262, New Orleans,
  Louisiana, June 2018. Association for Computational Linguistics.

\bibitem{khattakMaPLe}
Muhammad~Uzair khattak, Hanoona Rasheed, Muhammad Maaz, Salman Khan, and
  Fahad~Shahbaz Khan.
\newblock Maple: Multi-modal prompt learning.
\newblock In {\em The IEEE/CVF Conference on Computer Vision and Pattern
  Recognition}, 2023.

\bibitem{lester-etal-2021-power}
Brian Lester, Rami Al-Rfou, and Noah Constant.
\newblock The power of scale for parameter-efficient prompt tuning.
\newblock In {\em Proceedings of the 2021 Conference on Empirical Methods in
  Natural Language Processing}, pages 3045--3059, Online and Punta Cana,
  Dominican Republic, November 2021. Association for Computational Linguistics.

\bibitem{Lewis2019BARTDS}
Mike Lewis, Yinhan Liu, Naman Goyal, Marjan Ghazvininejad, Abdelrahman Mohamed,
  Omer Levy, Veselin Stoyanov, and Luke Zettlemoyer.
\newblock Bart: Denoising sequence-to-sequence pre-training for natural
  language generation, translation, and comprehension.
\newblock In {\em Annual Meeting of the Association for Computational
  Linguistics}, 2019.

\bibitem{li-liang-2021-prefix}
Xiang~Lisa Li and Percy Liang.
\newblock Prefix-tuning: Optimizing continuous prompts for generation.
\newblock In {\em Proceedings of the 59th Annual Meeting of the Association for
  Computational Linguistics and the 11th International Joint Conference on
  Natural Language Processing (Volume 1: Long Papers)}, pages 4582--4597,
  Online, August 2021. Association for Computational Linguistics.

\bibitem{Lian_2022_SSF}
Dongze Lian, Daquan Zhou, Jiashi Feng, and Xinchao Wang.
\newblock Scaling \& shifting your features: A new baseline for efficient model
  tuning.
\newblock In {\em Advances in Neural Information Processing Systems (NeurIPS)},
  2022.

\bibitem{Liu2022FewShotPF}
Haokun Liu, Derek Tam, Mohammed Muqeeth, Jay Mohta, Tenghao Huang, Mohit
  Bansal, and Colin Raffel.
\newblock Few-shot parameter-efficient fine-tuning is better and cheaper than
  in-context learning.
\newblock {\em ArXiv}, abs/2205.05638, 2022.

\bibitem{liu2021makes}
Jiachang Liu, Dinghan Shen, Yizhe Zhang, Bill Dolan, Lawrence Carin, and Weizhu
  Chen.
\newblock What makes good in-context examples for gpt-$3 $?
\newblock {\em arXiv preprint arXiv:2101.06804}, 2021.

\bibitem{liu2023pre}
Pengfei Liu, Weizhe Yuan, Jinlan Fu, Zhengbao Jiang, Hiroaki Hayashi, and
  Graham Neubig.
\newblock Pre-train, prompt, and predict: A systematic survey of prompting
  methods in natural language processing.
\newblock {\em ACM Computing Surveys}, 55(9):1--35, 2023.

\bibitem{Liu2021PTuningVP}
Xiao Liu, Kaixuan Ji, Yicheng Fu, Zhengxiao Du, Zhilin Yang, and Jie Tang.
\newblock P-tuning v2: Prompt tuning can be comparable to fine-tuning
  universally across scales and tasks.
\newblock {\em ArXiv}, abs/2110.07602, 2021.

\bibitem{Liu2019RoBERTaAR}
Yinhan Liu, Myle Ott, Naman Goyal, Jingfei Du, Mandar Joshi, Danqi Chen, Omer
  Levy, Mike Lewis, Luke Zettlemoyer, and Veselin Stoyanov.
\newblock Roberta: A robustly optimized bert pretraining approach.
\newblock {\em ArXiv}, abs/1907.11692, 2019.

\bibitem{liu2019roberta}
Yinhan Liu, Myle Ott, Naman Goyal, Jingfei Du, Mandar Joshi, Danqi Chen, Omer
  Levy, Mike Lewis, Luke Zettlemoyer, and Veselin Stoyanov.
\newblock Roberta: A robustly optimized bert pretraining approach.
\newblock {\em arXiv preprint arXiv:1907.11692}, 2019.

\bibitem{Ma2022XPromptET}
Fang Ma, Chen Zhang, Lei Ren, Jingang Wang, Qifan Wang, Wei~Yu Wu, Xiaojun
  Quan, and Dawei Song.
\newblock Xprompt: Exploring the extreme of prompt tuning.
\newblock In {\em Conference on Empirical Methods in Natural Language
  Processing}, 2022.

\bibitem{maddison2017concrete}
Chris~J Maddison, Andriy Mnih, and Yee~Whye Teh.
\newblock The concrete distribution: A continuous relaxation of discrete random
  variables.
\newblock {\em International Conference on Learning Representations}, 2017.

\bibitem{shu2022tpt}
Shu Manli, Nie Weili, Huang De-An, Yu~Zhiding, Goldstein Tom, Anandkumar Anima,
  and Xiao Chaowei.
\newblock Test-time prompt tuning for zero-shot generalization in
  vision-language models.
\newblock In {\em Advances in Neural Information Processing Systems (NeurIPS)},
  2022.

\bibitem{pennington-etal-2014-glove}
Jeffrey Pennington, Richard Socher, and Christopher Manning.
\newblock {G}lo{V}e: Global vectors for word representation.
\newblock In {\em Proceedings of the 2014 Conference on Empirical Methods in
  Natural Language Processing ({EMNLP})}, pages 1532--1543, Doha, Qatar,
  October 2014. Association for Computational Linguistics.

\bibitem{pilehvar-camacho-collados-2019-wic}
Mohammad~Taher Pilehvar and Jose Camacho-Collados.
\newblock {W}i{C}: the word-in-context dataset for evaluating context-sensitive
  meaning representations.
\newblock In {\em Proceedings of the 2019 Conference of the North {A}merican
  Chapter of the Association for Computational Linguistics: Human Language
  Technologies, Volume 1 (Long and Short Papers)}, pages 1267--1273,
  Minneapolis, Minnesota, June 2019. Association for Computational Linguistics.

\bibitem{radford2021learning}
Alec Radford, Jong~Wook Kim, Chris Hallacy, Aditya Ramesh, Gabriel Goh,
  Sandhini Agarwal, Girish Sastry, Amanda Askell, Pamela Mishkin, Jack Clark,
  Gretchen Krueger, and Ilya Sutskever.
\newblock Learning transferable visual models from natural language
  supervision.
\newblock In {\em International Conference on Machine Learning}, 2021.

\bibitem{raffel2020exploring}
Colin Raffel, Noam Shazeer, Adam Roberts, Katherine Lee, Sharan Narang, Michael
  Matena, Yanqi Zhou, Wei Li, and Peter~J Liu.
\newblock Exploring the limits of transfer learning with a unified text-to-text
  transformer.
\newblock {\em The Journal of Machine Learning Research}, 21(1):5485--5551,
  2020.

\bibitem{shazeer2018adafactor}
Noam Shazeer and Mitchell Stern.
\newblock Adafactor: Adaptive learning rates with sublinear memory cost.
\newblock In {\em International Conference on Machine Learning}, pages
  4596--4604. PMLR, 2018.

\bibitem{su-etal-2022-transferability}
Yusheng Su, Xiaozhi Wang, Yujia Qin, Chi-Min Chan, Yankai Lin, Huadong Wang,
  Kaiyue Wen, Zhiyuan Liu, Peng Li, Juanzi Li, Lei Hou, Maosong Sun, and Jie
  Zhou.
\newblock On transferability of prompt tuning for natural language processing.
\newblock In {\em Proceedings of the 2022 Conference of the North American
  Chapter of the Association for Computational Linguistics: Human Language
  Technologies}, pages 3949--3969, Seattle, United States, July 2022.
  Association for Computational Linguistics.

\bibitem{Sun2022MultiTaskPO}
Tianxiang Sun, Zhengfu He, Qinen Zhu, Xipeng Qiu, and Xuanjing Huang.
\newblock Multi-task pre-training of modular prompt for few-shot learning.
\newblock {\em ArXiv}, abs/2210.07565, 2022.

\bibitem{Oswald2022TransformersLI}
Johannes von Oswald, Eyvind Niklasson, E.~Randazzo, Jo{\~a}o Sacramento,
  Alexander Mordvintsev, Andrey Zhmoginov, and Max Vladymyrov.
\newblock Transformers learn in-context by gradient descent.
\newblock {\em ArXiv}, abs/2212.07677, 2022.

\bibitem{vu-etal-2022-spot}
Tu~Vu, Brian Lester, Noah Constant, Rami Al-Rfou{'}, and Daniel Cer.
\newblock {SP}o{T}: Better frozen model adaptation through soft prompt
  transfer.
\newblock In {\em Proceedings of the 60th Annual Meeting of the Association for
  Computational Linguistics (Volume 1: Long Papers)}, pages 5039--5059, Dublin,
  Ireland, May 2022. Association for Computational Linguistics.

\bibitem{wang2019superglue}
Alex Wang, Yada Pruksachatkun, Nikita Nangia, Amanpreet Singh, Julian Michael,
  Felix Hill, Omer Levy, and Samuel Bowman.
\newblock Superglue: A stickier benchmark for general-purpose language
  understanding systems.
\newblock {\em Advances in neural information processing systems}, 32, 2019.

\bibitem{wang2022learning}
Zifeng Wang, Zizhao Zhang, Chen-Yu Lee, Han Zhang, Ruoxi Sun, Xiaoqi Ren,
  Guolong Su, Vincent Perot, Jennifer Dy, and Tomas Pfister.
\newblock Learning to prompt for continual learning.
\newblock In {\em Proceedings of the IEEE/CVF Conference on Computer Vision and
  Pattern Recognition}, pages 139--149, 2022.

\bibitem{wei2021pretrained}
Colin Wei, Sang~Michael Xie, and Tengyu Ma.
\newblock Why do pretrained language models help in downstream tasks? an
  analysis of head and prompt tuning.
\newblock {\em Advances in Neural Information Processing Systems},
  34:16158--16170, 2021.

\bibitem{wu2022idpg}
Zhuofeng Wu, Sinong Wang, Jiatao Gu, Rui Hou, Yuxiao Dong, VG~Vydiswaran, and
  Hao Ma.
\newblock Idpg: An instance-dependent prompt generation method.
\newblock {\em arXiv preprint arXiv:2204.04497}, 2022.

\bibitem{wu-etal-2022-idpg}
Zhuofeng Wu, Sinong Wang, Jiatao Gu, Rui Hou, Yuxiao Dong, V.G.Vinod
  Vydiswaran, and Hao Ma.
\newblock {IDPG}: An instance-dependent prompt generation method.
\newblock In {\em Proceedings of the 2022 Conference of the North American
  Chapter of the Association for Computational Linguistics: Human Language
  Technologies}, pages 5507--5521, Seattle, United States, July 2022.
  Association for Computational Linguistics.

\bibitem{zhang2021beyond}
Aston Zhang, Yi~Tay, Shuai Zhang, Alvin Chan, Anh~Tuan Luu, Siu~Cheung Hui, and
  Jie Fu.
\newblock Beyond fully-connected layers with quaternions: Parameterization of
  hypercomplex multiplications with $1/n $ parameters.
\newblock {\em arXiv preprint arXiv:2102.08597}, 2021.

\bibitem{zhang2022automatic}
Zhuosheng Zhang, Aston Zhang, Mu~Li, and Alex Smola.
\newblock Automatic chain of thought prompting in large language models.
\newblock {\em arXiv preprint arXiv:2210.03493}, 2022.

\bibitem{zhou2022cocoop}
Kaiyang Zhou, Jingkang Yang, Chen~Change Loy, and Ziwei Liu.
\newblock Conditional prompt learning for vision-language models.
\newblock In {\em IEEE/CVF Conference on Computer Vision and Pattern
  Recognition (CVPR)}, 2022.

\bibitem{zhou2022coop}
Kaiyang Zhou, Jingkang Yang, Chen~Change Loy, and Ziwei Liu.
\newblock Learning to prompt for vision-language models.
\newblock {\em International Journal of Computer Vision (IJCV)}, 2022.

\end{thebibliography}


\begin{thebibliography}{10}

\bibitem{badaskar2008identifying}
Sameer Badaskar, Sachin Agarwal, and Shilpa Arora.
\newblock Identifying real or fake articles: Towards better language modeling.
\newblock In {\em Proceedings of the Third International Joint Conference on
  Natural Language Processing: Volume-II}, 2008.

\bibitem{black2022gpt}
Sid Black, Stella Biderman, Eric Hallahan, Quentin Anthony, Leo Gao, Laurence
  Golding, Horace He, Connor Leahy, Kyle McDonell, Jason Phang, et~al.
\newblock Gpt-neox-20b: An open-source autoregressive language model.
\newblock {\em Challenges \& Perspectives in Creating Large Language Models},
  page~95, 2022.

\bibitem{bojar2016findings}
Ond{\v{r}}ej Bojar, Rajen Chatterjee, Christian Federmann, Yvette Graham, Barry
  Haddow, Matthias Huck, Antonio~Jimeno Yepes, Philipp Koehn, Varvara
  Logacheva, Christof Monz, et~al.
\newblock Findings of the 2016 conference on machine translation.
\newblock In {\em Proceedings of the First Conference on Machine Translation:
  Volume 2, Shared Task Papers}, pages 131--198, 2016.

\bibitem{bommasani2021opportunities}
Rishi Bommasani, Drew~A Hudson, Ehsan Adeli, Russ Altman, Simran Arora, Sydney
  von Arx, Michael~S Bernstein, Jeannette Bohg, Antoine Bosselut, Emma
  Brunskill, et~al.
\newblock On the opportunities and risks of foundation models.
\newblock {\em arXiv preprint arXiv:2108.07258}, 2021.

\bibitem{brown2020language}
Tom Brown, Benjamin Mann, Nick Ryder, Melanie Subbiah, Jared~D Kaplan, Prafulla
  Dhariwal, Arvind Neelakantan, Pranav Shyam, Girish Sastry, Amanda Askell,
  et~al.
\newblock Language models are few-shot learners.
\newblock {\em Advances in neural information processing systems},
  33:1877--1901, 2020.

\bibitem{carlini2021extracting}
Nicholas Carlini, Florian Tramer, Eric Wallace, Matthew Jagielski, Ariel
  Herbert-Voss, Katherine Lee, Adam Roberts, Tom~B Brown, Dawn Song, Ulfar
  Erlingsson, et~al.
\newblock Extracting training data from large language models.
\newblock In {\em USENIX Security Symposium}, volume~6, 2021.

\bibitem{chakraborty2023possibilities}
Souradip Chakraborty, Amrit~Singh Bedi, Sicheng Zhu, Bang An, Dinesh Manocha,
  and Furong Huang.
\newblock On the possibilities of ai-generated text detection.
\newblock {\em arXiv preprint arXiv:2304.04736}, 2023.

\bibitem{chen2023gpt}
Yutian Chen, Hao Kang, Vivian Zhai, Liangze Li, Rita Singh, and Bhiksha
  Ramakrishnan.
\newblock Gpt-sentinel: Distinguishing human and chatgpt generated content.
\newblock {\em arXiv preprint arXiv:2305.07969}, 2023.

\bibitem{devlin-etal-2019-bert}
Jacob Devlin, Ming-Wei Chang, Kenton Lee, and Kristina Toutanova.
\newblock {BERT}: Pre-training of deep bidirectional transformers for language
  understanding.
\newblock In {\em Proceedings of the 2019 Conference of the North {A}merican
  Chapter of the Association for Computational Linguistics: Human Language
  Technologies}, pages 4171--4186, 2019.

\bibitem{Else2023AbstractsWB}
Holly Else.
\newblock Abstracts written by chatgpt fool scientists.
\newblock {\em Nature}, 613:423 -- 423, 2023.

\bibitem{fan-etal-2019-eli5}
Angela Fan, Yacine Jernite, Ethan Perez, David Grangier, Jason Weston, and
  Michael Auli.
\newblock {ELI}5: Long form question answering.
\newblock In {\em Proceedings of the 57th Annual Meeting of the Association for
  Computational Linguistics}, pages 3558--3567, Florence, Italy, July 2019.
  Association for Computational Linguistics.

\bibitem{fan-etal-2018-hierarchical}
Angela Fan, Mike Lewis, and Yann Dauphin.
\newblock Hierarchical neural story generation.
\newblock In {\em Proceedings of the 56th Annual Meeting of the Association for
  Computational Linguistics (Volume 1: Long Papers)}, pages 889--898,
  Melbourne, Australia, July 2018. Association for Computational Linguistics.

\bibitem{gao2022comparing}
Catherine~A Gao, Frederick~M Howard, Nikolay~S Markov, Emma~C Dyer, Siddhi
  Ramesh, Yuan Luo, and Alexander~T Pearson.
\newblock Comparing scientific abstracts generated by chatgpt to original
  abstracts using an artificial intelligence output detector, plagiarism
  detector, and blinded human reviewers.
\newblock {\em bioRxiv}, pages 2022--12, 2022.

\bibitem{gehrmann-etal-2019-gltr}
Sebastian Gehrmann, Hendrik Strobelt, and Alexander Rush.
\newblock {GLTR}: Statistical detection and visualization of generated text.
\newblock In {\em Proceedings of the 57th Annual Meeting of the Association for
  Computational Linguistics: System Demonstrations}, pages 111--116, 2019.

\bibitem{grechnikov2009detection}
EA~Grechnikov, GG~Gusev, AA~Kustarev, and AM~Raigorodsky.
\newblock Detection of artificial texts.
\newblock {\em RCDL2009 Proceedings. Petrozavodsk}, pages 306--308, 2009.

\bibitem{holtzmancurious}
Ari Holtzman, Jan Buys, Li~Du, Maxwell Forbes, and Yejin Choi.
\newblock The curious case of neural text degeneration.
\newblock In {\em International Conference on Learning Representations}, 2020.

\bibitem{ippolito-etal-2020-automatic}
Daphne Ippolito, Daniel Duckworth, Chris Callison-Burch, and Douglas Eck.
\newblock Automatic detection of generated text is easiest when humans are
  fooled.
\newblock In {\em Proceedings of the 58th Annual Meeting of the Association for
  Computational Linguistics}, pages 1808--1822, July 2020.

\bibitem{jin-etal-2019-pubmedqa}
Qiao Jin, Bhuwan Dhingra, Zhengping Liu, William Cohen, and Xinghua Lu.
\newblock {P}ub{M}ed{QA}: A dataset for biomedical research question answering.
\newblock In {\em Proceedings of the 2019 Conference on Empirical Methods in
  Natural Language Processing and the 9th International Joint Conference on
  Natural Language Processing (EMNLP-IJCNLP)}, pages 2567--2577, 2019.

\bibitem{kirchenbauer2023watermark}
John Kirchenbauer, Jonas Geiping, Yuxin Wen, Jonathan Katz, Ian Miers, and Tom
  Goldstein.
\newblock A watermark for large language models.
\newblock {\em arXiv preprint arXiv:2301.10226}, 2023.

\bibitem{krishna2023paraphrasing}
Kalpesh Krishna, Yixiao Song, Marzena Karpinska, John Wieting, and Mohit Iyyer.
\newblock Paraphrasing evades detectors of ai-generated text, but retrieval is
  an effective defense.
\newblock {\em arXiv preprint arXiv:2303.13408}, 2023.

\bibitem{le2012asymptotic}
Lucien Le~Cam.
\newblock {\em Asymptotic methods in statistical decision theory}.
\newblock Springer Science \& Business Media, 2012.

\bibitem{li2023origin}
Linyang Li, Pengyu Wang, Ke~Ren, Tianxiang Sun, and Xipeng Qiu.
\newblock Origin tracing and detecting of llms.
\newblock {\em arXiv preprint arXiv:2304.14072}, 2023.

\bibitem{li2023deepfake}
Yafu Li, Qintong Li, Leyang Cui, Wei Bi, Longyue Wang, Linyi Yang, Shuming Shi,
  and Yue Zhang.
\newblock Deepfake text detection in the wild.
\newblock {\em arXiv preprint arXiv:2305.13242}, 2023.

\bibitem{liang2023gpt}
Weixin Liang, Mert Yuksekgonul, Yining Mao, Eric Wu, and James Zou.
\newblock Gpt detectors are biased against non-native english writers.
\newblock {\em arXiv preprint arXiv:2304.02819}, 2023.

\bibitem{meritypointer}
Stephen Merity, Caiming Xiong, James Bradbury, and Richard Socher.
\newblock Pointer sentinel mixture models.
\newblock In {\em International Conference on Learning Representations}, 2017.

\bibitem{mitchell2023detectgpt}
Eric Mitchell, Yoonho Lee, Alexander Khazatsky, Christopher~D Manning, and
  Chelsea Finn.
\newblock Detectgpt: Zero-shot machine-generated text detection using
  probability curvature.
\newblock {\em arXiv preprint arXiv:2301.11305}, 2023.

\bibitem{DBLP:conf/icml/NaeemOUCY20}
Muhammad~Ferjad Naeem, Seong~Joon Oh, Youngjung Uh, Yunjey Choi, and Jaejun
  Yoo.
\newblock Reliable fidelity and diversity metrics for generative models.
\newblock In {\em Proceedings of the 37th International Conference on Machine
  Learning, {ICML} 2020, 13-18 July 2020, Virtual Event}, volume 119 of {\em
  Proceedings of Machine Learning Research}, pages 7176--7185. {PMLR}, 2020.

\bibitem{narayan-etal-2018-dont}
Shashi Narayan, Shay~B. Cohen, and Mirella Lapata.
\newblock Don{'}t give me the details, just the summary! topic-aware
  convolutional neural networks for extreme summarization.
\newblock In {\em Proceedings of the 2018 Conference on Empirical Methods in
  Natural Language Processing}, pages 1797--1807, 2018.

\bibitem{GPT35}
OpenAI.
\newblock {O}pen{AI} {M}odels - {GPT}3.5, 2022.

\bibitem{AITextClassifier}
OpenAI.
\newblock {AI} text classifier, Jan 2023.

\bibitem{openai2023gpt}
OpenAI.
\newblock Gpt-4 technical report.
\newblock {\em arXiv}, 2023.

\bibitem{ouyang2022training}
Long Ouyang, Jeffrey Wu, Xu~Jiang, Diogo Almeida, Carroll Wainwright, Pamela
  Mishkin, Chong Zhang, Sandhini Agarwal, Katarina Slama, Alex Ray, et~al.
\newblock Training language models to follow instructions with human feedback.
\newblock {\em Advances in Neural Information Processing Systems},
  35:27730--27744, 2022.

\bibitem{polyanskiy2022information}
Yury Polyanskiy and Yihong Wu.
\newblock Information theory: From coding to learning, 2022.

\bibitem{radford2019language}
Alec Radford, Jeffrey Wu, Rewon Child, David Luan, Dario Amodei, Ilya
  Sutskever, et~al.
\newblock Language models are unsupervised multitask learners.
\newblock {\em OpenAI blog}, 1(8):9, 2019.

\bibitem{raffel2020exploring}
Colin Raffel, Noam Shazeer, Adam Roberts, Katherine Lee, Sharan Narang, Michael
  Matena, Yanqi Zhou, Wei Li, and Peter~J Liu.
\newblock Exploring the limits of transfer learning with a unified text-to-text
  transformer.
\newblock {\em The Journal of Machine Learning Research}, 21(1):5485--5551,
  2020.

\bibitem{sadasivan2023can}
Vinu~Sankar Sadasivan, Aounon Kumar, Sriram Balasubramanian, Wenxiao Wang, and
  Soheil Feizi.
\newblock Can ai-generated text be reliably detected?
\newblock {\em arXiv preprint arXiv:2303.11156}, 2023.

\bibitem{schulman2022chatgpt}
J~Schulman, B~Zoph, C~Kim, J~Hilton, J~Menick, J~Weng, JFC Uribe, L~Fedus,
  L~Metz, M~Pokorny, et~al.
\newblock Chatgpt: Optimizing language models for dialogue, 2022.

\bibitem{solaiman2019release}
Irene Solaiman, Miles Brundage, Jack Clark, Amanda Askell, Ariel Herbert-Voss,
  Jeff Wu, Alec Radford, Gretchen Krueger, Jong~Wook Kim, Sarah Kreps, et~al.
\newblock Release strategies and the social impacts of language models.
\newblock {\em arXiv preprint arXiv:1908.09203}, 2019.

\bibitem{stokel2022ai}
Chris Stokel-Walker.
\newblock Ai bot chatgpt writes smart essays-should academics worry?
\newblock {\em Nature}, 2022.

\bibitem{thoppilan2022lamda}
Romal Thoppilan, Daniel De~Freitas, Jamie Hall, Noam Shazeer, Apoorv
  Kulshreshtha, Heng-Tze Cheng, Alicia Jin, Taylor Bos, Leslie Baker, Yu~Du,
  et~al.
\newblock Lamda: Language models for dialog applications.
\newblock {\em arXiv preprint arXiv:2201.08239}, 2022.

\bibitem{GPTZero}
Edward Tian.
\newblock Gptzero: An ai text detector, 2023.

\bibitem{touvron2023llama}
Hugo Touvron, Thibaut Lavril, Gautier Izacard, Xavier Martinet, Marie-Anne
  Lachaux, Timoth{\'e}e Lacroix, Baptiste Rozi{\`e}re, Naman Goyal, Eric
  Hambro, Faisal Azhar, et~al.
\newblock Llama: Open and efficient foundation language models.
\newblock {\em arXiv preprint arXiv:2302.13971}, 2023.

\bibitem{vaswani2017attention}
Ashish Vaswani, Noam Shazeer, Niki Parmar, Jakob Uszkoreit, Llion Jones,
  Aidan~N Gomez, {\L}ukasz Kaiser, and Illia Polosukhin.
\newblock Attention is all you need.
\newblock {\em Advances in neural information processing systems}, 30, 2017.

\bibitem{verma2023ghostbuster}
Vivek Verma, Eve Fleisig, Nicholas Tomlin, and Dan Klein.
\newblock Ghostbuster: Detecting text ghostwritten by large language models,
  2023.

\bibitem{wang2023bot}
Hong Wang, Xuan Luo, Weizhi Wang, and Xifeng Yan.
\newblock Bot or human? detecting chatgpt imposters with a single question.
\newblock {\em arXiv preprint arXiv:2305.06424}, 2023.

\bibitem{wang2023m4}
Yuxia Wang, Jonibek Mansurov, Petar Ivanov, Jinyan Su, Artem Shelmanov, Akim
  Tsvigun, Chenxi Whitehouse, Osama~Mohammed Afzal, Tarek Mahmoud, Alham~Fikri
  Aji, and Preslav Nakov.
\newblock M4: Multi-generator, multi-domain, and multi-lingual black-box
  machine-generated text detection, 2023.

\bibitem{wasserman2013}
Larry Wasserman.
\newblock Lecture notes for stat 705: Advanced data analysis.
\newblock \url{https://www.stat.cmu.edu/~larry/=stat705/Lecture27.pdf}, 2013.
\newblock Accessed on April 9, 2023.

\bibitem{wu2023llmdet}
Kangxi Wu, Liang Pang, Huawei Shen, Xueqi Cheng, and Tat-Seng Chua.
\newblock Llmdet: A large language models detection tool, 2023.

\bibitem{yu2023bag}
Weichen Yu, Tianyu Pang, Qian Liu, Chao Du, Bingyi Kang, Yan Huang, Min Lin,
  and Shuicheng Yan.
\newblock Bag of tricks for training data extraction from language models.
\newblock {\em arXiv preprint arXiv:2302.04460}, 2023.

\bibitem{yu2023gpt}
Xiao Yu, Yuang Qi, Kejiang Chen, Guoqiang Chen, Xi~Yang, Pengyuan Zhu, Weiming
  Zhang, and Nenghai Yu.
\newblock Gpt paternity test: Gpt generated text detection with gpt genetic
  inheritance.
\newblock {\em arXiv preprint arXiv:2305.12519}, 2023.

\bibitem{zellers2019defending}
Rowan Zellers, Ari Holtzman, Hannah Rashkin, Yonatan Bisk, Ali Farhadi,
  Franziska Roesner, and Yejin Choi.
\newblock Defending against neural fake news.
\newblock {\em Advances in neural information processing systems}, 32, 2019.

\bibitem{zhan2023g3detector}
Haolan Zhan, Xuanli He, Qiongkai Xu, Yuxiang Wu, and Pontus Stenetorp.
\newblock G3detector: General gpt-generated text detector.
\newblock {\em arXiv preprint arXiv:2305.12680}, 2023.

\bibitem{zhang2022opt}
Susan Zhang, Stephen Roller, Naman Goyal, Mikel Artetxe, Moya Chen, Shuohui
  Chen, Christopher Dewan, Mona Diab, Xian Li, Xi~Victoria Lin, et~al.
\newblock Opt: Open pre-trained transformer language models.
\newblock {\em arXiv preprint arXiv:2205.01068}, 2022.

\end{thebibliography}
\clearpage
\appendix

 \appendix
 \begin{center}
    {\LARGE Appendix: Dynamic Prompting}
 \end{center}
\section{Broader Impacts and Limitations}
Due to the extensive workload of experiments, we only test our methods for classification tasks from the SuperGLUE benchmark. Additional text generation tasks could be exploited in future work. Also, our conclusions are drawn from the encoder-decoder architecture such as T5, BERT and RoBERTa models, vision pre-trained model ViT-B and a pre-trained ViT-B/16 CLIP model. And it is worth investigating whether dynamic prompting still holds for  decoder-only GPT. Besides, our approach introduces additional parameters like anneal temperature or learning network, which could increase the difficulty of optimization.
\section{Derivation of the Unified View of Prompt Tuning}\label{app:derivation}

For the new query $x' = [P_1; x; P_2] $, the attention head module becomes:

\begin{align}
    \textit{Head} & = \textit{Attn}\left([P_1; x; P_2]W^Q, \right.
    \left. [P_1; x; P_2]W^K, [P_1; x; P_2]W^V\right)\\
    &= \softmax\left(\frac{Q'*K'^T}{\sqrt{d}}\right)V'\\
    \intertext{omitting $\sqrt{d}$ for brevity}\nonumber\\
    & =\left[ \softmax(P_1W^QK'^T)V'; \softmax(xW^Q K'^T)V' ; \softmax(P_2W^Q K'^T)V' \right], 
\end{align}

 where 
\begin{align}
    &\softmax(P_1W^QK'^T)V' \nonumber\\
    & = \softmax\left(P_1W^Q [K_1^T; K^T; K_2^T] \right) \begin{bmatrix} V_1 \\ V \\ V_2 \end{bmatrix}\\
    & = \lambda_1*\softmax(Q_1 K_1^T)V_1 + \lambda_2*\softmax(Q_1 K_2^T)V_2 + (1 - \lambda_1 - \lambda_2)*\softmax(Q_1 K^T)V \\
    & = \lambda_1* {Attn(Q_1, K_1, V_1)} + \lambda_2*{Attn(Q_1, K_2, V_2)} +( 1 - \lambda_1 - \lambda_2)* {Attn(Q_1, K, V )},
\end{align}
 
where $\lambda_1$ and $\lambda_2$ are normalized weights:
\small{
$$ \lambda_1 = \frac{\sum_i exp(Q_1 K_1^T)_i} { \sum_{j=1}^{2} \sum_i exp(Q_1 K_j^T)_i + \sum_i exp(Q_1 K^T)_i }, \quad
 \lambda_2 = \frac{\sum_i exp(Q_1 K_2^T)_i} {  \sum_{j=1}^{2} \sum_i exp(Q_1 K_j^T)_i + \sum_i exp(Q_1 K^T)_i }. $$} \normalsize
\begin{align}
    &\softmax(xW^Q K'^T)V'\nonumber\\
    &= \softmax\left(xW^Q [K_1^T; K^T; K_2^T] \right) \begin{bmatrix} V_1 \\ V \\ V_2 \end{bmatrix} \\
    &= \beta_1*\softmax(Q K_1^T)V_1 + \beta_2*\softmax(Q K_2^T)V_2 + (1 - \beta_1 - \beta_2)*\softmax(Q K^T)V \\
    &= \beta_1*  {Attn(Q, K_1, V_1)} + \beta_2* {Attn(Q, K_2, V_2)} +( 1 - \beta_1 - \beta_2)*  {Attn(Q, K, V )},
\end{align}

where $\beta_1$ and $\beta_2$ are normalized weights:
\small{
$$ \beta_1 = \frac{\sum_i exp(Q K_1^T)_i} { \sum_{j=1}^{2} \sum_i exp(Q K_j^T)_i + \sum_i exp(Q K^T)_i }, \quad \beta_2 = \frac{\sum_i exp(Q K_2^T)_i} {  \sum_{j=1}^{2} \sum_i exp(Q K_j^T)_i + \sum_i exp(Q K^T)_i }.$$}\normalsize

\begin{align}
    & \softmax(p_2W^Q K'^T)V'\nonumber\\
    &= \softmax\left(p_2W^Q [K_1^T; K^T; K_2^T] \right) \begin{bmatrix} V_1 \\ V \\ V_2 \end{bmatrix}\\
    & = \gamma_1*\softmax(Q_2 K_1^T)V_1 + \gamma_2*\softmax(Q_2 K_2^T)V_2 + (1 - \gamma_1 - \gamma_2)*\softmax(Q_2 K^T)V \\
    & = \gamma_1*  {Attn(Q_2, K_1, V_1)} + \gamma_2* {Attn(Q_2, K_2, V_2)} +( 1 - \gamma_1 - \gamma_2)* {Attn(Q_2, K, V )},
\end{align}

where $\gamma_1$ and $\gamma_2$ are normalized weights:
\small{
$$ \gamma_1 = \frac{\sum_i exp(Q_2 K_1^T)_i} { \sum_{j=1}^{2} \sum_i exp(Q_2 K_j^T)_i + \sum_i exp(Q_2 K^T)_i }, \quad \gamma_2 = \frac{\sum_i exp(Q_2 K_2^T)_i} {  \sum_{j=1}^{2} \sum_i exp(Q_2 K_j^T)_i + \sum_i exp(Q_2 K^T)_i }. $$
}\normalsize

Finally, we can write them together:

\small{
\begin{align}
    &  Head= Attn(x', K', V')\nonumber \\ 
    & = \left [ \lambda_1* \underbrace{Attn(Q_1, K_1, V_1)}_{\textit{ \textcolor{red} {prompt tuning} }} + \lambda_2*\underbrace{Attn(Q_1, K_2, V_2)}_{\textit{  {postfix} }}\right.  +( 1 - \lambda_1 - \lambda_2)*\underbrace{Attn(Q_1, K, V )}_{ \textcolor{red} {\textit{ prompt tuning}}} ; \nonumber \\
    & \beta_1* \underbrace{Attn(Q, K_1, V_1)}_{\textit{\textcolor{red} {prompt tuning} }} + \beta_2*\underbrace{Attn(Q, K_2, V_2)}_{\textit{postfix}} +( 1 - \beta_1 - \beta_2)*\underbrace{ \textcolor{blue} {Attn(Q, K, V )} }_{\textit{ \textcolor{blue} { standard} }};\nonumber\\
    & \gamma_1* \underbrace{Attn(Q_2, K_1, V_1)}_{\textit{postfix}} + \gamma_2*\underbrace{Attn(Q_2, K_2, V_2)}_{\textit{postfix}}\left. +( 1 - \gamma_1 - \gamma_2)*\underbrace{Attn(Q_2, K, V )}_{\textit{postfix}} \right ].
\end{align}
}\normalsize

\section{Dynamic Insertion Position with Gumbel-Softmax}
\label{sec:gumbel}

We use the Gumbel-Max trick \citep{maddison2017concrete} to dynamically decide insertion position for soft prompts. 
Specifically, to decide $dpos$, we have $(l+1)$ positions to choose. Let
\{$\alpha_0$, $\cdots$,$\alpha_{l}$\} represent the log probabilities \{$\log$($p_0$),$\cdots$, $\log$($p_{l}$)\} of different insertion positions. $\bm\alpha$ is the output logit of the network $POS_\theta$.
Thus, we can draw samples in the following way: we first draw i.i.d samples \{$g_0$, $\cdots$, $g_{l}$\} from a Gumbel distribution, i.e., $g = -\log(-\log(z)) \sim Gumbel$, 
where $z$ $\sim$ Uniform(0, 1). Then we produce the discrete sample by adding $g$ to introduce stochasticity:
\begin{gather}
dpos = \argmax_{i}[\alpha_i + g_i], i \in \{0,\cdots,l\}.
\end{gather}

\begin{gather}
\label{eq:mu}
\mu_i = \exp{((\alpha_i+g_i})/\tau)/ (\sum^{l}_{\hat{i}=0}\exp{((\alpha_{\hat{i}}+g_{\hat{i}}})/\tau)),i \in \{0,\cdots,l\}
\end{gather}

The $\argmax$ operation is non-differentiable, but we can use the $\softmax$ as a continuously differentiable approximation to it (Eq. \eqref{eq:mu}). $\tau$ is the temperature to control the discreteness. Thus, we use the $\argmax$ to make the discrete selection on the forward pass, while approximating it with softmax on the backward pass, which is called the straight-through estimator \citep{jang2017categorical}.

\section{Dynamic Length}\label{app:dl}
The huggingface transformers \footnote{https://huggingface.co/} implemented the attention mask mechanism by giving infinite minus value to padded tokens so that the calculated attention score will reach zero.
\begin{equation}
Attention = \softmax(\frac{Q*K^{T}}{\sqrt{d_k}} + M)*V, 
\end{equation}
where 
\begin{equation}
M =
\begin{cases}
0, & \text{mask=1}\\
-\infty, & \text{mask=0}.
\end{cases}
\end{equation}

It seems natural to just treat truncated prompts as padding tokens with $mask=0$. 
However, when the prompt length $l_i$ is dynamically updated in input instance $x_i$, the logits returned by Gumbel-Softmax can not be directly applied to the attention mask matrix M since M can not provide gradients. Therefore, we adopt a surrogate strategy by 
\begin{equation}
P_{new} = [0*P_{before} ; P_{after}],
\end{equation}
where $P_{before} \in \mathbb{R}^{(l-l_i) \times d} $ and $P_{after} \in \mathbb{R}^{l_i \times d} $.
In this way, although the attention score after $\softmax$ is $\frac{e^0}{\sum_{i} (q_i*k_i)}$ rather than $0$, the corresponding value in $V$ is $0$. Such implementation is not optimal, but we stick to it for simplicity.

\section{Experimental Settings}\label{sec:expsetting}

\subsection{Implementation details}\label{exp:gumbel}
We find the initialization of prompts could have influential effects on the final performance. Initializing from the vocabulary of LMs almost always gives better results. We thus follow the default setting in OpenPrompt using the list of embeddings in front of the token vocabulary as the initialization of soft prompt vectors. Setting warm-up steps to 500 yields consistent gains for the small, base, and large models. However, for T5-XL, a warm-up step of 500 does not lead to convergence, and we reduce it to 10 steps. Also, we only run limited experiments with T5-XL due to the computational resource constraint.

We empirically find the annealing temperature is very important for Gumbel-Softmax \citep{DBLP:conf/iclr/JangGP17} to behave well. We follow \citep{jang2017categorical}, and adjust the annealing temperature by 
\begin{equation}
\tau = max(\tau * exp( -\frac{ \gamma * iterations}{step}), 0.5),\end{equation}
where initial $\tau$ is set to $1.0$ and the annealing rate $\gamma \in \{3e^{-7}, 3e^{-5}, 3e^{-3}\}$ and the $step$ is picked from $[0.1, 1, 10, 30, 100, 200, 600]$. 
In practice, the Gumbel-Softmax simulates more closely to the true categorical distribution when $\tau$ is reaching 0.5.

\stitle{Hardware.}\label{exp:hardware}
We use NVIDIA A40 48 GB for T5-XL and RTX 6000 24 GB for all other models.

\subsection{Model hyperparameters for experiments on T5-series models}
For all one-layer networks for learning instance-dependent dynamic position, length, or representation information, we adopt a single linear layer followed by ReLU activation. 
The input embedding dimension $d$ is 512, 768, 1024, 2048 for T5-small, base, large, and XL, respectively. We illustrate the details of our additional parameters in Table \ref{tab:param}. The batch size is 32 and 16 for dynamic prompting and finetuning. All inputs are truncated to a maximum of 480 tokens. 
For each method, we tune the learning temperature via grid search in the range $\{10^{-0.5},10^{-1},10^{-1.5},10^{-2},10^{-2.5},10^{-3}\}$ to obtain the best performances. $k$ is set to 8 for all settings. In all settings except for comparison on varying length $l$, we set $l$=20.
\begin{table}[h]
\centering
\caption{Tuned parameters in our Dynamic Prompting methods. PTs are soft prompts and $\Theta_{NN}$ is the learning network. $d$ is the input vocabulary embedding size of LMs. $l$ is the prompt length. $k$ is the number of prompts in the prompt pool.
}\label{tab:param}
\scalebox{0.8}{
\begin{tabular}{l@{\hspace{0.8\tabcolsep}}|cccc}
\hline
\multirow{3}{*}{\textbf{Methods}}
 & \multicolumn{4}{c}{\textbf{Number of parameters}}   \\
  \cline{2-5}
 &  \thead{Fixed\\Position}  & \thead{Adaptive \\ pos} & \thead{Adaptive \\ ins\_pos} & \thead{Adaptive \\ ins\_vec} \\
\hline
PTs & $l*d$ & $l*d$ & $l*d$  & $(l+1)*d*k$ \\
$\Theta_{NN}$ & $0$ & $l+1$ & $d*(l+1)$ & $d*k$  \\
\hline
\end{tabular}
}

\end{table}

\subsection{Model hyperparameters for experiments on P-tuning V2}
Since we intend to test if our approach can lead to accuracy gains when our dynamic insertion position, as discussed in Sec. \ref{unified}, is embedded in different prompt learning frameworks, we use the default setting in the GitHub repo\footnote{https://github.com/THUDM/P-tuning-v2} of P-tuning V2 \citep{Liu2021PTuningVP}. In the repo, the scripts for datasets cb, multirc, and record are missing. The detailed setting is summarized in Table \ref{tab:pt2para}.

\begin{table}[htbp]
\centering
\caption{Parameter setting for P-tuning V2 related evaluation.} \label{tab:pt2para}
\setlength\tabcolsep{1.6pt}
\scalebox{1.}{
\scriptsize{
\begin{tabular}{c|cccc|c|ccccc}
\hline
                                      & \multicolumn{4}{c|}{\textbf{BERT-Large}}                                                                                           &         & \multicolumn{5}{c}{\textbf{RoBERTa-Large}}                                                                                                                       \\ \cline{2-11} 
\multirow{-2}{*}{\textbf{Dataset} }            & \multicolumn{1}{c|}{batch\_size} & \multicolumn{1}{c|}{lr} & \multicolumn{1}{c|}{dropout} & prompt length & \#epoch & \multicolumn{1}{c|}{batch\_size} & \multicolumn{1}{c|}{lr} & \multicolumn{1}{c|}{dropout } & \multicolumn{1}{c|}{prompt length} & \#epoch \\ \hline
{\color[HTML]{1F2328} \textbf{BoolQ}} & \multicolumn{1}{c|}{32}          & \multicolumn{1}{c|}{5.00E-03}      & \multicolumn{1}{c|}{0.1}          & 40            & 100     & \multicolumn{1}{c|}{16}          & \multicolumn{1}{c|}{7.00E-03}      & \multicolumn{1}{c|}{0.1}          & \multicolumn{1}{c|}{8}             & 100     \\ \hline
\textbf{MultiRC (F1a)}                & \multicolumn{1}{c|}{16}          & \multicolumn{1}{c|}{5.00E-03}      & \multicolumn{1}{c|}{0.1}          & 40            & 100     & \multicolumn{1}{c|}{16}          & \multicolumn{1}{c|}{7.00E-03}      & \multicolumn{1}{c|}{0.1}          & \multicolumn{1}{c|}{8}             & 100     \\ \hline
{\color[HTML]{1F2328} \textbf{WiC}}   & \multicolumn{1}{c|}{16}          & \multicolumn{1}{c|}{1.00E-04}      & \multicolumn{1}{c|}{0.1}          & 20            & 80      & \multicolumn{1}{c|}{32}          & \multicolumn{1}{c|}{1.00E-02}      & \multicolumn{1}{c|}{0.1}          & \multicolumn{1}{c|}{8}             & 50      \\ \hline
{\color[HTML]{1F2328} \textbf{CB}}    & \multicolumn{1}{c|}{32}          & \multicolumn{1}{c|}{5.00E-03}      & \multicolumn{1}{c|}{0.1}          & 40            & 100     & \multicolumn{1}{c|}{16}          & \multicolumn{1}{c|}{7.00E-03}      & \multicolumn{1}{c|}{0.1}          & \multicolumn{1}{c|}{8}             & 100     \\ \hline
{\color[HTML]{1F2328} \textbf{RTE}}   & \multicolumn{1}{c|}{16}          & \multicolumn{1}{c|}{1.00E-02}      & \multicolumn{1}{c|}{0.1}          & 20            & 60      & \multicolumn{1}{c|}{32}          & \multicolumn{1}{c|}{5.00E-03}      & \multicolumn{1}{c|}{0.1}          & \multicolumn{1}{c|}{128}           & 100     \\ \hline
{\color[HTML]{1F2328} \textbf{COPA}}  & \multicolumn{1}{c|}{16}          & \multicolumn{1}{c|}{1.00E-02}      & \multicolumn{1}{c|}{0.1}          & 16            & 80      & \multicolumn{1}{c|}{8}           & \multicolumn{1}{c|}{9.00E-03}      & \multicolumn{1}{c|}{0.1}          & \multicolumn{1}{c|}{8}             & 120     \\ \hline
{\color[HTML]{1F2328} \textbf{WSC}}   & \multicolumn{1}{c|}{16}          & \multicolumn{1}{c|}{5.00E-03}      & \multicolumn{1}{c|}{0.1}          & 20            & 80      & \multicolumn{1}{c|}{16}          & \multicolumn{1}{c|}{1.00E-02}      & \multicolumn{1}{c|}{0.1}          & \multicolumn{1}{c|}{8}             & 10      \\ \hline
\textbf{ReCoRD (F1)}                  & \multicolumn{1}{c|}{20}          & \multicolumn{1}{c|}{5.00E-03}      & \multicolumn{1}{c|}{0.1}          & 40            & 100     & \multicolumn{1}{c|}{16}          & \multicolumn{1}{c|}{7.00E-03}      & \multicolumn{1}{c|}{0.1}          & \multicolumn{1}{c|}{8}             & 100     \\ \hline
\end{tabular}
}
}
\end{table}

\subsection{Model hyperparameters for experiments on VPT }
For all datasets on VPT \citep{jia2022vpt}, we use the default setting in the repo\footnote{https://github.com/KMnP/vpt}. The pre-trained backbone we used is ViT-B (``sup\_vitb16\_imagenet21k''). SGD with momentum 0.9, base learning rate 0.25 and weight\_decay 0.001 is used as the optimizer. The batch size is set to 32. The random seed is 0.

\subsection{Model hyperparameters for experiments on MaPLe}
We follow the setting in MaPLe \citep{khattakMaPLe} repo\footnote{https://github.com/muzairkhattak/multimodal-prompt-learning} and use a few-shot training strategy in all experiments at 16 shots which are randomly sampled for each class. For each dataset, we run three times using seed values 1, 2, and 3, and report the average accuracy. We apply prompt tuning on a pretrained ViT-B/16 CLIP model. We set prompt depth $J$ to 9 and
the language and vision prompt lengths to 4. All models are trained for 12 epochs with a batch size of 4 and a learning rate of 0.0035 via SGD optimizer. We initialize the language prompts of the first layer with the pretrained CLIP word embeddings of the template category ``for sure This is a photo of <category>''. This is different from the original setting in MaPLe which use only prompt length 2 with initialization using ``a photo
of a <category>'', as we need to leave room for our algorithm to select a good insertion position. The setting for the MaPLe baseline also follows the same setting for a fair comparison. Since the task depends on the calculation of similarity between text and image embeddings, and there is no instance-dependent information, we only validate the $adap\_pos$ setting.

\subsection{Details of datasets}\label{sec:datasetapp}
Following previous work \citep{ding-etal-2022-openprompt}, we evaluate our approach on five SuperGlue \citep{wang2019superglue} datasets to test the natural language understanding ability, namely BoolQ \citep{clark-etal-2019-boolq}, MultiRC \citep{khashabi-etal-2018-looking}, CB \citep{de2019commitmentbank}, RTE \citep{giampiccolo-etal-2007-third}, and WiC \citep{pilehvar-camacho-collados-2019-wic}. We use the default train/dev/test split and report the default metric on the validation set since the test set is not directly available. For comparison with P-tuning V2, we also use SuperGlue datasets. For the vision prompt tuning setting, we follow \citep{jia2022vpt} and use the well-known FGVC benchmark datasets consisting of 5 benchmarked Fine-Grained Visual Classification tasks including CUB-200-2011, NABirds, Oxford Flowers, Stanford Dogs, and Stanford Cars. For the vision-language setting, we follow MaPLe \citep{khattakMaPLe} and use 11 datasets including Caltech101,
OxfordPets, StanfordCars, Flowers102, Food101, FGVCAircraft, SUN397, UCF101, DTD, EuroSAT, and ImageNet-R.

\section{Case and Ablation Study}\label{analysis}
In this section, we conduct a case study to explore how the learned position differs from each other and an ablation study to investigate the influence of original prompt length and instance representation.

\begin{figure}
\centering
\includegraphics[width=.4\textwidth]{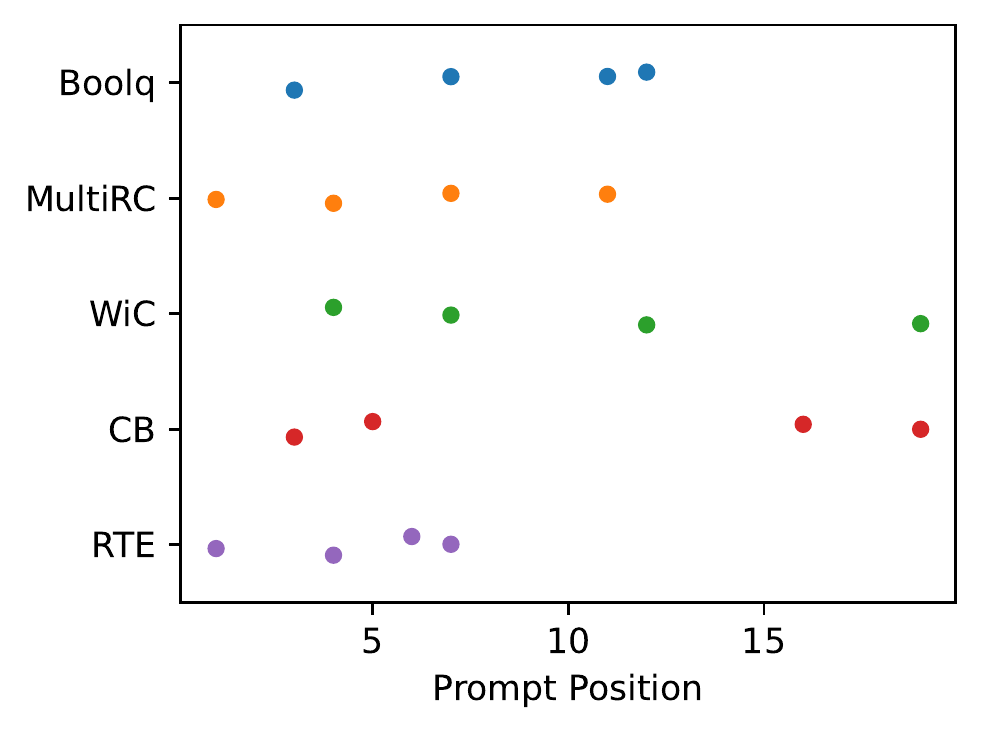} 
    \caption{ Comparison of learned best position on 5 datasets under 4 random seeds. }\label{aba:position}
\end{figure}

\stitle{Adaptive Position.}
We first look at how the learned dynamic position differs from each other across five datasets. We run three experiments on T5-small and record the final optimal position on each task, where the initial prompt length is fixed at 20. Figure \ref{aba:position} shows that the most suitable position varies across tasks and individual runs. This suggests that the optimal position depends on the specific tasks and prompt representations. There is no one-for-all solution. 

\stitle{Influence of Input Representation.}\label{aba:input_rep}
Since our instance-dependent dynamic prompting requires generating the sentence representation, where we run a forward of the LMs to get $LM(X)$. In IDPG \citep{wu2022idpg}, the authors use the Glove \citep{pennington-etal-2014-glove} embedding as the sentence representation and obtain close prompt tuning performance. And they also suggest caching for applications in downstream tasks. But we believe those methods add additional complexity for deployment.
To overcome the drawback of feeding one instance twice into the LMs, we try an alternative method: use the initial vocabulary representation from T5 of the input sentence as the input for the learning networks to generate dynamically learned prompts.
We perform $adap\_ins\_pos$ experiments on RTE and CB datasets in T5-small and T5-large. The results are shown in Table \ref{tab:vocab}. As we can see, the performance is almost maintained for all cases. Our operation does not introduce additional models or complexity, thus suitable for various downstream tasks.
\begin{table}
\centering
\caption{Ablation study on the effects of sentence representation.
}\label{tab:vocab}
\scalebox{0.8}{
\begin{tabular}{l@{\hspace{0.8\tabcolsep}}|cc|cc}
\hline
\multirow{2}{*}{\textbf{Methods}}
 & \multicolumn{2}{c|}{\textbf{T5-LM-Small}}  & \multicolumn{2}{c}{\textbf{T5-LM-Large}}  \\
  \cline{2-5}
 &  CB & RTE &  CB  & RTE \\
\hline
T5-LMs & $83.93$  & $66.79$   & $89.29$  & $85.71$ \\
Vocabulary  & $83.39$ & $66.53$  & $88.91$ & $85.29$  \\
\hline
\end{tabular}
}
\end{table}

\stitle{Influence of Prompt Length.}
We conduct an ablation study to investigate whether our dynamic position approach still works when the total prompt length varies. As shown in Table \ref{tab:fixed1} and Appendix \ref{app:fix_l}, we first perform a greedy search over $L = [2, 4, 8, 16, 20, 32]$ to find the optimal prompt length. Then based on the searched best length, we run different dynamic position experiments. The results show that our adaptive position consistently increases the performance when $L$ is set to $32$. Besides, combining a prompt pool (adaptive vector) can further improve the results. In particular, when the $L$ is set to only $4$ for T5-Large, our adaptive position can still maintain superiority on most datasets. These results show that our dynamic position can universally improve dynamic prompting for varying prompt lengths, even if $L$ is extremely small. Recall the unified view in Sec. \ref{unified}, and these results validate the hypothesis that prepended prompts are not enough, and a slight change leads to significant improvements.

\section{Additional Results}\label{sec:appadditional}

\subsection{Additional comparison of adaptive vector} \label{app:adap_vec}
\begin{figure}[htbp]
  \centering
  \begin{minipage}[b]{0.49\linewidth}
    \centering
    \includegraphics[width=1\textwidth]{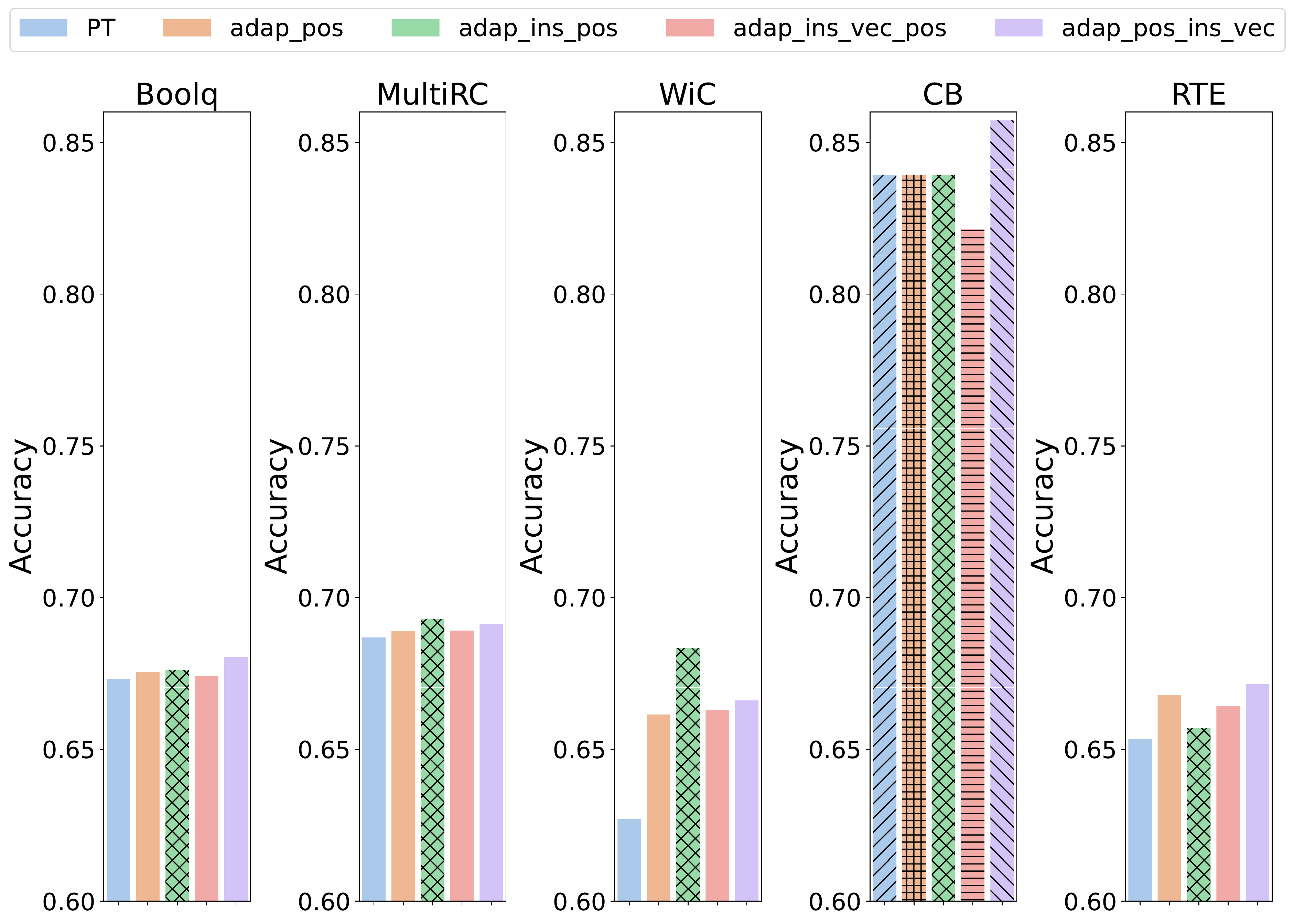} %adaptive_vector_small.pdf} 
    \caption{Results on SuperGLUE with T5-Small. } \label{app:adap_vec_small}
  \end{minipage}
  %\hspace{0.5cm}
  \begin{minipage}[b]{0.49\linewidth}
    \centering
    \includegraphics[width=1\textwidth]{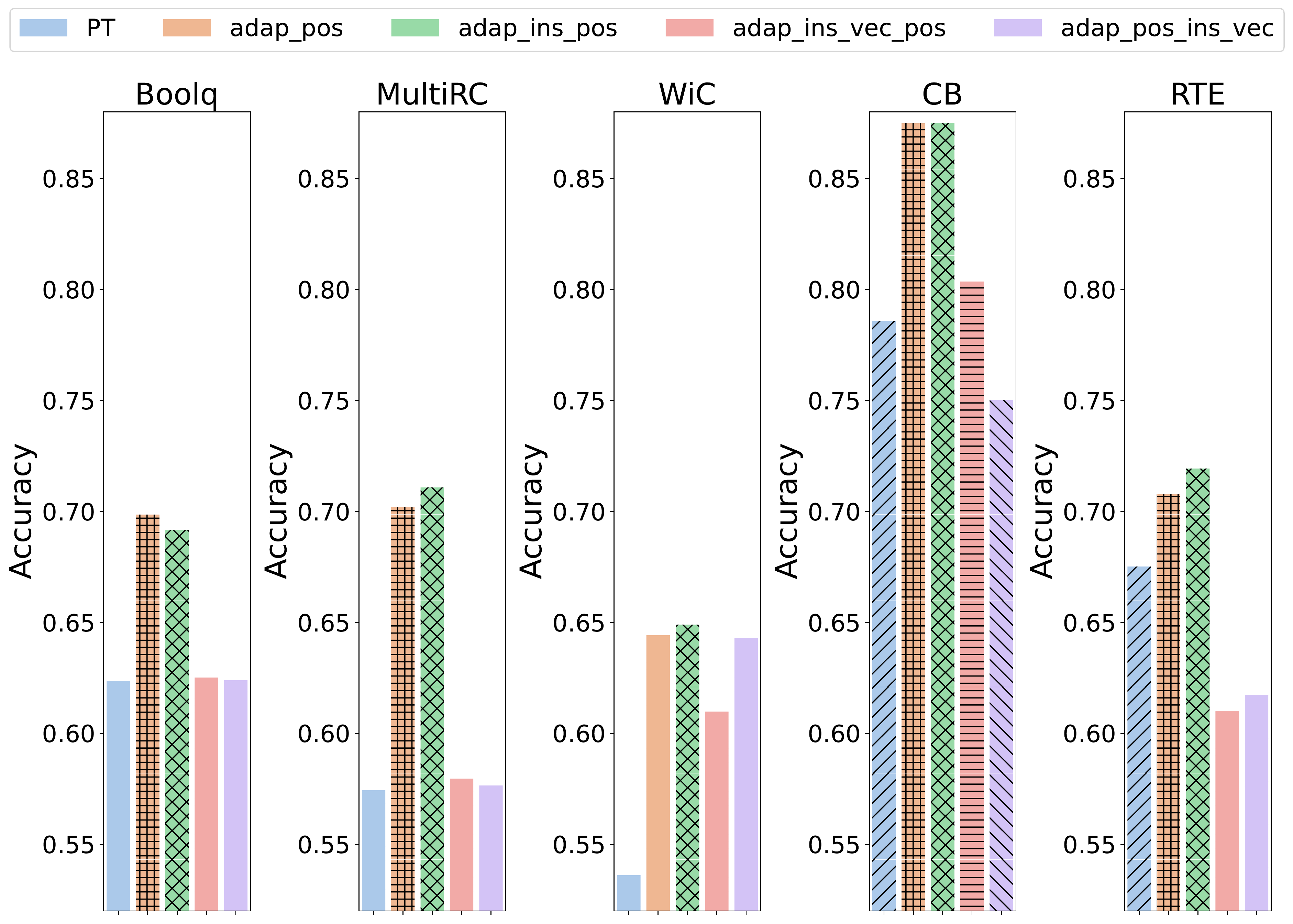}%adaptive_vector_base.pdf} 
    \caption{Results on SuperGLUE with T5-Base. } \label{app:adap_vec_base}
  \end{minipage}
\end{figure}

Here we show the additional results for T5-Small in Figure \ref{app:adap_vec_small}, and T5-Base in Figure \ref{app:adap_vec_base}. We observe a similar trend as T5-Large in Sec. \ref{5:adap_vec}, but T5-Base demonstrates worse results.

\subsection{Additional influence of prompt length }\label{app:fix_l}
In this section, we present the comprehensive results obtained from T5-Small (Table \ref{tab:fixed1}), T5-Base (Table \ref{tab:fixed2}), and T5-Large (Table \ref{tab:fixed3}) models, respectively. To ensure a thorough analysis, we conducted experiments using different prompt lengths, namely 2, 4, 8, 16, 20, and 32. Our aim was to identify the optimal prompt length for each model size across the five SuperGlue datasets and validate the effectiveness of our technique over the best run of baseline prompt tuning.
For T5-Small, we discovered that a prompt length of 32 consistently yielded the highest average accuracy across the SuperGlue datasets. Consequently, for this particular model size, we proceeded to employ our approaches, adapting the prompt position and representation in an adaptive manner. The prompt lengths of 8 and 4 were adopted for T5-Base and T5-Large models respectively.

An in-depth analysis of the results reveals a remarkable trend across all T5 models. Our dynamic prompting techniques consistently outperformed the best run achieved by traditional prompt tuning with a fixed length. This compelling observation unequivocally illustrates the effectiveness and prowess of our approach. Furthermore, as the model size increased from T5-Small to T5-Base and T5-Large, we observed a corresponding increase in the accuracy gain facilitated by our dynamic prompting techniques. This empirical evidence reaffirms the potency of our methodology as model complexity grows.

The findings presented here provide substantial evidence of the superiority of our dynamic prompting techniques over traditional fixed-length prompt tuning. This knowledge empowers researchers and practitioners alike to leverage the full potential of dynamic prompts, unlocking new avenues for improved performance in various natural language processing tasks.
\begin{table*}
\centering
\caption{The left column shows the results of prompt tuning under different lengths, and $L$=32 performs the best. The right column shows different dynamic position strategies by using the $L$=32.
}\label{tab:fixed1}
\setlength\tabcolsep{3.7pt}
\scalebox{0.9}{
\begin{tabular}{l@{\hspace{0.8\tabcolsep}}|cccccc|cccc}
\hline
\multirow{3}{*}{\textbf{Dataset}}
 & \multicolumn{6}{c|}{\textbf{T5-LM-Small}} & \multicolumn{4}{c}{\textbf{T5-LM-Small}}  \\
 %\cline{2-7}
 %& \multicolumn{3}{c|}{test}  & \multicolumn{3}{c}{test} \\
  \cline{2-11}
 & \thead{ Fixed\\L=2} & \thead{ Fixed\\L=4} & \thead{ Fixed\\L=8} & \thead{ Fixed\\L=16} & \thead{ Fixed\\L=20} & \thead{ Fixed\\L=32}  &  \thead{Adaptive\\Position}  & \thead{Adaptive \\ Ins\_pos} & \thead{Adaptive \\Ins\_vec\_pos } & \thead{Adaptive \\ Pos\_ins\_vec}\\
\hline
Boolq & $62.20$ & $62.96$ & $65.23$   & $67.25$ & $67.31$ & $\done \textbf{67.61}$     & $67.00$ & $68.35$ & $67.83$   & $\textbf{67.95}$    \\
MultiRC & $66.17$ & $67.57$ & $68.42$     & $68.30$ & $\textbf{68.69}$ & $\done68.67$      & $68.83$ & $68.38$ & $69.27$   & $\textbf{69.41}$    \\
WiC & $61.29$ & $61.76$ & $63.95$     & $63.48$ & $62.69$ & $\done \textbf{64.11}$       & $65.20$ & $66.14$ & $65.83$   & $\textbf{68.85}$    \\
CB & $75.00$ & $83.93$ & $82.14$     & $83.93$ & $83.93$ & $\done \textbf{87.50}$       & $\textbf{91.07}$ & $85.71$ & $91.07$   & $89.29$    \\
RTE & $61.37$ & $64.98$ & $66.79$     & $\textbf{68.95}$ & $65.34$ & $\done68.23$       & $67.15$ & $\textbf{67.87}$ & $66.79$   & $66.43$    \\
\hline
Avg. & $65.21$ & $68.24$ & $69.31$     & $70.38$ & $69.59$ & $\done \textbf{71.22}$       & $71.85$ & $71.29$ & $72.16$   & $\textbf{72.39}$    \\
\hline
\end{tabular}
}
\end{table*}

\begin{table*}[!t]
\centering
\caption{The left column shows the results of prompt tuning under different lengths, and $L$=8 performs the best. The right column shows different dynamic position strategies by using the $L$=8.
}\label{tab:fixed2}
\setlength\tabcolsep{3.7pt}
\scalebox{0.90}{
\begin{tabular}{l@{\hspace{0.8\tabcolsep}}|cccccc|cccc}
\hline
\multirow{3}{*}{\textbf{Dataset}}
 & \multicolumn{6}{c|}{\textbf{T5-LM-Base}} & \multicolumn{4}{c}{\textbf{T5-LM-Base}}  \\
 %\cline{2-7}
 %& \multicolumn{3}{c|}{test}  & \multicolumn{3}{c}{test} \\
  \cline{2-11}
 &  \thead{ Fixed\\L=2} & \thead{ Fixed\\L=4} & \thead{ Fixed\\L=8} & \thead{ Fixed\\L=16} & \thead{ Fixed\\L=20} & \thead{ Fixed\\L=32}  &  \thead{Adaptive\\Position}  & \thead{Adaptive \\ Ins\_position} & \thead{Adaptive \\Ins\_vec\_pos } & \thead{Adaptive \\ Pos\_ins\_vec}\\
\hline
Boolq & $62.32$ & $65.02$ & $\done62.20$   & $\textbf{70.18}$ & $62.35$ & $62.26$     & $67.95$ & $68.04$ & $\textbf{68.41}$   & $66.18$    \\
MultiRC & $\textbf{58.07}$ & $57.49$ & $\done56.93$     & $57.88$ & $57.43$ & $57.32$    & $69.25$ & $68.13$ & $65.08$   & $\textbf{69.35}$    \\
WiC & $61.12$ & $60.50$ & $\done60.19$     & $\textbf{63.64}$ & $53.61$ & $52.66$    & $65.20$ & $\textbf{66.14}$ & $65.36$   & $63.32$    \\
CB & $85.71$ & $78.57$ & $\done \textbf{94.64}$     & $80.36$ & $78.57$ & $82.14$    & $87.50$ & $87.50$ & $\textbf{92.86}$   & $89.29$    \\
RTE & $59.57$ & $57.04$ & $\done \textbf{68.23}$     & $67.15$ & $67.51$ & $55.60$     & $\textbf{68.59}$ & $68.59$ & $60.29$   & $67.51$    \\
\hline
Avg. & $65.36$ & $63.72$ & $\done \textbf{68.44}$     & $67.84$ & $63.89$ & $62.00$    & $\textbf{71.70}$ & $71.68$ & $70.40$   & $71.13$    \\
\hline
\end{tabular}
}
\end{table*}										

\begin{table}[!t]
\centering
\caption{ The left column shows the results of prompt tuning under different lengths, and $L=4$ performs the best. The right column shows different dynamic position strategies by using the $L=4$.
}\label{tab:fixed3}
\setlength\tabcolsep{3.1pt}
\scalebox{0.9}{
\begin{tabular}{l@{\hspace{0.8\tabcolsep}}|cccccc|cccc}
\hline
\multirow{3}{*}{\textbf{Dataset}}
 & \multicolumn{6}{c|}{\textbf{T5-LM-Large}} & \multicolumn{4}{c}{\textbf{T5-LM-Large}}  \\
 %\cline{2-7}
 %& \multicolumn{3}{c|}{test}  & \multicolumn{3}{c}{test} \\
  \cline{2-11}
 &  \thead{ Fixed\\L=2} & \thead{ Fixed\\L=4} & \thead{ Fixed\\L=8} & \thead{ Fixed\\L=16} & \thead{ Fixed\\L=20} & \thead{ Fixed\\L=32}  &  \thead{Adaptive\\Position}& \thead{Adaptive\\Ins\_position} & \thead{Adaptive\\Ins\_vec\_pos} & \thead{Adaptive\\Pos\_ins\_vec}   \\
\hline
Boolq & $79.08$ & $\done81.87$ & $82.57$   & $75.11$ & $81.20$ & $\textbf{84.80}$     & $81.83$ & $80.15$ & $81.50$ & $\textbf{81.87}$ \\
MultiRC & $67.66$ & $\done76.13$ & $75.64$     & $\textbf{79.19}$ & $58.00$ & $68.52$     & $\textbf{79.08}$ & $70.15$ & $75.39$ & $73.64$  \\
WiC & $57.84$ & $\done67.87$ & $65.05$     & $\textbf{70.22}$ & $69.30$ & $69.75$     & $67.87$ & $67.40$ & $\textbf{71.47}$ & $69.75$ \\
CB & $\textbf{96.43}$ & $\done89.29$ & $80.36$     & $82.14$ & $87.50$ & $82.14$     & $87.50$ & $87.50$ &  $\textbf{98.21}$ & $96.43$\\
RTE & $74.73$ & $\done77.62$ & $82.31$     & $\textbf{84.48}$ & $82.60$ & $82.67$     & $80.86$ & $77.26$ & $\textbf{84.12}$ & $81.23$ \\
\hline
Avg. & $75.15$ & $\done \textbf{78.55}$ & $77.18$     & $78.23$ & $75.72$ & $77.58$     & $79.43$ & $76.49$ & $\textbf{82.14}$ & $80.58$\\
\hline
\end{tabular}
}
\end{table}	

\subsection{Additional results on P-tuning V2}\label{sec:apppt2}

\begin{table}[htbp]
\centering
\caption{Comparison of PT-2 with Adaptive Position and original PT-2 on SuperGLUE datasets.
}\label{tab:pv2}
\small{
\begin{tabular}{c|cc|cc}
\hline
{\color[HTML]{000000} }                       & \multicolumn{2}{c|}{{\color[HTML]{000000} BERT-Large}}                                                   & \multicolumn{2}{c}{{\color[HTML]{000000} RoBERTa-Large}}                                                \\ \cline{2-5} 
\multirow{-2}{*}{{\color[HTML]{000000} }}     & \multicolumn{1}{c|}{{\color[HTML]{000000} PT-2}} & {\color[HTML]{000000} PT-2+\textit{adap\_pos} (Ours)} & \multicolumn{1}{c|}{{\color[HTML]{000000} PT-2}} & {\color[HTML]{000000} PT-2+\textit{adap\_pos} (Ours)} \\ \hline
{\color[HTML]{000000} \textbf{BoolQ}}         & \multicolumn{1}{c|}{{\color[HTML]{000000} 73.46}} & {\color[HTML]{000000} 74.62}                         & \multicolumn{1}{c|}{{\color[HTML]{000000} 84.07}} & {\color[HTML]{000000} 84.46}                         \\ \hline
{\color[HTML]{000000} \textbf{MultiRC (F1a)}} & \multicolumn{1}{c|}{{\color[HTML]{000000} 66.05}} & {\color[HTML]{000000} 66.05}                         & \multicolumn{1}{c|}{{\color[HTML]{000000} 70.72}} & {\color[HTML]{000000} 71.76}                         \\ \hline
{\color[HTML]{000000} \textbf{WiC}}           & \multicolumn{1}{c|}{{\color[HTML]{000000} 69.59}} & {\color[HTML]{000000} 74.61}                         & \multicolumn{1}{c|}{{\color[HTML]{000000} 70.69}} & {\color[HTML]{000000} 72.73}                         \\ \hline
{\color[HTML]{000000} \textbf{CB}}            & \multicolumn{1}{c|}{{\color[HTML]{000000} 83.93}} & {\color[HTML]{000000} 83.93}                         & \multicolumn{1}{c|}{{\color[HTML]{000000} 94.64}} & {\color[HTML]{000000} 96.43}                         \\ \hline
{\color[HTML]{000000} \textbf{RTE}}           & \multicolumn{1}{c|}{{\color[HTML]{000000} 76.90}}  & {\color[HTML]{000000} 76.90}                          & \multicolumn{1}{c|}{{\color[HTML]{000000} 86.64}} & {\color[HTML]{000000} 88.81}                         \\ \hline
{\color[HTML]{000000} \textbf{COPA}}           & \multicolumn{1}{c|}{{\color[HTML]{000000} 74.00}}  & {\color[HTML]{000000} 76.00}                          & \multicolumn{1}{c|}{{\color[HTML]{000000} 86.00}} & {\color[HTML]{000000} 88.00}                         \\ \hline
{\color[HTML]{000000} \textbf{WSC}}           & \multicolumn{1}{c|}{{\color[HTML]{000000} 63.46}}  & {\color[HTML]{000000} 69.23}                          & \multicolumn{1}{c|}{{\color[HTML]{000000} 63.46}} & {\color[HTML]{000000} 63.46}                         \\ \hline
{\color[HTML]{000000} \textbf{ReCoRD(F1)}}           & \multicolumn{1}{c|}{{\color[HTML]{000000} 66.02}}  & {\color[HTML]{000000} 66.05}                          & \multicolumn{1}{c|}{{\color[HTML]{000000} 70.72}} & {\color[HTML]{000000} 72.01}                         \\ \hline
{\color[HTML]{000000} \textbf{Avg.}}          & \multicolumn{1}{c}{{\color[HTML]{000000} 71.68}} & {\color[HTML]{000000} 73.42}                         & \multicolumn{1}{c}{{\color[HTML]{000000} 78.37}} & {\color[HTML]{000000} 79.71}                         \\ %\hline
                                              & \multicolumn{1}{c}{}                             & {\color[HTML]{3166FF} +1.74}                         & \multicolumn{1}{c}{}                             & {\color[HTML]{3166FF} +1.34}                         \\ \hline
\end{tabular}
}
\end{table}

To ensure a fair comparison, we adopted the identical setup employed by Tang et al. \citep{Liu2021PTuningVP}, employing the backbone models BERT-Large \citep{devlin2018bert} and RoBERTa-Large \citep{liu2019roberta} on the SuperGlue datasets. The detailed results of the two models on the SuperGlue dataset are depicted in Table \ref{tab:pv2}. The obtained results serve as a testament to the efficacy of our adaptive insertion position approach. Remarkably, significant performance improvements are observed across the majority of datasets. The underlying theoretical foundation, as elaborated in Sec. \ref{unified}, sheds light on the essence of our technique. By enabling soft prompts to encompass the input, we are able to capture supplementary semantic information that conventional prefix or postfix prompt tuning methods fail to capture.

As we delve deeper into the manipulation of prompts across multiple transformer layers, the effectiveness of our dynamic insertion position approach becomes increasingly apparent. This observation highlights the inherent power and adaptability of our methodology, as it successfully exploits the intricacies and hierarchies within the transformer architecture to enhance performance. By dynamically optimizing the insertion position of prompts, we can tap into additional layers of contextual understanding, enabling our approach to surpass traditional methods.

Overall, the results demonstrate that our dynamic insertion position approach is a promising avenue for enhancing the performance of downstream tasks. It offers a novel perspective on prompt adaptation, leveraging the strengths of pre-trained transformers to capture a more comprehensive representation of the input data.

\subsection{Additional results on vision prompt tuning (VPT)}\label{sec:vpt_app}
\begin{figure}[ht]
    \centering
    \subfigure[VPT-shallow]{\includegraphics[width=2.5in]{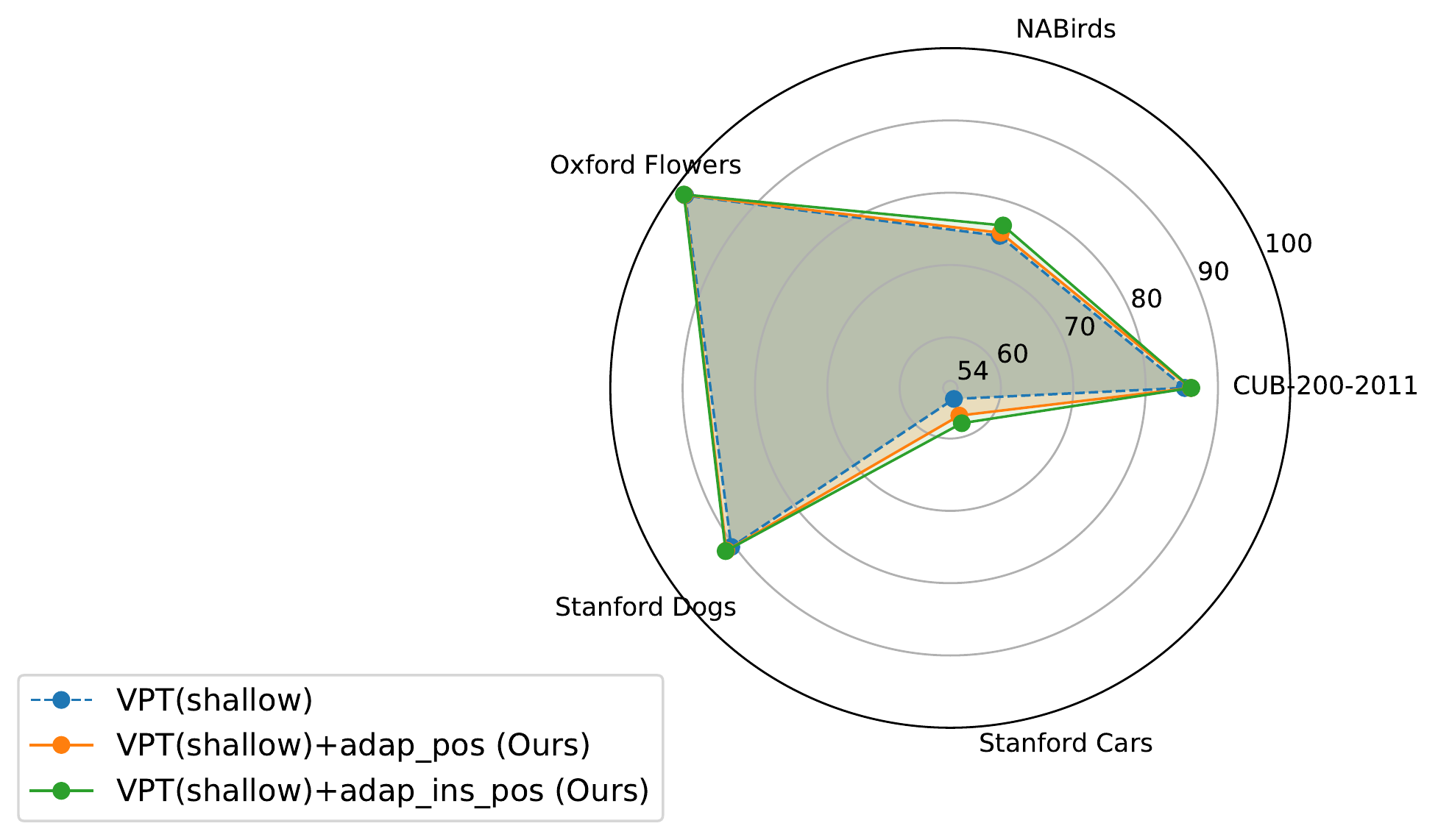}\label{fig:vpt_shallow}}
     \hspace{1cm}
    \subfigure[VPT-deep]{\includegraphics[width=2.5in]{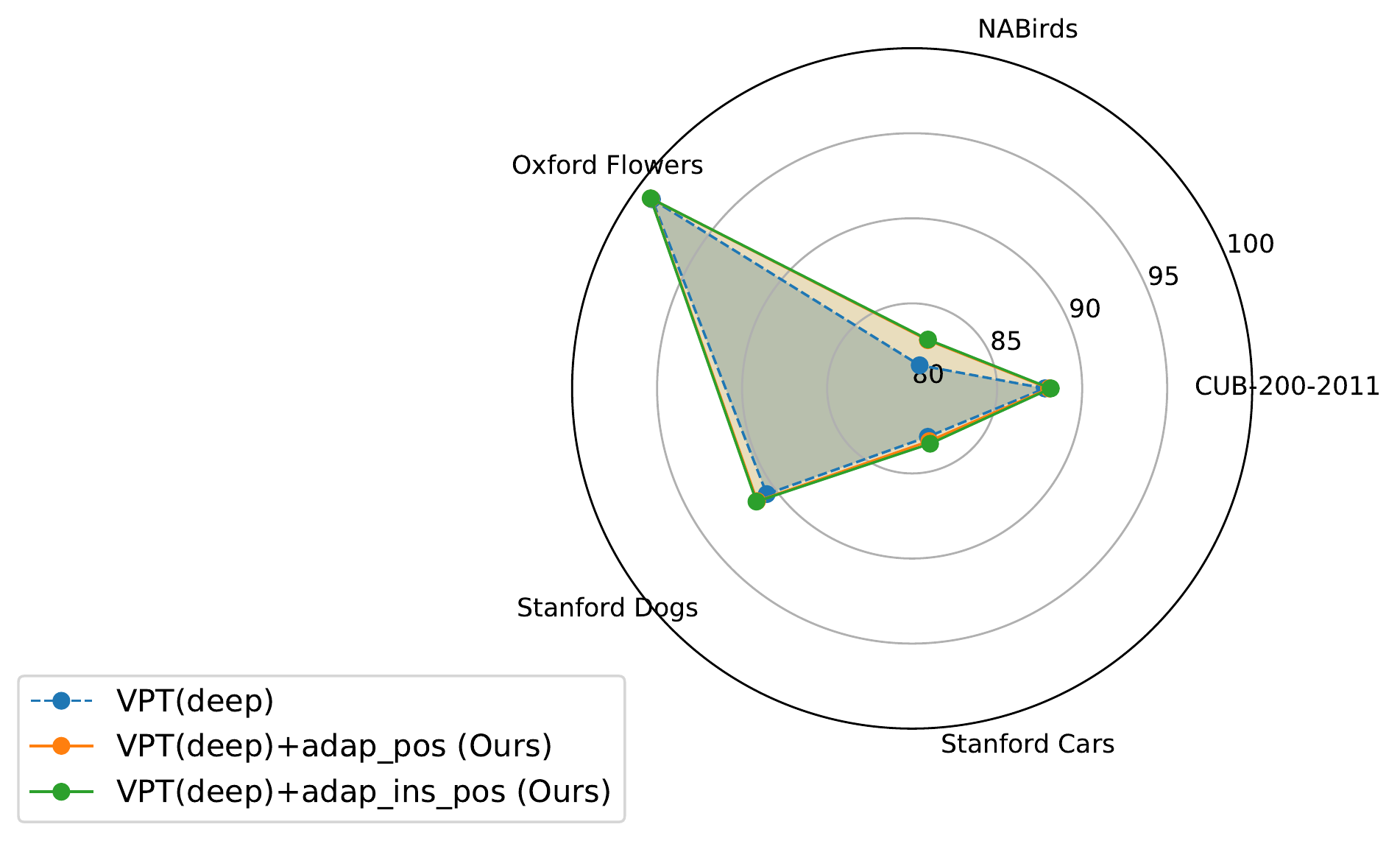}}\label{fig:vpt_deep}
    \caption{Comparison of basic VPT model and VPT with adaptive position (ours) on different datasets. }
    \label{fig:vpt}\vspace{-0.3cm}
\end{figure}

In the realm of adapting large pre-trained Transformers for downstream vision tasks, a remarkable piece of work known as Vision Prompt Tuning (VPT) has emerged. We thus further include our methods in the VPT framework. We follow the same setting as in \citep{jia2022vpt} and optimize insertion position on both shallow and deep settings. The results are depicted in Figure \ref{fig:vpt}, which demonstrates the performance gains achieved by dynamically optimizing the insertion position for prompts. Remarkably, these benefits manifest across both shallow and deep settings of VPT, underscoring the robustness and efficacy of our approach.

\subsection{Additional results on vision language modeling}\label{sec:vision_language_app}
\begin{figure}
\centering
    \includegraphics[width=.48\textwidth]{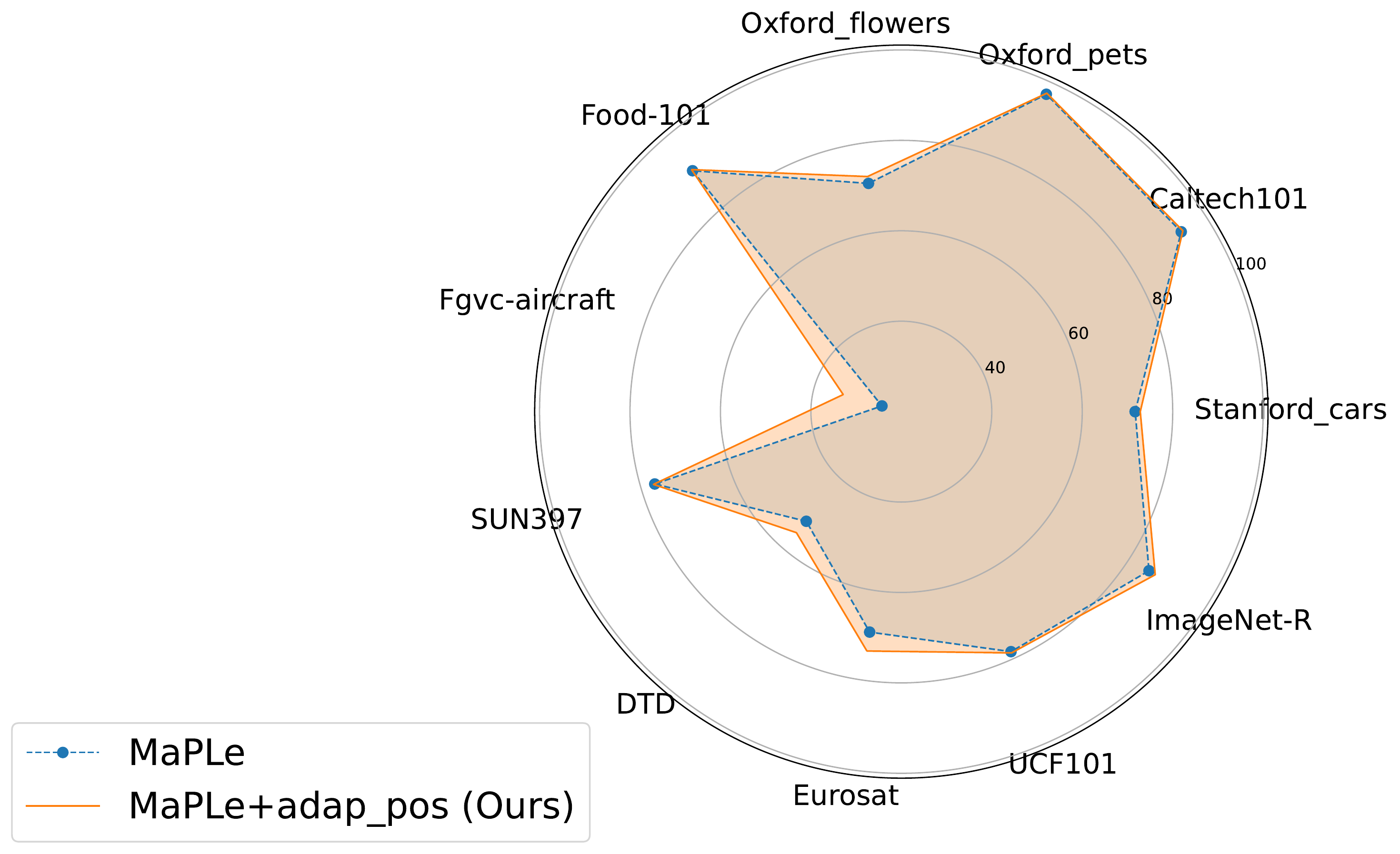} 
    \caption{\label{fig:vision_language} Comparison of MaPLe with dynamic prompt position and original MaPLe average over 11 datasets on the vision-language pretrained model prompt tuning for novel class generalization task. MaPLe with our adaptive position surpasses the original MaPLe on 11 diverse image recognition datasets for novel class generalization tasks. }
\end{figure}

\begin{table*}[htbp]
\centering
\caption{Comparison of vision language model prompt tuning results between basic MaPLe model and MaPLe with adaptive position (ours) average over 11 datasets.
}\label{tab:visionlanguage}
\setlength\tabcolsep{2.5pt}
\scalebox{1.}{
\scriptsize{
\begin{tabular}{
>{\columncolor[HTML]{FFFFFF}}c |
>{\columncolor[HTML]{FFFFFF}}c 
>{\columncolor[HTML]{FFFFFF}}c |
>{\columncolor[HTML]{FFFFFF}}c 
>{\columncolor[HTML]{FFFFFF}}c |
>{\columncolor[HTML]{FFFFFF}}c |
>{\columncolor[HTML]{FFFFFF}}c 
>{\columncolor[HTML]{FFFFFF}}c 
>{\columncolor[HTML]{FFFFFF}}c 
>{\columncolor[HTML]{FFFFFF}}c |
>{\columncolor[HTML]{FFFFFF}}c }
\hline
\cellcolor[HTML]{FFFFFF}                          & \multicolumn{5}{c|}{\cellcolor[HTML]{FFFFFF}MaPLe}                                                                                                                                                                                                                                                                                       & \multicolumn{5}{c}{\cellcolor[HTML]{FFFFFF}MaPLe + \textit{adap\_pos} (Ours)}                                                                                                                                                                                                                                                                                                                 \\ \cline{2-11} 
\cellcolor[HTML]{FFFFFF}                          & \multicolumn{2}{c|}{\cellcolor[HTML]{FFFFFF}Base}                & \multicolumn{2}{c|}{\cellcolor[HTML]{FFFFFF}Novel}               & \cellcolor[HTML]{FFFFFF}                                                                                 & \multicolumn{2}{c|}{\cellcolor[HTML]{FFFFFF}Base}                                                                             & \multicolumn{2}{c|}{\cellcolor[HTML]{FFFFFF}Novel}                               & \cellcolor[HTML]{FFFFFF}                                                                                 \\ \cline{2-5} \cline{7-10}
\multirow{-3}{*}{\cellcolor[HTML]{FFFFFF}Dataset} & \multicolumn{1}{c|}{\cellcolor[HTML]{FFFFFF}Acc.}    & Macro\_F1 & \multicolumn{1}{c|}{\cellcolor[HTML]{FFFFFF}Acc.}    & Macro\_F1 & \multirow{-2}{*}{\cellcolor[HTML]{FFFFFF}\begin{tabular}[c]{@{}c@{}}HM\\ (base+novel Acc.)\end{tabular}} & \multicolumn{1}{c|}{\cellcolor[HTML]{FFFFFF}Acc.}             & \multicolumn{1}{c|}{\cellcolor[HTML]{FFFFFF}Macro\_F1}        & \multicolumn{1}{c|}{\cellcolor[HTML]{FFFFFF}Acc.}             & Macro\_F1        & \multirow{-2}{*}{\cellcolor[HTML]{FFFFFF}\begin{tabular}[c]{@{}c@{}}HM\\ (base+novel Acc.)\end{tabular}} \\ \hline
stanford\_cars                                    & \multicolumn{1}{c|}{\cellcolor[HTML]{FFFFFF}76.43} & 75.70   & \multicolumn{1}{c|}{\cellcolor[HTML]{FFFFFF}71.70} & 69.97   & 73.99                                                                                                  & \multicolumn{1}{c|}{\cellcolor[HTML]{FFFFFF}76.50}          & \multicolumn{1}{c|}{\cellcolor[HTML]{FFFFFF}75.87}          & \multicolumn{1}{c|}{\cellcolor[HTML]{FFFFFF}72.83}          & 71.37          & 74.62                                                                                                  \\ \hline
caltech101                                        & \multicolumn{1}{c|}{\cellcolor[HTML]{FFFFFF}98.37} & 96.77   & \multicolumn{1}{c|}{\cellcolor[HTML]{FFFFFF}93.57} & 93.43   & 95.91                                                                                                  & \multicolumn{1}{c|}{\cellcolor[HTML]{FFFFFF}98.40}          & \multicolumn{1}{c|}{\cellcolor[HTML]{FFFFFF}96.83}          & \multicolumn{1}{c|}{\cellcolor[HTML]{FFFFFF}94.00}          & 93.63          & 96.15                                                                                                  \\ \hline
oxford\_pets                                      & \multicolumn{1}{c|}{\cellcolor[HTML]{FFFFFF}95.23} & 95.23   & \multicolumn{1}{c|}{\cellcolor[HTML]{FFFFFF}97.17} & 97.17   & 96.19                                                                                                  & \multicolumn{1}{c|}{\cellcolor[HTML]{FFFFFF}95.47}          & \multicolumn{1}{c|}{\cellcolor[HTML]{FFFFFF}95.47}          & \multicolumn{1}{c|}{\cellcolor[HTML]{FFFFFF}97.43}          & 97.43          & 96.44                                                                                                  \\ \hline
oxford\_flowers                                   & \multicolumn{1}{c|}{\cellcolor[HTML]{FFFFFF}97.53} & 97.37   & \multicolumn{1}{c|}{\cellcolor[HTML]{FFFFFF}71.00} & 65.50   & 82.17                                                                                                  & \multicolumn{1}{c|}{\cellcolor[HTML]{FFFFFF}97.07}          & \multicolumn{1}{c|}{\cellcolor[HTML]{FFFFFF}96.90}          & \multicolumn{1}{c|}{\cellcolor[HTML]{FFFFFF}72.53}          & 67.47          & 83.02                                                                                                  \\ \hline
{\color[HTML]{1F2328} food-101}                   & \multicolumn{1}{c|}{\cellcolor[HTML]{FFFFFF}89.87} & 89.83   & \multicolumn{1}{c|}{\cellcolor[HTML]{FFFFFF}90.50} & 90.50   & 90.18                                                                                                  & \multicolumn{1}{c|}{\cellcolor[HTML]{FFFFFF}90.03}          & \multicolumn{1}{c|}{\cellcolor[HTML]{FFFFFF}90.03}          & \multicolumn{1}{c|}{\cellcolor[HTML]{FFFFFF}90.83}          & 90.80          & 90.43                                                                                                  \\ \hline
{\color[HTML]{1F2328} fgvc-aircraft}              & \multicolumn{1}{c|}{\cellcolor[HTML]{FFFFFF}39.83} & 37.53   & \multicolumn{1}{c|}{\cellcolor[HTML]{FFFFFF}24.47} & 21.13   & 21.54                                                                                                  & \multicolumn{1}{c|}{\cellcolor[HTML]{FFFFFF}40.17}          & \multicolumn{1}{c|}{\cellcolor[HTML]{FFFFFF}37.70}          & \multicolumn{1}{c|}{\cellcolor[HTML]{FFFFFF}33.40}          & 28.70          & 36.34                                                                                                  \\ \hline
{\color[HTML]{1F2328} SUN397}                     & \multicolumn{1}{c|}{\cellcolor[HTML]{FFFFFF}81.50} & 81.27   & \multicolumn{1}{c|}{\cellcolor[HTML]{FFFFFF}76.83} & 76.03   & 79.09                                                                                                  & \multicolumn{1}{c|}{\cellcolor[HTML]{FFFFFF}81.57}          & \multicolumn{1}{c|}{\cellcolor[HTML]{FFFFFF}81.37}          & \multicolumn{1}{c|}{\cellcolor[HTML]{FFFFFF}77.13}          & 76.30          & 79.29                                                                                                  \\ \hline
{\color[HTML]{1F2328} DTD}                        & \multicolumn{1}{c|}{\cellcolor[HTML]{FFFFFF}79.97} & 79.77   & \multicolumn{1}{c|}{\cellcolor[HTML]{FFFFFF}52.10} & 50.87   & 62.94                                                                                                  & \multicolumn{1}{c|}{\cellcolor[HTML]{FFFFFF}81.07}          & \multicolumn{1}{c|}{\cellcolor[HTML]{FFFFFF}81.00}          & \multicolumn{1}{c|}{\cellcolor[HTML]{FFFFFF}55.43}          & 53.53          & 65.71                                                                                                  \\ \hline
{\color[HTML]{1F2328} eurosat}                    & \multicolumn{1}{c|}{\cellcolor[HTML]{FFFFFF}91.40} & 91.40   & \multicolumn{1}{c|}{\cellcolor[HTML]{FFFFFF}69.27} & 65.33   & 78.76                                                                                                  & \multicolumn{1}{c|}{\cellcolor[HTML]{FFFFFF}92.27}          & \multicolumn{1}{c|}{\cellcolor[HTML]{FFFFFF}92.27}          & \multicolumn{1}{c|}{\cellcolor[HTML]{FFFFFF}73.50}          & 71.30          & 81.63                                                                                                  \\ \hline
{\color[HTML]{1F2328} UCF101}                     & \multicolumn{1}{c|}{\cellcolor[HTML]{FFFFFF}84.20} & 83.17   & \multicolumn{1}{c|}{\cellcolor[HTML]{FFFFFF}78.37} & 76.20   & 81.12                                                                                                  & \multicolumn{1}{c|}{\cellcolor[HTML]{FFFFFF}84.27}          & \multicolumn{1}{c|}{\cellcolor[HTML]{FFFFFF}83.20}          & \multicolumn{1}{c|}{\cellcolor[HTML]{FFFFFF}78.67}          & 76.53          & 81.35                                                                                                  \\ \hline
{\color[HTML]{1F2328} imagenet-r}                 & \multicolumn{1}{c|}{\cellcolor[HTML]{FFFFFF}86.80} & 85.03   & \multicolumn{1}{c|}{\cellcolor[HTML]{FFFFFF}85.10} & 85.03   & 85.94                                                                                                  & \multicolumn{1}{c|}{\cellcolor[HTML]{FFFFFF}86.73}          & \multicolumn{1}{c|}{\cellcolor[HTML]{FFFFFF}85.03}          & \multicolumn{1}{c|}{\cellcolor[HTML]{FFFFFF}86.73}          & 85.03          & 86.73                                                                                                  \\ \hline
{\color[HTML]{1F2328} \textbf{Avg.}}               & \multicolumn{1}{c|}{\cellcolor[HTML]{FFFFFF}83.74} & 83.01   & \multicolumn{1}{c|}{\cellcolor[HTML]{FFFFFF}73.64} & 71.92   & 77.08                                                                                                  & \multicolumn{1}{c|}{\cellcolor[HTML]{FFFFFF}\textbf{83.96}} & \multicolumn{1}{c|}{\cellcolor[HTML]{FFFFFF}\textbf{83.24}} & \multicolumn{1}{c|}{\cellcolor[HTML]{FFFFFF}\textbf{75.68}} & \textbf{73.83} & \textbf{79.25}                                                                                         \\ \hline
\end{tabular}
}
}
\end{table*}
\begin{table}
\centering

\caption{Comparison with MaPLe on base-to-novel generalization. The adaptive position of the prompt on MaPLe will improve generalization performance over existing methods on 11 recognition datasets. Absolute gains over basic MaPLe are indicated in blue.
}
\label{tab:visionLanguage_detail}
\subtable[stanford\_cars]{
\setlength\tabcolsep{2.5pt}
\scalebox{1.}{
\scriptsize{
\begin{tabular}{
>{\columncolor[HTML]{FFFFFF}}c 
>{\columncolor[HTML]{FFFFFF}}c 
>{\columncolor[HTML]{FFFFFF}}c |
>{\columncolor[HTML]{FFFFFF}}c }
\hline
\multicolumn{1}{c}{\cellcolor[HTML]{FFFFFF}Method}                        & Base Acc. & Novel Acc. & \begin{tabular}[c]{@{}c@{}}HM \\ (Base+Novel)\end{tabular} \\ \hline
\multicolumn{1}{c}{\cellcolor[HTML]{FFFFFF}MaPLe}                         & 76.43   & 71.70    & 73.97                                                    \\ \hline
\multicolumn{1}{c}{\cellcolor[HTML]{FFFFFF}MaPLe+\textit{adap\_pos}} & 76.50   & 72.83    & 74.61                                                    \\ %\hline
                                                                            & \textcolor{blue!50}{+0.07}      & \textcolor{blue!50}{+1.13}       & \textcolor{blue!50}{+0.64}                                                        \\ \cline{2-4} 
\end{tabular}
}
}
\label{tab:stanfordcars}
}
\subtable[caltech101]{
\setlength\tabcolsep{2.5pt}
\scalebox{1.}{
\scriptsize{
\begin{tabular}{
>{\columncolor[HTML]{FFFFFF}}c 
>{\columncolor[HTML]{FFFFFF}}c 
>{\columncolor[HTML]{FFFFFF}}c |
>{\columncolor[HTML]{FFFFFF}}c }
\hline
\multicolumn{1}{c}{\cellcolor[HTML]{FFFFFF}Method}                        & Base Acc. & Novel Acc. & \begin{tabular}[c]{@{}c@{}}HM \\ (Base+Novel)\end{tabular} \\ \hline
\multicolumn{1}{c}{\cellcolor[HTML]{FFFFFF}MaPLe}                         & 98.37   & 93.57    & 95.91                                                    \\ \hline
\multicolumn{1}{c}{\cellcolor[HTML]{FFFFFF}MaPLe+\textit{adap\_pos}} & 98.40   & 94.00    & 96.15                                                    \\ %\hline
                                                                            & \textcolor{blue!50}{+0.03}      & \textcolor{blue!50}{+0.43}       & \textcolor{blue!50}{+0.24}                                                        \\ \cline{2-4} 
\end{tabular}
}
}
\label{tab:caltech101}
}

\subtable[oxford\_pets]{
\setlength\tabcolsep{2.5pt}
\scalebox{1.}{
\scriptsize{
\begin{tabular}{
>{\columncolor[HTML]{FFFFFF}}c 
>{\columncolor[HTML]{FFFFFF}}c 
>{\columncolor[HTML]{FFFFFF}}c |
>{\columncolor[HTML]{FFFFFF}}c }
\hline
\multicolumn{1}{c}{\cellcolor[HTML]{FFFFFF}Method}                        & Base Acc. & Novel Acc. & \begin{tabular}[c]{@{}c@{}}HM \\ (Base+Novel)\end{tabular} \\ \hline
\multicolumn{1}{c}{\cellcolor[HTML]{FFFFFF}MaPLe}                         & 95.23   & 97.17    & 96.19                                                    \\ \hline
\multicolumn{1}{c}{\cellcolor[HTML]{FFFFFF}MaPLe+\textit{adap\_pos}} & 95.46   & 97.43    & 96.44                                                    \\ %\hline
                                                                            & \textcolor{blue!50}{+0.23}      & \textcolor{blue!50}{+0.26}       & \textcolor{blue!50}{+0.25}                                                        \\ \cline{2-4} 
\end{tabular}
}
}
\label{tab:oxfordpets}
}
\subtable[oxford\_flowers]{
\setlength\tabcolsep{2.5pt}
\scalebox{1.}{
\scriptsize{
\begin{tabular}{
>{\columncolor[HTML]{FFFFFF}}c 
>{\columncolor[HTML]{FFFFFF}}c 
>{\columncolor[HTML]{FFFFFF}}c |
>{\columncolor[HTML]{FFFFFF}}c }
\hline
\multicolumn{1}{c}{\cellcolor[HTML]{FFFFFF}Method}                        & Base Acc. & Novel Acc. & \begin{tabular}[c]{@{}c@{}}HM \\ (Base+Novel)\end{tabular} \\ \hline
\multicolumn{1}{c}{\cellcolor[HTML]{FFFFFF}MaPLe}                         & 97.53   & 71.00    & 82.17                                                    \\ \hline
\multicolumn{1}{c}{\cellcolor[HTML]{FFFFFF}MaPLe+\textit{adap\_pos}} & 97.07   & 72.53    & 83.02                                                    \\ %\hline
                                                                            & \textcolor{red!50}{-0.46}      & \textcolor{blue!50}{+1.53}       & \textcolor{blue!50}{+0.85}                                                        \\ \cline{2-4} 
\end{tabular}
}
}
\label{tab:oxfordflowers}
}

\subtable[food-101]{
\setlength\tabcolsep{2.5pt}
\scalebox{1.}{
\scriptsize{
\begin{tabular}{
>{\columncolor[HTML]{FFFFFF}}c 
>{\columncolor[HTML]{FFFFFF}}c 
>{\columncolor[HTML]{FFFFFF}}c |
>{\columncolor[HTML]{FFFFFF}}c }
\hline
\multicolumn{1}{c}{\cellcolor[HTML]{FFFFFF}Method}                        & Base Acc. & Novel Acc. & \begin{tabular}[c]{@{}c@{}}HM \\ (Base+Novel)\end{tabular} \\ \hline
\multicolumn{1}{c}{\cellcolor[HTML]{FFFFFF}MaPLe}                         & 89.87   & 90.50    & 90.18                                                    \\ \hline
\multicolumn{1}{c}{\cellcolor[HTML]{FFFFFF}MaPLe+\textit{adap\_pos}} & 90.03   & 90.83    & 90.43                                                    \\ %\hline
                                                                            & \textcolor{blue!50}{+0.16}      & \textcolor{blue!50}{+0.33}       & \textcolor{blue!50}{+0.25}                                                        \\ \cline{2-4} 
\end{tabular}
}
}
\label{tab:food101}
}
\subtable[fgvc\_aircraft]{
\setlength\tabcolsep{2.5pt}
\scalebox{1.}{
\scriptsize{
\begin{tabular}{
>{\columncolor[HTML]{FFFFFF}}c 
>{\columncolor[HTML]{FFFFFF}}c 
>{\columncolor[HTML]{FFFFFF}}c |
>{\columncolor[HTML]{FFFFFF}}c }
\hline
\multicolumn{1}{c}{\cellcolor[HTML]{FFFFFF}Method}                        & Base Acc. & Novel Acc. & \begin{tabular}[c]{@{}c@{}}HM \\ (Base+Novel)\end{tabular} \\ \hline
\multicolumn{1}{c}{\cellcolor[HTML]{FFFFFF}MaPLe}                         & 39.83   & 24.47    & 21.54                                                    \\ \hline
\multicolumn{1}{c}{\cellcolor[HTML]{FFFFFF}MaPLe+\textit{adap\_pos}} & 40.17   & 33.40    & 36.34                                                    \\ %\hline
                                                                            & \textcolor{blue!50}{+0.34}      & \textcolor{blue!50}{+8.93}       & \textcolor{blue!50}{+14.80}                                                        \\ \cline{2-4} 
\end{tabular}
}
}
\label{tab:fgvcaircraft}
}

\subtable[SUN397]{
\setlength\tabcolsep{2.5pt}
\scalebox{1.}{
\scriptsize{
\begin{tabular}{
>{\columncolor[HTML]{FFFFFF}}c 
>{\columncolor[HTML]{FFFFFF}}c 
>{\columncolor[HTML]{FFFFFF}}c |
>{\columncolor[HTML]{FFFFFF}}c }
\hline
\multicolumn{1}{c}{\cellcolor[HTML]{FFFFFF}Method}                        & Base Acc. & Novel Acc. & \begin{tabular}[c]{@{}c@{}}HM \\ (Base+Novel)\end{tabular} \\ \hline
\multicolumn{1}{c}{\cellcolor[HTML]{FFFFFF}MaPLe}                         & 81.50   & 76.83    & 79.09                                                    \\ \hline
\multicolumn{1}{c}{\cellcolor[HTML]{FFFFFF}MaPLe+\textit{adap\_pos}} & 81.57   & 77.13    & 79.29                                                    \\ %\hline
                                                                            & \textcolor{blue!50}{+0.07}      & \textcolor{blue!50}{+0.30}       & \textcolor{blue!50}{+0.20}                                                        \\ \cline{2-4} 
\end{tabular}
}
}
\label{tab:SUN397}
}
\subtable[DTD]{
\setlength\tabcolsep{2.5pt}
\scalebox{1.}{
\scriptsize{
\begin{tabular}{
>{\columncolor[HTML]{FFFFFF}}c 
>{\columncolor[HTML]{FFFFFF}}c 
>{\columncolor[HTML]{FFFFFF}}c |
>{\columncolor[HTML]{FFFFFF}}c }
\hline
\multicolumn{1}{c}{\cellcolor[HTML]{FFFFFF}Method}                        & Base Acc. & Novel Acc. & \begin{tabular}[c]{@{}c@{}}HM \\ (Base+Novel)\end{tabular} \\ \hline
\multicolumn{1}{c}{\cellcolor[HTML]{FFFFFF}MaPLe}                         & 79.97   & 52.10    & 62.94                                                    \\ \hline
\multicolumn{1}{c}{\cellcolor[HTML]{FFFFFF}MaPLe+\textit{adap\_pos}} & 81.07   & 55.43    & 65.71                                                    \\ %\hline
                                                                            & \textcolor{blue!50}{+1.10}      & \textcolor{blue!50}{+3.33}       & \textcolor{blue!50}{+2.77}                                                        \\ \cline{2-4} 
\end{tabular}
}
}
\label{tab:DTD}
}

\subtable[eurosat]{
\setlength\tabcolsep{2.5pt}
\scalebox{1.}{
\scriptsize{
\begin{tabular}{
>{\columncolor[HTML]{FFFFFF}}c 
>{\columncolor[HTML]{FFFFFF}}c 
>{\columncolor[HTML]{FFFFFF}}c |
>{\columncolor[HTML]{FFFFFF}}c }
\hline
\multicolumn{1}{c}{\cellcolor[HTML]{FFFFFF}Method}                        & Base Acc. & Novel Acc. & \begin{tabular}[c]{@{}c@{}}HM \\ (Base+Novel)\end{tabular} \\ \hline
\multicolumn{1}{c}{\cellcolor[HTML]{FFFFFF}MaPLe}                         & 91.40   & 69.27    & 78.76                                                    \\ \hline
\multicolumn{1}{c}{\cellcolor[HTML]{FFFFFF}MaPLe+\textit{adap\_pos}} & 92.27   & 73.50    & 81.63                                                    \\ %\hline
                                                                            & \textcolor{blue!50}{+0.87}      & \textcolor{blue!50}{+4.23}       & \textcolor{blue!50}{+2.87}                                                        \\ \cline{2-4} 
\end{tabular}
}
}
\label{tab:eurosat}
}
\subtable[UCF101]{
\setlength\tabcolsep{2.5pt}
\scalebox{1.}{
\scriptsize{
\begin{tabular}{
>{\columncolor[HTML]{FFFFFF}}c 
>{\columncolor[HTML]{FFFFFF}}c 
>{\columncolor[HTML]{FFFFFF}}c |
>{\columncolor[HTML]{FFFFFF}}c }
\hline
\multicolumn{1}{c}{\cellcolor[HTML]{FFFFFF}Method}                        & Base Acc. & Novel Acc. & \begin{tabular}[c]{@{}c@{}}HM \\ (Base+Novel)\end{tabular} \\ \hline
\multicolumn{1}{c}{\cellcolor[HTML]{FFFFFF}MaPLe}                         & 84.20   & 78.37    & 81.12                                                    \\ \hline
\multicolumn{1}{c}{\cellcolor[HTML]{FFFFFF}MaPLe+\textit{adap\_pos}} & 84.27   & 78.67    & 81.35                                                    \\ %\hline
                                                                            & \textcolor{blue!50}{+0.07}      & \textcolor{blue!50}{+0.30}       & \textcolor{blue!50}{+0.23}                                                        \\ \cline{2-4} 
\end{tabular}
}
}
\label{tab:UCF101}
}

\subtable[imagenet-r]{
\setlength\tabcolsep{2.5pt}
\scalebox{1.}{
\scriptsize{
\begin{tabular}{
>{\columncolor[HTML]{FFFFFF}}c 
>{\columncolor[HTML]{FFFFFF}}c 
>{\columncolor[HTML]{FFFFFF}}c |
>{\columncolor[HTML]{FFFFFF}}c }
\hline
\multicolumn{1}{c}{\cellcolor[HTML]{FFFFFF}Method}                        & Base Acc. & Novel Acc. & \begin{tabular}[c]{@{}c@{}}HM \\ (Base+Novel)\end{tabular} \\ \hline
\multicolumn{1}{c}{\cellcolor[HTML]{FFFFFF}MaPLe}                         & 86.80   & 85.10    & 85.94                                                    \\ \hline
\multicolumn{1}{c}{\cellcolor[HTML]{FFFFFF}MaPLe+\textit{adap\_pos}} & 86.73   & 86.73    & 86.73                                                    \\ %\hline
                                                                            & \textcolor{red!50}{-0.07}      & \textcolor{blue!50}{+1.63}       & \textcolor{blue!50}{+0.79}                                                        \\ \cline{2-4} 
\end{tabular}
}
}
\label{tab:imagenetr}
}
\end{table}
Vision-language (V-L) models, such as the remarkable CLIP, have drawn significant attention recently. As the pioneering work of MaPLe \citep{khattakMaPLe} introduced a coupling function to effectively condition vision prompts based on their language counterparts, it bridges the gap between the vision and text modalities. Here, we also incorporate our adaptive insertion position approach into the text input layer, leveraging the power of dynamic prompt manipulation. We summarize our results in detail in Table \ref{tab:visionlanguage} and Table \ref{tab:visionLanguage_detail}, which substantiates the potency of our approach. By incorporating adaptive insertion position into MaPLe, we achieve an impressive absolute average gain of 2.04\% on novel classes' average accuracy and 2.17\% on the 
harmonic mean (HM) averaged over 3 runs (seeds) of both base and novel classes respectively. This substantial enhancement in performance stands as a compelling testament, providing evidence of the effectiveness and potency of our dynamic prompting methodology. It firmly establishes our approach as a powerful ingredient in the realm of Vision-Language (V-L) models, significantly elevating their capabilities and pushing the boundaries of what can be achieved.

\section{Parameter Sensitivity Analysis}\label{sec:apppara}
\begin{figure*}[ht]
\centering
    \includegraphics[width=.48\textwidth]{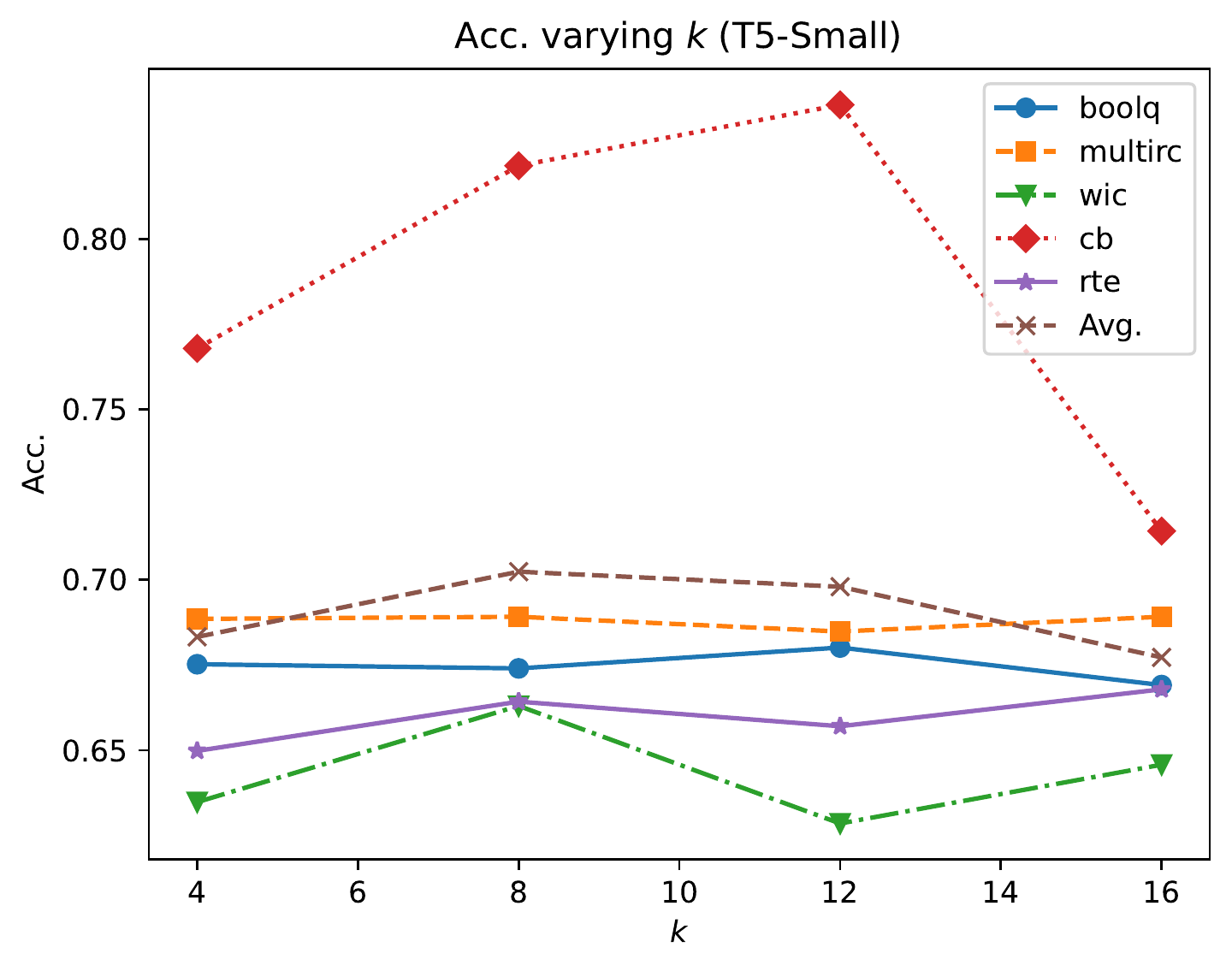} 
    \caption{Parameter Sensitivity of $k$. } \label{app:parameter}
\end{figure*}
We also analyzed the hyperparameters of our approach in this section. Basically, our approach only has one hyper-parameter, i.e., $k$ for \textit{adap\_pos\_ins\_vec} and \textit{adap\_ins\_pos\_vec}. Different choices of the number of mixture soft-prompts do not significantly impact the model performance. Using T5-Small as an example, we present the accuracy with respect to $k$ in Figure \ref{app:parameter}. The figure shows that our algorithm is not very sensitive to hyperparameter $k$. In our setting, we thus empirically set $k=8$ for all experiments. 

\end{document}